\documentclass[pdflatex,sn-mathphys-num]{sn-jnl}

\usepackage{graphicx}%
\usepackage{multirow}%
\usepackage{amsmath,amssymb,amsfonts}%
\usepackage{amsthm}%
\usepackage{pifont}
\usepackage{mathrsfs}%
\usepackage[title]{appendix}%
\usepackage[table]{xcolor}%
\usepackage{textcomp}%
\usepackage{manyfoot}%
\usepackage{caption}
\usepackage{xurl}      
\usepackage{booktabs}%
\usepackage{algorithm}%
\usepackage{algorithmicx}%
\usepackage{algpseudocode}%
\usepackage{listings}%
\usepackage{tikz}
\usepackage{array}
\usetikzlibrary{shapes.geometric, arrows.meta, positioning}
\usepackage{forest}

\definecolor{phaseA}{RGB}{70, 130, 180}   
\definecolor{phaseB}{RGB}{244, 164, 96}   
\definecolor{phaseC}{RGB}{95, 158, 160}   
\definecolor{phaseD}{RGB}{175, 140, 175}  

\definecolor{lightpink}{RGB}{250,237,234}

\usepackage{hhline}

\usepackage{xcolor}        
\usepackage{graphicx}      

\usepackage{multirow}
\usepackage{makecell}
\usepackage{float}

\usepackage{booktabs}     
\usepackage{natbib}       
\usepackage{array}        

\usepackage{bookmark}  
\bookmarksetup{rellevel=1}

\theoremstyle{thmstyleone}%
\theoremstyle{thmstyletwo}%

\theoremstyle{thmstylethree}%

\begin{document}

\title{Personalization Meets Safety: Mechanisms, Risks, and Mitigations in Personalized LLMs}

\author[1]{\fnm{Yanyan} \sur{Luo}}\email{luoyanyanit@chinamobile.com}
\author*[1]{\fnm{Xue} \sur{Han}}\email{hanxueit@chinamobile.com}
\author[1]{\fnm{Ruiqiao} \sur{Bai}}\email{bairuiqiaoit@chinamobile.com}
\author[1]{\fnm{Xin} \sur{Huang}}\email{huangxinit01@chinamobile.com}
\author[1]{\fnm{Yitong} \sur{Wang}}\email{wangyitongit@chinamobile.com}
\author[1]{\fnm{Qian} \sur{Hu}}\email{huqianit@chinamobile.com}
\author[1]{\fnm{Qing} \sur{Wang}}\email{wangqingit@chinamobile.com}
\author[1]{\fnm{Chunxu} \sur{Zhao}}\email{zhaochunxuit@chinamobile.com}
\author[1]{\fnm{Jie} \sur{Liu}}\email{liujieit@chinamobile.com}
\author[1]{\fnm{Cong} \sur{Geng}}\email{gengcongit@chinamobile.com}
\author[1]{\fnm{Lehao} \sur{Xing}}\email{xinglehaoit@chinamobile.com}
\author[2]{\fnm{Pengwei} \sur{Hu}}\email{hpw@ms.xjb.ac.cn}
\author*[1]{\fnm{Junlan} \sur{Feng}}\email{fengjunlanit@chinamobile.com}


\affil*[1]{\orgdiv{Jiutian Research}, \orgname{China Mobile Jiutian Artificial Intelligence Technology (Beijing) Co., Ltd.}, \orgaddress{\city{Beijing}, \country{China}}}

\affil[2]{\orgdiv{Xinjiang Technical Institute of Physics and Chemistry}, \orgname{Chinese Academy of Sciences}, \orgaddress{\city{Beijing}, \country{China}}}



\abstract{

Large Language Models (LLMs) have demonstrated strong capabilities across diverse real-world applications, driving growing demand for user-specific personalization that adapts model behavior to individual preferences, contexts, and long-term interaction histories. However, the core mechanisms that make personalization effective—including enhanced user-specific representation, persistent memory, adaptive fine-tuning, and autonomous agent frameworks—simultaneously reshape and expand the safety landscape in ways that existing literature has not systematically addressed. Existing reviews of LLM personalization tend to overlook safety issues, while studies focused on LLM safety mostly adopt models without personalization. This important intersection has not been thoroughly investigated. This survey provides the first comprehensive, safety-aware review of personalized LLMs. We organize the space of personalization along three dimensions—what user information to represent, how to incorporate it, and how to evaluate the result—and introduce a unified taxonomy of safety risks that spans this full landscape. At the representation level, we analyze over-exposure, sensitive attribute inference, and information persistence risks arising from structured profiles, textual personas, preference signals, and memory-based representations. Covering mainstream personalization paradigms, we delineate unique vulnerabilities inherent to prompting, retrieval augmentation, parameter fine-tuning, reinforcement learning, Mixture-of-Experts (MoE), model pruning, agent frameworks and multimodal personalization, and synthesize mitigation strategies across the model lifecycle. Beyond these fine-grained risks, we characterize paradigm-agnostic safety risks arising inherently from personalized adaptation itself, independent of implementation. We further summarize personalized datasets and evaluation methodologies. Through a case study of OpenClaw, a rapidly adopted autonomous personal agent, we analyze emerging deployment trends in real-world personalized agent ecosystems. Our analysis reveals three structural inadequacies in the existing literature: safety is evaluated as user-invariant rather than relational, personalization techniques are analyzed in isolation rather than in composition, and evaluation frameworks cannot capture emergent long-term risks. By jointly examining personalized representations, personalization paradigms, fine-grained and paradigm-agnostic safety risks, and defenses across the full lifecycle, this survey provides a unified framework for understanding and building safe personalized LLM systems and identifies key open challenges for future research.

}

\keywords{Large Language Models, Personalization, Personalized LLMs, Safety}



\maketitle

\section{Introduction}\label{sec1}

Large Language Models (LLMs) have achieved remarkable success across diverse real-world applications, including education \citep{chu-etal-2025-llm}, healthcare \citep{xu2024talk2care}, finance \citep{easin2024intelligent}, and decision support \citep{huang2024makinglargelanguagemodels}. As these systems become increasingly involved in user-facing scenarios, there is growing demand for user-specific personalization—enabling models to adapt to individual preferences, needs, and contextual constraints rather than producing uniform responses. Research on LLM personalization has advanced along three complementary dimensions. The first focuses on \textbf{what user-specific information to model}, encompassing structured profiles  \citep{wu2026personalizedsafetyllmsbenchmark}, textual personas \citep{dash2025polypersonapersonagroundedllmsynthetic,guo2025personalityguidedcodegenerationusing,zhong2023memorybankenhancinglargelanguage}, preference signals \citep{jang2023personalizedsoupspersonalizedlarge}, and long-term interaction histories \citep{zhang2025prime}.  The second investigates \textbf{how to incorporate such information}, via prompting-based methods \citep{dash2025polypersonapersonagroundedllmsynthetic,guo2025personalityguidedcodegenerationusing,zhong2023memorybankenhancinglargelanguage}, retrieval-augmented generation \citep{salemi2024memory,li2025lifealignlifelongalignmentlarge,mysore2023rag,salemi2024lamplargelanguagemodels,wu2024understandingroleuserprofile}, parameter-efficient fine-tuning \citep{pi2024personalizedvisualinstructiontuning,huang2024selectivepromptingtuningpersonalized,dash2025polypersonapersonagroundedllmsynthetic,hebert2024persomapersonalizedsoftprompt}, and reinforcement learning from personalized feedback \citep{jang2023personalizedsoupspersonalizedlarge,wu2023finegrainedhumanfeedbackgives,park2024rlhfheterogeneousfeedbackpersonalization,huang2024makinglargelanguagemodels,lee2024aligningthousandspreferencesmessage}. The third explores \textbf{how to evaluate the resulting personalization quality}, moving from generic language-quality metrics toward preference-following and long-term behavioral consistency \citep{zhao2025llmsrecognizepreferencesevaluating,salemi2025reasoningenhancedselftraininglongformpersonalized,zollo2025personalllmtailoringllmsindividual}.

Despite this progress, most personalization techniques are developed under a uniform safety and alignment framework that treats acceptable model behavior as consistent across users, regardless of individual characteristics, vulnerability levels, or contextual risk. However, safety in personalized systems is inherently user-conditional rather than universally defined. The same model response—for example, guidance on medication dosage, detailed financial risk information, or emotionally charged content—may be appropriate for one user while posing potential harm to another depending on their background, situational context, or expertise. Critically, this mismatch is not marginal: a large-scale evaluation across seven sensitive domains reports that incorporating personalized user context improves safety scores by 43.2\% \citep{wu2026personalizedsafetyllmsbenchmark}, highlighting important limitations of user-invariant safety evaluation frameworks. Complementary work evaluating twenty widely used LLMs further shows that current models exhibit limited ability to adapt their safety behavior to individual user attributes \citep{in2025safety}.


Beyond this foundational mismatch, deeper user adaptation of LLMs introduces new or amplified safety risks compared with generic deployments. First, fine-tuning on user-specific data—even benign data—may gradually degrade the safety alignment established during pre-training, as even narrow task adaptation has been shown to open pathways to harmful outputs the base model would have refused \citep{qi2023fine}. Second, retrieval-augmented personalization creates a leakage pipeline in which sensitive personal records stored in external memory are injected into model prompts, potentially exposing them to unintended access, adversarial probing, or infrastructure-level leakage; retrieval corpora can moreover be poisoned to redirect model outputs without touching any model weights \citep{rag-are-not-safer-2025-rag}. Third, long-term memory—the defining capability of personalized agents—is vulnerable to injection attacks that contaminate the memory store through seemingly benign interactions, with injected records persisting silently across sessions and activating on semantically related future queries \citep{chen2024agentpoisonredteamingllmagents,clop2024backdooredretrieverspromptinjection}. Fourth, preference-based personalization methods may reinforce sycophantic tendencies, encouraging models to prioritize user agreement over factual accuracy \citep{wei2024simplesyntheticdatareduces}. Such behavior may increase misinformation risks, particularly in high-stakes domains where reliable and evidence-grounded responses are essential. Fifth, architecture-level personalization via Mixture-of-Experts (MoE) routing may expose side-channel information through data-dependent expert selection \citep{ding2025moechoexploitingsidechannelattacks}, while personalized pruning can be exploited to embed stealthy backdoors that remain dormant in dense models but activate after compression or sparsification \citep{guo2026silentsparsebackdoorattacks}. These mechanisms further expand the attack surface of personalized deployment pipelines.



This survey provides a comprehensive and integrated review of personalized LLMs from a safety-aware perspective. We systematically examine personalization stack spanning personalized representations, adaptation techniques, architecture- and system-level personalization, and multimodal personalization. Building upon this taxonomy, we analyze their associated fine-grained safety risks and corresponding mitigation strategies throughout the personalization lifecycle. Beyond these fine-grained risks, we further summarize broader safety challenges arising from personalized adaptation itself, as well as challenges related to personalized datasets and evaluation methodologies. We also organize mitigation and evaluation strategies from three complementary perspectives and further discuss emerging deployment trends in real-world personalized agent ecosystems. By integrating personalization stack, safety risks, evaluation methodologies, and mitigation strategies within a unified framework, this survey aims to provide a foundation for future research toward safe, dependable, and deployable personalized LLM systems.

\begin{figure}[htbp]
    \centering
    \resizebox{0.82\linewidth}{!}{
    \begin{forest}
      for tree={
        draw,
        align=center,
        thick,
        fill=blue!5, 
        rounded corners=3pt,
        grow'=east,    
        child anchor=west,    
        parent anchor=east,       
        anchor=west,
        l sep=0.8 cm,   
        s sep=0.3cm,                   
        edge path={
          \noexpand\path [draw, \forestoption{edge}] 
          (!u.parent anchor) -- +(0.5,0) |- (.child anchor)\forestoption{edge label};
        },
      }
    [ {\rotatebox{-90}{\strut\fontsize{15}{15}\selectfont\sffamily\bfseries\scshape Personalization Meets Safety in LLMs}},fill=red!20,rounded corners=4pt,minimum width=1cm,minimum height=20cm,align=center
      [
        Personalized Representations and Associated
        Safety Risks \\(\textcolor{red}{\S3}),fill=violet!20
        [Representation Types\\
          (\textcolor{red}{\S3.1})
           ]
          [Representation Properties\\
          (\textcolor{red}{\S3.2})
          ]
          [Representation Management \\
          (\textcolor{red}{\S3.3})
          ]
          [Safety Concerns\\
          (\textcolor{red}{\S3.4})
          ]         
      ]
      [
        Personalization Techniques in LLM pipeline and Associated Safety Risks\\(\textcolor{red}{\S4}),fill=violet!20
        [
          Non-Parametric  Personalization\\
            (\textcolor{red}{\S4.1})
            [
              Prompting-based\\ (\textcolor{red}{\S4.1.1}) \\,fill=green!20!gray!10
            ]
            [
              Retrieval-augmented\\ (\textcolor{red}{\S4.1.2}),fill=green!20!gray!10
            ]
        ]
        [
          Parametric Personalization \\
           (\textcolor{red}{\S4.2})
            [
              Fine-tuning-based \\ (\textcolor{red}{\S4.2.1}),fill=green!20!gray!10
            ]
            [
              Reinforcement learning-based\\(\textcolor{red}{\S4.2.2}),fill=green!20!gray!10
            ]
        ]
      ]
      [
        Architecture-level Personalization and Associated Safety Risks\\(\textcolor{red}{\S5}),fill=violet!20
        [
          Mixture-of-Experts (MoE)-based\\(\textcolor{red}{\S5.1})
        ]
        [
          Pruning-based\\(\textcolor{red}{\S5.2})
        ]
      ]
      [
        System-level Personalization and Associated Safety Risks\\(\textcolor{red}{\S6}),fill=violet!20
        [
          System-level Personalization Techniques\\(\textcolor{red}{\S6.1})
        ]
        [
          Safety Risks\\(\textcolor{red}{\S6.2})
        ]
        [
          Defense and Mitigation Strategies\\(\textcolor{red}{\S6.3})
        ]
      ]
      [
        Multimodal Personalization and Associated Safety Risks\\(\textcolor{red}{\S7}),fill=violet!20
        [
          Multimodal Personalization Techniques
        ]
        [
          Safety and Privacy Risks
        ]
        [
          Defense and Mitigation Strategies
        ]
      ]
      [
        Paradigm-agnostic Safety Risks of LLM Personalization\\(\textcolor{red}{\S8}),fill=violet!20
        [
          Bias Reinforcement
        ]
        [
          Anthropomorphism
        ]
        [
          Algorithmic and Inferential Profiling
        ]
        [
          Safety Gaming and Evasion
        ]
      ]
      [
        Personalization Evaluation Benchmarks\\(\textcolor{red}{\S9}),fill=violet!20
        [
          Language Metrics
        ]
        [
          Classification Metrics
        ]
        [
          Dialogue Metircs
        ]
        [
          Personalization Safety Metrics
        ]
      ]
      [
        Mitigation and Evaluation of Safety Concerns in LLM Personalization\\(\textcolor{red}{\S10}),fill=violet!20
        [
          Training-free Solutions\\
          (\textcolor{red}{\S10.1})
        ]
        [
          Training-based Solutions\\
          (\textcolor{red}{\S10.2})
        ]
        [
           Evaluation-based Solutions\\
           (\textcolor{red}{\S10.3})
        ]
      ]
      [
        Technical Trend for Personalized Application Market\\(\textcolor{red}{\S11}),fill=violet!20
        [
          An Overview of Representative Personalized Applications in the LLM Agent Market\\(\textcolor{red}{\S11.1})
        ]
        [
          The Practical Implementation of Frontier Technologies in Personalized Agents\\(\textcolor{red}{\S11.2})
        ]
        [
          Security Risks and Challenges of Current Personalized Agent Applications \\(\textcolor{red}{\S11.3})
        ]
      ]
    ]
    \end{forest}}
    \caption{Overview of the personalization landscape in LLMs and its associated safety implications. The figure summarizes personalized representations and personalization paradigms, together with their emerging safety risks and corresponding mitigations across the personalization stack, as well as broader risks arising inherently from personalized adaptation, together with personalization evaluation benchmarks. It further groups mitigation and evaluation strategies into three categories: training-based approaches, training-free methods, and personalized safety evaluation frameworks, and additionally summarizes deployment considerations and emerging technical trends in personalized intelligent agent ecosystems.}
    \label{taxnomy_overview}
\end{figure}
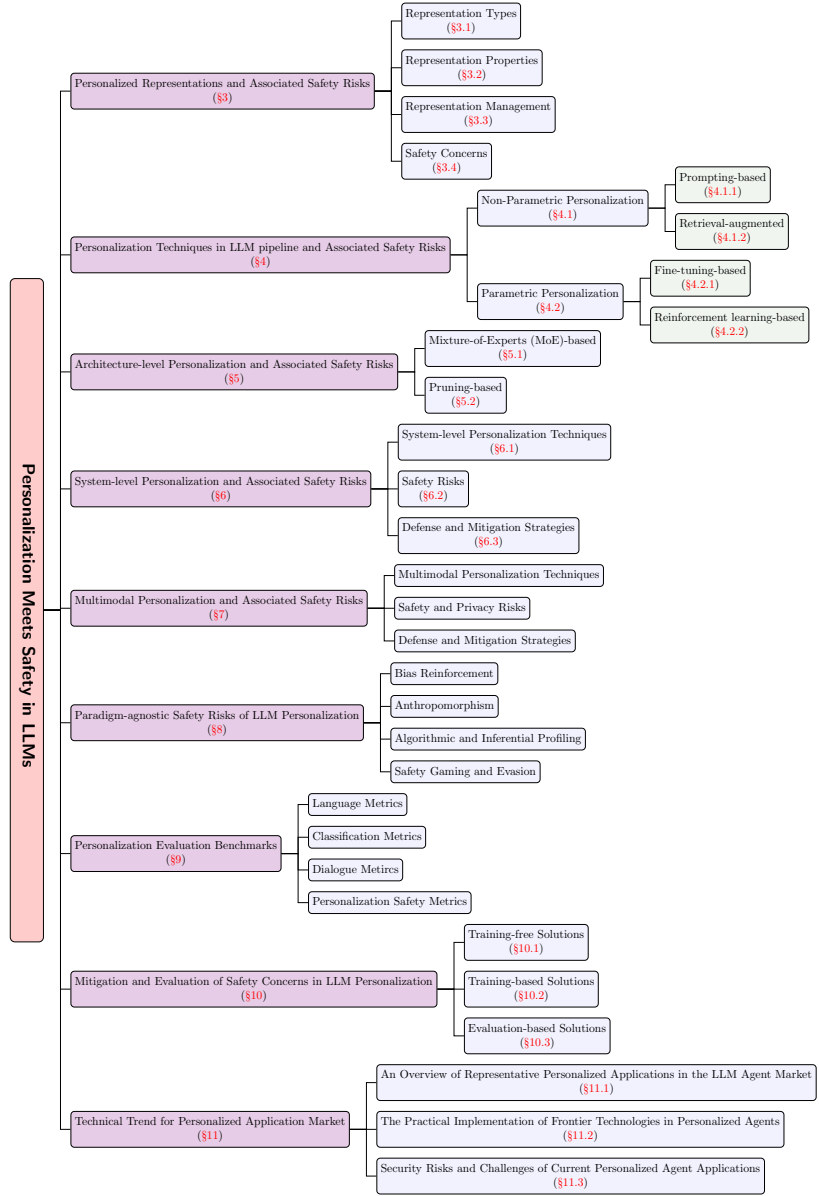

\section{A Safety-aware Taxonomy of Personalized LLMs}


This section surveys recent advances in personalized LLMs through the lens of the taxonomy illustrated in Figure \ref{taxnomy_overview}. The taxonomy provides a unified safety-aware framework for understanding personalized LLMs, systematically connecting personalized representations, personalization paradigms, safety risks, mitigation strategies, evaluation methodologies, and deployment considerations. It captures both fine-grained risks associated with different personalization mechanisms and paradigm-agnostic risks that arise inherently from personalized adaptation itself. Beyond personalization paradigms, the taxonomy further incorporates evaluation benchmarks and methodologies, highlighting emerging challenges in assessing personalization effectiveness and safety. In addition, the taxonomy systematically organizes mitigation and evaluation strategies for personalized safety risks. Finally, using OpenClaw as a representative case study, we summarize emerging technical and application trends in real-world personalized agent ecosystems.


As shown in Table \ref{table-survey}, existing surveys on LLM personalization and LLM safety often study these two areas independently. Furthermore, safety is commonly treated as a global property of LLM behavior, while the fine-grained safety implications arising throughout the personalization stack—from low-level personalized representations and adaptation mechanisms to architecture- and system-level personalization—remain insufficiently explored. Compared with existing studies, our survey provides a comprehensive, safety-aware review of personalized LLMs across the full personalization stack, systematically covering both fine-grained and paradigm-agnostic safety risks, together with their corresponding mitigation strategies.

\begin{table*}[htb]
\small
\begin{center}
\resizebox{\textwidth}{!}
{\begin{tabular}{>{\raggedright\arraybackslash}m{9cm}cccc}
\toprule
\rowcolor{gray!21}
\bf Title & \bf Personalization  & \bf Fine-grained Safety& \bf Global Safety  & \bf Threat Mitigation 
\\ \hline
 A Survey of Personalized Large Language Models: Progress and Future Directions \citep{liu2025survey} & \checkmark & \ding{55} & \ding{55} & \ding{55}  \\ 
\rowcolor{gray!21}
A Survey on Personalized Alignment--The Missing Piece for Large Language Models in Real-world Applications \citep{guan2025surveypersonalizedalignment} & \checkmark & \ding{55} & \checkmark & \ding{55}\\ 
A Survey on Personalized and Pluralistic Preference Alignment in Large Language Models \citep{xie2025surveypersonalizedpluralisticpreference} & \checkmark  & \ding{55} & \ding{55} & \ding{55}\\ 
\rowcolor{gray!21}
Personalized Generation in Large Model Era: A Survey \citep{xu2025personalized} & \checkmark  & \ding{55} & \ding{55} & \ding{55}\\ 
Personalization of Large Language Models: A Survey \citep{zhang2024personalization} & \checkmark  & \ding{55} & \ding{55} & \ding{55}\\ 
\rowcolor{gray!21}
Personalized Multimodal Large Language Models: A Survey \citep{wu2024personalized} & \checkmark  & \ding{55} & \ding{55} & \ding{55}\\ 
Two Tales of Persona in LLMs: A Survey of Role-playing and Personalization \citep{tseng2024talespersonallmssurvey} & \checkmark  & \ding{55} & \checkmark  & \ding{55}\\
\rowcolor{gray!21}
Large Language Model Safety: A Holistic Survey \citep{shi2024largelanguagemodelsafety} & \ding{55}  & \ding{55} & \checkmark  & \checkmark\\
On Large Language Models Safety, Security, and Privacy: A Survey \citep{ZHANG2025100301} & \ding{55}  & \ding{55} & \checkmark  & \checkmark\\
\rowcolor{gray!21}
Ours & \checkmark  & \checkmark & \checkmark  & \checkmark\\
\bottomrule
\end{tabular}
}
\end{center}
\caption{A summary of existing related surveys on LLM personalization and safety.}
\label{table-survey}
\end{table*}

\begin{figure}[htbp]
\centering
\includegraphics[width=\linewidth, trim=0.3cm 0.1cm 0.9cm 0.1cm,clip]{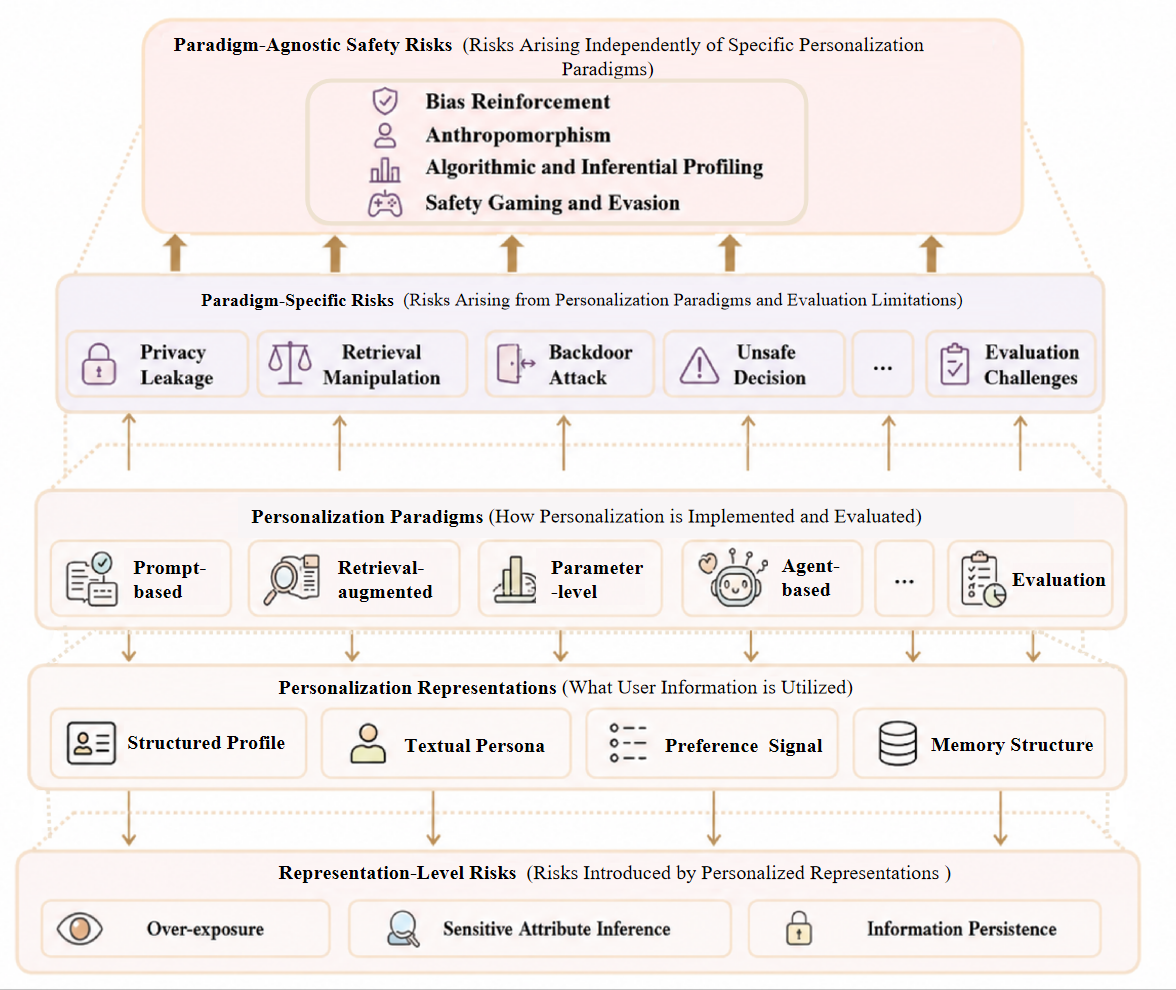}
\caption{\textbf{Overview of personalization representations, techniques, and associated safety risks}. Safety risks and challenges span the entire personalization lifecycle, from user representation to personalization implementation, interaction, and evaluation. Different representation forms expose user information at different levels. Personalization paradigms further build upon these representations to adapt model behaviors, introducing paradigm-specific vulnerabilities. For example, retrieval-augmented personalization methods may suffer from retrieval manipulation, while agent-based systems face risks related to unsafe planning and decision-making. Beyond paradigm-specific threats, some risks arise inherently from personalized adaptation itself rather than any specific personalization paradigm.}
\label{fig:overview} 
\end{figure}

Figure \ref{fig:overview} further summarizes the layered personalization stack of personalized LLMs, including representation-level personalization (what user information is utilized) and personalization paradigms (how personalization is implemented and evaluated). It also highlights the associated safety risks that emerge throughout the personalization lifecycle (risks arising from personalization paradigms and limitations in personalized safety evaluation). Different representation forms expose user information at varying semantic, behavioral, and temporal levels, thereby introducing representation-level risks such as over-exposure, sensitive attribute inference, and information persistence. Built upon these representations, personalization paradigms further adapt and evaluate model behaviors, introducing paradigm-specific vulnerabilities such as retrieval manipulation, backdoor attacks, unsafe decision-making, and evaluation challenges. Beyond these paradigm-specific threats, some risks emerge inherently from personalized adaptation itself (paradigm-agnostic safety risks), independent of the underlying personalization paradigm.

\section{Personalized Representations and Associated Safety Risks}
\label{sec:representation}

Personalized representations refer to the forms in which user-specific information is encoded and maintained within LLM systems. Instead of generating uniform responses for all users, personalized LLMs leverage such representations to adapt outputs according to individual characteristics, preferences, and interaction histories \citep{liu2025survey,zhang2024personalization}. Existing approaches employ diverse representation forms, including structured profiles, textual personas, preference signals, and memory-based representations, which further differ in properties such as observability, compressibility and temporal persistence.



Formally, let $S$ denote the set of user-related information. This information is transformed into a compact user representation, as shown in Equation \ref{eq:user_rep}.

\begin{equation}
U = f(S)
\label{eq:user_rep}
\end{equation}

\noindent where $f(\cdot)$ denotes a process that converts heterogeneous user information into a usable personalization representation, and $U$ captures compact user characteristics. Many personalized systems additionally maintain external memory structures \citep{afzoon2026persobenchbenchmarkingpersonalizedresponse} to preserve historical user-LLM interactions. Let $H$ denote the sequence of historical interactions. The corresponding memory representation $M$ can be formulated as Equation \ref{eq:user_mem}.

\begin{equation}
M = g(H)
\label{eq:user_mem}
\end{equation}

\noindent where $g(\cdot)$ denotes a memory construction process that organizes historical interactions into contextual memory representations, $M$ preserves detailed contextual information derived from past interactions. Personalized response generation $y$ can therefore be expressed as Equation \ref{eq:y}.

\begin{equation}
y = \mathrm{LLM}(x, U, M)
\label{eq:y}
\end{equation}

\noindent where $x$ is the current user query.

However, the increasing precision and richness of personalized representations also introduce new safety challenges. As models encode more detailed user information and accumulate long-term interaction histories, the risk of privacy exposure and unintended misuse of personal data correspondingly increases \citep{PPILeakage}. Against this background, understanding the forms and properties of personalized representations is essential for analyzing their associated safety risks. 

\subsection{Representation Types}

\begin{figure}[h]
\centering
\includegraphics[width=\linewidth]{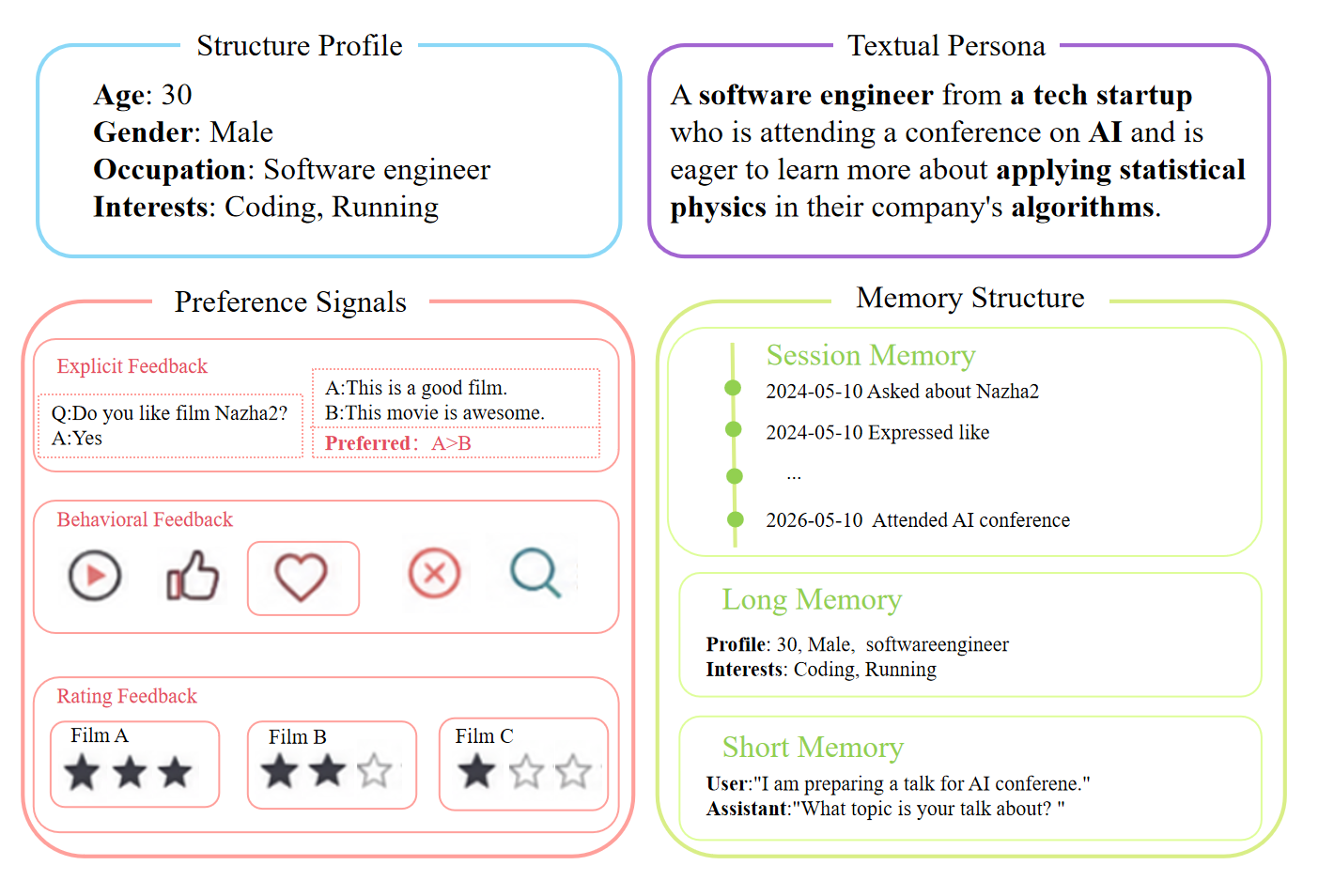}
\caption{An overview of personalization representation types. This figure presents four forms of personalization information as inputs for LLMs. \textbf{Structured profile} encodes user attributes in explicit key–value formats (e.g., age-30). \textbf{Textual persona} represents user characteristics through natural language descriptions and semantic narratives. \textbf{Preference signals} capture user preferences from explicit or implicit feedback, such as pairwise choices, ratings, clicks, corrections, and behavioral interactions. \textbf{Memory structure} organizes personalized information across different temporal scopes, including short-term conversational context, session-level interaction history, and persistent long-term user memory.}
\label{fig:representation}
\end{figure}

As illustrated in Figure \ref{fig:representation}, personalized representations generally include structured profiles, textual personas, preference signals, and memory-based representations. In practice, the first three are often incorporated into compact user representations $U$, while the last category maintains external memory representations $M$ derived from historical interactions.

\textbf{Structured profile-based representations} encode explicit user attributes, such as demographics, roles, or declared interests, in structured forms, such as key–value pairs or predefined schema fields \citep{wu2026personalizedsafetyllmsbenchmark,li2016personabasedneuralconversationmodel}. These representations typically rely on structured metadata that can be directly incorporated into prompts or contextual information of LLMs.

\textbf{Textual persona-based representations} describe or summarize user characteristics, such as personality traits, interests, or communication styles \citep{zhang2018personalizingdialogueagentsi,zhao2025llmsrecognizepreferencesevaluating,ge2025scalingsyntheticdatacreation}, through natural language statements rather than predefined schema fields. Persona-based dialogue systems commonly adopt such representations, where persona statements or summaries serve as contextual conditioning information guiding response generation.

\textbf{Preference-based representations} capture user-specific behavioral tendencies inferred from historical interactions, such as content interests, stylistic preferences, or response expectations \citep{jang2023personalizedsoupspersonalizedlarge,li20251,ryan2025synthesizemeinducingpersonaguidedprompts}. These representations enable models to better align generated responses with individual user preferences.
 
\textbf{Memory-based representations} preserve contextual information derived from historical interactions through external memory structures. Such memories may store dialogue histories, behavioral traces, or extracted factual information from past conversations, which can later be retrieved to support personalized response generation \citep{afzoon2026persobenchbenchmarkingpersonalizedresponse,liu2026simplemem,yang2026beyond}.


\subsection{Representation Properties}

Representation properties characterize the structural and functional characteristics of personalized representations, including their observability, compressibility, memory persistence and temporal evolution.

\textbf{Observability} characterizes whether user characteristics are explicitly observable or implicitly inferred from behavioral signals. Explicit characteristics are directly provided by users, such as demographic information or declared preferences, whereas implicit characteristics are inferred from interaction histories through representation learning processes, including latent interests, communication styles, and behavioral tendencies \citep{zhao2025llmsrecognizepreferencesevaluating}.

\textbf{Compressibility} characterizes the extent to which heterogeneous user information can be compactly encoded into personalized representations \citep{jiang-etal-2024-personallm,zhong2024memorybank}. Highly compressed representations typically summarize user characteristics into dense latent vectors or compact textual descriptions, thereby improving storage efficiency and inference speed. In contrast, less compressed representations may preserve richer contextual and behavioral information, but incur higher memory and computational costs. Moreover, overly verbose or less structured representations may introduce substantial redundancy, which can negatively affect memory retrieval and downstream reasoning, and may further lead to middle-context degradation phenomena \citep{liu2026simplemem}.

\textbf{Memory persistence} characterizes the extent to which personalized representations preserve and utilize long-term historical interaction information. Unlike compact representations that summarize user characteristics into aggregated representations, memory-based personalization systems maintain persistent memory structures that store dialogue histories, behavioral traces, or extracted factual knowledge from past interactions \citep{memorysurvey}. During inference, relevant information can be retrieved from memory to provide contextual grounding for personalized response generation.

\textbf{Temporal evolution} describes whether personalized representations remain static or dynamically evolve over time. Since user interests, preferences, and interaction patterns may continuously change, both compact user representations $U$ and memory representations $M$ may be updated based on newly observed interactions. Modeling temporal evolution enables personalized systems to capture shifts in user characteristics and behavioral patterns across time \citep{memorysurvey}.

\subsection{Representation Management}
\label{sec:memory_mgmt}
Beyond representation forms and properties, personalized LLM systems must additionally manage how personalized information is stored, updated, compressed, and retrieved throughout long-term interactions. Such management is essential for maintaining personalization quality, behavioral consistency, and scalability over time.

Although $U$ and $M$ are conceptually distinct, they are often tightly coupled in practical personalized systems. Earlier approaches commonly maintained $M$ as raw or extended interaction histories through full-context extension or heuristic filtering mechanisms, while more recent methods employ structured summarization and semantic compression to improve scalability and long-term consistency \citep{memorysurvey}. For example, SimpleMem is an efficient memory framework inspired by Complementary Learning Systems (CLS) theory \citep{liu2026simplemem} and built upon structured semantic compression. Conversely, compact user representations $U$ can also be dynamically summarized from retrieved information in $M$. For instance, GraphRAG constructs entity-level knowledge graphs from source documents and further generates community summaries over groups of closely related entities \citep{edge2025localglobalgraphrag}. Motivated by the central role of memory management in long-term and agent-based personalization, we next focus on memory management mechanisms in personalized LLM systems.

\textbf{Memory management mechanisms} play a critical role in the effectiveness of memory-based personalization systems. Existing methods can generally be categorized into optimization-oriented and infrastructure-oriented approaches.

Optimization-oriented approaches primarily focus on improving the efficiency and quality of memory utilization, aiming to retrieve relevant user information with low latency and minimal information loss. These methods typically optimize memory retrieval strategies, memory selection, or memory representation mechanisms for personalized interactions.

In contrast, infrastructure-oriented approaches focus on designing scalable and structured memory architectures that support long-term conversational personalization. For example, some systems \citep{liu2026simplemem} integrate semantic, lexical, and symbolic retrieval mechanisms to improve the recall of relevant interaction records, while others \citep{chhikara2025mem0} combine text-based and graph-based memory structures to better organize and maintain historical knowledge.

Table \ref{tb:per_rep_mem_app} summarizes optimization-oriented methods, while Table \ref{tb:per_rep_mem_infra} summarizes infrastructure-oriented methods. Some approaches span both categories and are therefore included in both tables.

\begin{table}[htbp]
\footnotesize
\setlength{\tabcolsep}{4pt}
  \centering
    \begin{tabular}{p{1.5cm}p{5.8cm}p{4.4cm}p{1cm}}
    \toprule
\textbf{Method Name} &
\textbf{Method Overview} &
\textbf{Problem Solved or Optimized} &
\textbf{Year} \\\midrule

MemoChat \citep{lu2023memochat} &
Memo construction technique; summarizes past dialogues into structured notes to maintain conversational consistency. &
Information loss and context drift in multi-turn, long-context dialogues.  &
2023 \\\hline

UIA \citep{zeng2023personalized} &
Personalized dense retrieval; tailors information access and search results based on unique user history and context. &
Generic information retrieval that fails to account for individual user intent.  &
2023 \\\hline

Pearl \citep{mysore2024pearl} &
Generation-calibrated retrieval; uses generation-phase feedback to refine the retrieval of personalized writing history. &
Mismatch between retrieved user history and current stylistic task requirements.  &
2024 \\\hline

Persona-db \citep{sun2025persona} &
Collaborative data refinement; distills shared persona traits while maintaining individual user nuances. &
High computational cost and noise in large-scale user persona modeling.  &
2025 \\\hline

Reasoning- Bank \citep{ouyang2025reasoningbank} &
Scaling reasoning memory; stores and reuses complex reasoning traces tailored to specific tasks or users. &
Repetitive reasoning failures and inability to scale complex logic across similar tasks.  &
2025 \\\hline

ReMe \citep{cao2025remember} &
Dynamic procedural memory; experience-driven evolution that refines task execution based on past user feedback. &
Inability to adapt procedural workflows to specific user preferences or constraints.  &
2025 \\\hline

MACLA \citep{forouzandeh2025learning} &
Bayesian procedural selection; uses contrastive refinement to select the most effective multi-step procedures for a user. &
Difficulty in selecting and refining effective multi-step procedures for personalized tasks.  &
2025 \\\hline

MemEvolve \citep{zhang2025memevolve} &
Meta-evolution of memory; agent optimizes its own memory structure and retrieval parameters over time. &
Fixed memory architectures that cannot adapt to increasing user interaction complexity.  &
2025 \\\hline

Hindsight \citep{latimer2025hindsight} &
Retention, recall, and reflection; agent reflects on past history to adjust its persona and behavioral patterns. &
Lack of self-correction and behavioral refinement in agents over long-term engagements.  &
2025 \\\hline

TACITREE \citep{li2025toward} &
Hierarchical tree reasoning; uses a large-scale dataset to infer implicit user preferences in multi-session chats. &
Inability to utilize implicit user background information in personalized conversations.  &
2025 \\\hline

EMG-RAG \citep{patel2025engram} &
Editable memory graphs; represents user data as a dynamic, updatable graph structure. &
Static knowledge representations that cannot adapt to changing user facts or preferences.  &
2025 \\\hline

memobase \citep{memobase} &
User-profile centric system; stores structured attributes and time-aware events for virtual companions. &
High latency and cost in maintaining detailed, evolving user profiles for long-term engagement. &
2025 \\\hline

LangMem \citep{langmem} &
Background memory manager; automatically extracts user preferences from chat to refine prompts and behavior. &
Manual overhead in updating user state and maintaining consistency across sessions. &
2025 \\\hline

Mem0 \citep{chhikara2025mem0} &
Adaptive personalization layer; remembers preferences across sessions and evolves through continuous learning. &
Fragmented user experiences across different applications and session boundaries. &
2025 \\\hline

SimpleMem \citep{liu2026simplemem} &
Lifelong memory module; efficient storage and retrieval of interaction history for agent state maintenance. &
Inconsistency in agent behavior across long-term interactions. &
2026 \\\hline

RP-Reasoner \citep{feng2026does} &
Rational preference benchmarking; optimizes how memory shapes reasoning regarding specific user choices. &
Failure of agents to adhere to user-specific logical preferences and choice patterns.  &
2026 \\\hline

MemRL \citep{zhang2026memrl} &
Reinforcement learning on episodic memory; agent learns successful personalized strategies through runtime feedback. &
Static agent behavior that fails to improve based on past successful interactions.  &
2026 \\\hline

MemSkill \citep{zhang2026memskill} &
Evolving memory skills; specializes in cognitive routines for managing and utilizing personalized stored information. &
Lack of specialized routines for effectively leveraging long-term personalized data.  &
2026 \\\hline

ProMem \citep{yang2026beyond} &
Proactive memory extraction; identifies and stores key conversational information before it is explicitly requested. &
Reactive memory systems that lead to fragmented personalization and delayed recall.  &
2026 \\\hline

MCMA \citep{liang2026learning} &
Meta-cognitive management; optimizes how the agent organizes and transfers structured knowledge about a user. &
Poor transfer of personalized knowledge between different functional domains or tasks. &
2026 \\

     \bottomrule
    \end{tabular}%
\caption{Optimization methods for personalization applications based on external memory. The publication years of these works are taken from their formal papers or technical reports and used for the statistics.}
\label{tb:per_rep_mem_app}
\end{table}%
\begin{table}[htbp]
\footnotesize
\setlength{\tabcolsep}{3pt}
  \centering
  \begin{tabular}{>{\raggedright\arraybackslash}p{1.5cm}>{\raggedright\arraybackslash}p{1cm}>{\raggedright\arraybackslash}p{5.1cm}>{\raggedright\arraybackslash}p{4.7cm}>{\raggedright\arraybackslash}p{1cm}}
    \toprule
    
\textbf{Method Name} &
\textbf{Type} &
\textbf{Key Features of the Method} &
\textbf{Applicable Application Scenarios}  &
\textbf{Year} \\\midrule

MemGPT \citep{packer2023memgpt} &
System &
Virtual memory paging; manages memory like an operating system to overcome context limits. &
Long-horizon tasks (coding, research) where context management is the primary bottleneck. &
2023 \\\hline

ReadAgent \citep{lee2024human} &
System &
Human-inspired reading mechanism; compresses long contexts into gists for high-efficiency reading. &
Long-document analysis and scenarios where context exceeds standard model limits. &
2024 \\\hline

HippoRAG \citep{jimenez2024hipporag} &
KG &
PageRank and spreading activation on KG; mimics neurobiological associative recall. &
Open-domain QA and tasks requiring complex multi-hop reasoning over large datasets. &
2024 \\\hline

MemoryBank \citep{zhong2024memorybank} &
Text/ Vector &
Long-term memory enhancement; hierarchical vector storage; recency-weighted retrieval. &
General-purpose persistence for long-term LLM chatbots. &
2024 \\\hline

AriGraph \citep{anokhin2024arigraph} &
KG &
Learns KG world models with episodic memory; builds a map of the environment for agent navigation. &
Embodied agents and gaming agents navigating complex or evolving worlds. &
2024 \\\hline

EMG-RAG \citep{wang2024craftingpersonalizedagentsretrievalaugmented} &
KG &
Graph-based retrieval and editing; supports dynamic updates to represent evolving world/user states. &
Personal agents that require a persistent, editable factual knowledge base. &
2024 \\\hline

MACLA \citep{forouzandeh2025learning} &
Hierar- chical &
Bayesian selection and contrastive refinement of hierarchical procedures for task execution. &
Complex task automation where agents must refine multi-step scripts over time. &
2025 \\\hline

ENGRAM \citep{patel2025engram} &
System &
Lightweight orchestration; focuses on low-overhead updates and effective recall for mobile agents. &
Conversational agents on mobile or resource-limited platforms. &
2025 \\\hline

GraphRAG \citep{edge2025localglobalgraphrag} &
KG &
Global summarization via community detection in graphs; identifies global themes from local data. &
Query-focused summarization over massive unstructured document collections. &
2025 \\\hline

Zep \citep{rasmussen2025zep} &
KG &
Temporal knowledge graph; tracks how facts and relationships change over time. &
Enterprise agents where temporal contexts are critical. &
2025 \\\hline

A-MEM \citep{xu2025mem} &
System &
Agent-centric memory management; provides a unified interface for agents to manage their own state. &
Multi-agent systems that need a standardized way to share and persist state information. &
2025 \\\hline

MemOS \citep{li2025memos} &
System &
Unified memory OS; manages heterogeneous storage and provides a centralized API for state. &
Large-scale AI agent deployments that require centralized state management. &
2025 \\\hline


LangMem \citep{langmem} &
Text &
Background extraction and prompt refinement; integrates with LangGraph for long-term state. &
Production agents requiring seamless persistent state and continuous learning. &
2025 \\\hline

Mem0 \citep{chhikara2025mem0} &
Text \& KG &
Universal memory layer; multi-level state (User, Session, Agent). &
Personalized AI assistants, customer support. &
2025 \\\hline

AMemGYM \citep{jiayangamemgym} &
System &
Interactive platform for benchmarking memory management in long-horizon dialogues. &
Research and evaluation of memory-augmented agents in realistic conversational settings. &
2026 \\\hline

xMemory \citep{hu2026beyond} &
System &
Decouples retrieval from aggregation; uses multiple specialized indices to improve recall precision. &
Information retrieval in dense data environments where high precision is required. &
2026 \\\hline

PlugMem \citep{yangplugmem} &
System &
Task-agnostic plugin module; retrofits existing LLMs with memory without requiring fine-tuning. &
Adding persistence to existing pre-trained models across diverse domains. &
2026 \\\hline

AgeMem \citep{yu2026agentic} &
System &
Unified management of long-term and short-term memory; employs priority-based eviction and paging. &
Resource-constrained deployments requiring efficient management of persistent state. &
2026 \\\hline

SYNAPSE \citep{jiang2026synapse} &
KG &
Spreading activation on episodic-semantic graphs; links events to concepts for associative recall. &
Complex reasoning tasks requiring fusion of personal experience and general knowledge. &
2026 \\\hline

ProMem \citep{yang2026beyond} &
Text/ Structured &
Proactive extraction from live streams; identifies significant events for immediate indexing. &
Real-time conversational assistants and proactive monitoring systems. &
2026 \\\hline

MCMA \citep{liang2026learning} &
Meta- cognitive &
Meta-cognitive management; optimizes memory organization for cross-domain knowledge transfer. &
Advanced agents requiring self-reflective capabilities and transferable knowledge. &
2026 \\

      \bottomrule
      \end{tabular}
\caption{Memory infrastructure and frameworks supporting personalization. The publication years of these works are taken from their formal papers or technical reports and used for the statistics.}
\label{tb:per_rep_mem_infra}
\end{table}

While the methods summarized above primarily focus on memory construction and retrieval, recent work has increasingly addressed two additional dimensions of the memory lifecycle: \textbf{updating} and \textbf{forgetting}.

For updating, a shared trend is moving from append-only storage toward conflict-aware revision. Rather than accumulating all historical facts, these systems explicitly detect conflicts between new and existing information and apply structured strategies—such as explicit add, update, and delete primitives (e.g., Mem0 \citep{chhikara2025mem0}) or temporal invalidation (e.g., Zep \citep{rasmussen2025zep}). A common principle is separating raw interaction evidence from derived preference state, revising the latter while preserving the former for auditability.

For forgetting, a shared philosophy is favoring retrieval-time suppression over physical deletion, since aggressive eviction risks losing rarely accessed but critical facts (e.g., allergy information). Representative strategies include cognitive-science-inspired decay (e.g., MemoryBank's Ebbinghaus curve \citep{zhong2024memorybank}), multi-tier retention policies (e.g., ENGRAM's hierarchy \citep{patel2025engram}), and activation-based inhibition (e.g., SYNAPSE \citep{jiang2026synapse}). Together, these update and forgetting mechanisms reflect a broader shift from passive storage toward lifecycle-aware memory management.


\subsection{Safety Concerns in Personalized Representations}

While richer and more expressive personalized representations improve the ability of LLM systems to adapt to individual users, they also increase the amount of personal information processed and stored, thereby expanding the potential attack surface. From a representation perspective, these risks can be broadly understood as forms of \textbf{representation leakage}, which arises when user information encoded in representations is exposed, inferred, or entangled  beyond its intended use.

\textbf{Over-exposure} refers to the direct or indirect leakage of
sensitive user information through model outputs or analysis of internal representations. Personalized representations—such as structured profiles, preference embeddings, or textual personas—often encode fine-grained attributes that can act as quasi-identifiers, enabling adversaries to link outputs with external data sources and re-identify users \citep{PPILeakage,pii}.

\textbf{Attribute inference} occurs when latent representations
implicitly encode sensitive traits that are not explicitly disclosed.
Even in the absence of direct identifiers, attackers may infer personal attributes by analyzing model behavior or response patterns \citep{chen2024janus,carlini2021extracting}.

\textbf{Information persistence} arises when user-specific information remains embedded in representations or memory structures over time. Such persistence makes sensitive information difficult to isolate or remove, and may lead to unintended propagation across interactions, increasing the risk of leakage or adversarial exploitation \citep{memory26snail,dong2026memoryinjectionattacksllm}.

These risks arise from different representation forms and storage mechanisms. As summarized in Table \ref{tab:presentation_risk}, each personalized representation is associated with different, though often overlapping, attack surfaces and characteristic leakage patterns.

\begin{table*}[t]
\centering
\small
\setlength{\tabcolsep}{4pt}
\resizebox{\textwidth}{!}
{
\begin{tabular}{p{2.5cm} p{3.8cm} p{3.8cm} p{3.8cm}}
\toprule
\textbf{Personalized Representation} & \textbf{Storage Mechanism} & \textbf{Attack Surface} & \textbf{Characteristic Leakage Risks} \\
\midrule
Structured profile & User profile database & Record linkage/data extraction & Exposure and linkage of explicit user attributes \\
\hline
Textual persona & Persona description repository & Stylometric and semantic analysis & Identity and behavioral trait leakage \\
\hline
Preference-based & User interaction logs & Behavioral pattern analysis and preference inference & Sensitive attribute inference from inferred preferences \\
\hline
Memory-based & External memory store ($M$) & Memory retrieval and retention mechanisms & Information persistence and cross-session leakage \\
\bottomrule
\end{tabular}
}
\caption{Mapping between personalized representation types, storage mechanisms, associated attack surfaces, and their characteristic, often overlapping leakage risks \citep{PPILeakage,pii,chen2024janus,carlini2021extracting,memory26snail,dong2026memoryinjectionattacksllm}.}
\label{tab:presentation_risk}
\end{table*}

\section{Personalization Techniques in LLM pipeline and Associated Safety Risks}
\label{sec:pipeline}

Building on the various forms of personalized representations discussed above, personalization techniques center on how user-specific information is utilized, updated, and integrated into model behavior. These approaches cover a wide spectrum, including non-parametric methods, parametric methods, and hybrid strategies that combine both.

\subsection{Non-parametric Personalization}
\label{sec:non-parametric}
In this paper, non-parametric personalization refers to personalization methods that operationalize $U$ and $M$ without updating the backbone LLM parameters. User-specific information is typically incorporated at inference time by augmenting the model input with external context, retrieved knowledge, or lightweight personalized representations. In this setting, $U$ and $M$ are not primarily encoded in the backbone model weights, but are instead explicitly injected into prompts or auxiliary contextual representations.

\subsubsection{Prompting-based Personalization} 
\label{sec:prompting}

\begin{figure}[ht]
    \centering
    \resizebox{\columnwidth}{!}{
    \begin{forest}
      for tree={
        font=\small,
        draw,
        align=center,
        thick,
        fill=blue!5, 
        rounded corners=3pt,
        grow'=east,    
        child anchor=west,    
        parent anchor=east,       
        anchor=west,
        l sep=0.8cm,   
        s sep=0.3cm,                   
        edge path={
          \noexpand\path [draw, \forestoption{edge}] 
          (!u.parent anchor) -- +(0.5,0) |- (.child anchor)\forestoption{edge label};
        },
      }
      [Prompting-based Personalization, fill=red!20
        [Personalized Techniques, fill=violet!20
          [Hard Prompt
            [Untuned Summarizer \cite{wang2023cue,richardson2023integrating,liu2024once}, fill=green!20!gray!10]
            [Task-specific Tuned Summarizer \cite{li2024matryoshka}, fill=green!20!gray!10]
            [Structured Profiles \cite{wu2026personalizedsafetyllmsbenchmark,li2016personabasedneuralconversationmodel}, fill=green!20!gray!10]
          ]
          [Soft Prompt
            [Input Prefix \cite{doddapaneni2024user,hebert2024persoma,sayana2025beyond}, fill=green!20!gray!10]
            [Cross-attention Mechanism \cite{ning2025user,liu2023recap}, fill=green!20!gray!10]
            [Output Logit \cite{wu2021personalized}, fill=green!20!gray!10]
          ]
        ]
        [Safety Attacks, fill=violet!20
          [Membership Inference \& Linkage Attacks
            [Hard Prompt \cite{duan2023flocks,edemacu2025privacy}, fill=green!20!gray!10]
            [Soft Prompt \cite{duan2023flocks,wang2025efficient}, fill=green!20!gray!10]
          ]
        ]
        [Safety Defenses, fill=violet!20
          [Hard Prompt
            [Data Anonymization \cite{karavdic2025handling,kan2023protecting,chen2023hide,zhang2024cogenesis}, fill=green!20!gray!10]
            [Differential Privacy \cite{duan2023flocks}, fill=green!20!gray!10]
          ]
          [Soft Prompt
            [Knowledge Distillation \cite{wang2025efficient}, fill=green!20!gray!10]
            [Differential Privacy \cite{duan2023flocks}, fill=green!20!gray!10]
          ]
        ]
      ]
    \end{forest}
    }
    \caption{Overview of prompting-based personalization.}
    \label{prompting_overview}
\end{figure}
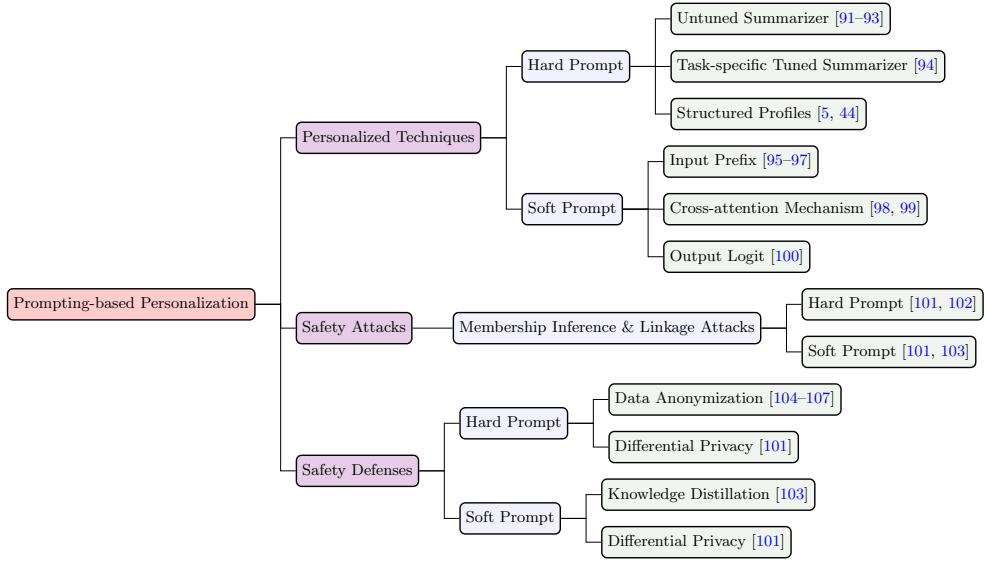

\paragraph{Personalized Prompting Techniques}
Prompting-based approaches integrate $U$ directly into the model input (see Figure \ref{prompting_overview} for an overview) \cite{liu2025survey}. One common strategy is profile-augmented prompting, where raw user data are summarized into natural language descriptions that encode user preferences and profiles to enrich the LLM context \cite{liu2025survey}. The summarization model may be either an untuned pre-trained LLM  \cite{wang2023cue,richardson2023integrating,liu2024once} or a task-specific fine-tuned model \cite{li2024matryoshka}. Structured user profiles or preference attributes can also be directly appended to the input context to guide generation \cite{wu2026personalizedsafetyllmsbenchmark,li2016personabasedneuralconversationmodel}. Another line of methods leverages soft prompting, which encodes personalized information into continuous soft embeddings rather than discrete prompt tokens \cite{liu2025survey}. These methods optimize lightweight personalized representations while keeping the backbone LLM frozen. Such representations can be incorporated through input prefixes \cite{doddapaneni2024user,hebert2024persoma,sayana2025beyond}, cross-attention mechanisms \cite{ning2025user,liu2023recap}, or inference-time logit modulation strategies \cite{wu2021personalized}.

Formally, given a query $x$, prompting-based personalization generates responses as $\mathrm{LLM}(x, U)$, where $U$ is explicitly injected into the model input or prompt representations. This design enables flexible and controllable personalization without updating the backbone LLM parameters.



\begin{figure}[h]
\centering
\includegraphics[width=\linewidth]{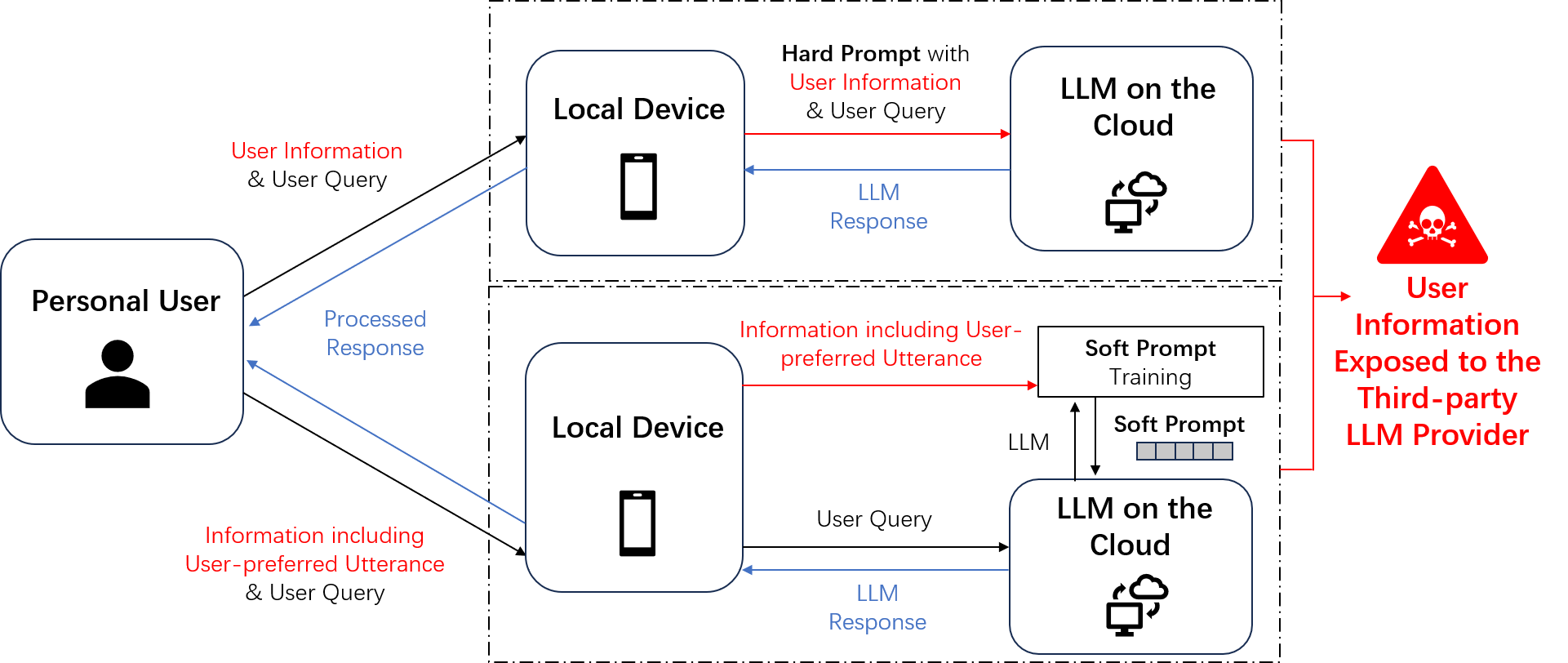}
\caption{User information exposed to the third-party LLM provider caused by hard and soft prompting-based personalization \cite{edemacu2025privacy, wang2025efficient}.}
\label{prompt-figure}
\end{figure}

\paragraph{Safety and Privacy Risks}
Nevertheless, user information embedded via hard (i.e., natural language-based) or soft prompts is vulnerable to membership inference attacks (which are leveraged to deduce whether a personal data instance was used in the hard prompt or the training of the soft prompt \cite{duan2023flocks}) through model outputs, particularly under adversarial querying or probing. This can further enable linkage attacks that infer user identities over time. Besides, as shown in Figure~\ref{prompt-figure}, prompting-based personalization embeds user representations in prompts, creating privacy risks as sensitive information becomes immediately accessible to the LLM and any third-party service during each interaction \cite{edemacu2025privacy}. This risk is further amplified in soft prompt-based methods, which often require sharing private data with the third-party LLM provider for soft prompt construction or tuning, leading to more exposure and reduced controllability \citep{wang2025efficient}.



\paragraph{Defense and Mitigation Strategies}

Various solutions have been proposed to address the safety risks caused by prompting-based personalization methods. For instance, to avoid sharing hard prompts containing personal data with third-party LLM providers, some studies adopt a privacy-preserving pipeline \cite{karavdic2025handling,kan2023protecting,chen2023hide,zhang2024cogenesis}: hard prompts are first processed by an anonymization module to ensure that personal information is not transmitted to the LLM, the anonymized data is then sent to the third-party model, and the resulting LLM output subsequently undergoes de-anonymization—which may be model-based or rule-based—to restore a personalized response. For soft prompt-based methods, to eliminate the need for sending private data to an external LLM provider for fine-tuning the soft prompt, the POST (Privacy Of Soft-prompt Transfer) technique \cite{wang2025efficient} employs knowledge distillation to derive a small language model from the external large LLM, and then tunes the soft prompt locally on the small model before transferring it back to the large LLM. In addition, prompt learning techniques based on Differential Privacy (DP) have been proposed to defend against membership inference and linkage attacks by introducing controlled perturbations into prompt representations. Examples include PromptPATE \cite{duan2023flocks}, which constructs an ensemble of LLMs using distinct hard prompts derived from private datasets, and PromptDPSGD \cite{duan2023flocks}, which performs private gradient descent (via gradient clipping and noise injection) during soft prompt training. While such DP–based approaches help strengthen privacy protection, they often degrade personalization performance due to injected noise, leaving this problem an open challenge.

\subsubsection{Retrieval-augmented Personalization}
\label{sec:retrieval}
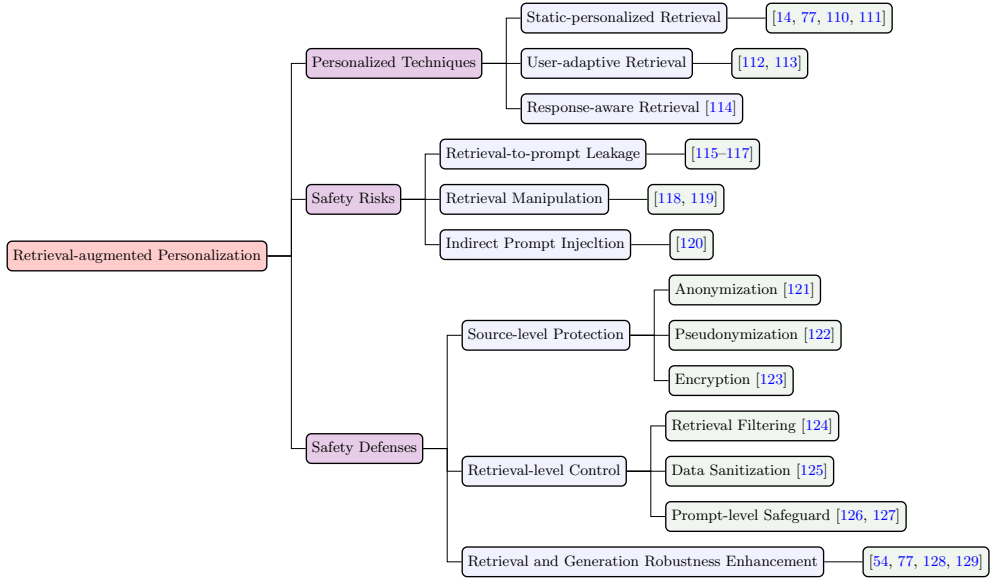
\begin{figure}[ht]
    \centering
    \resizebox{\columnwidth}{!}{
    \begin{forest}
      for tree={
        font=\small,
        draw,
        align=center,
        thick,
        fill=blue!5, 
        rounded corners=3pt,
        grow'=east,    
        child anchor=west,    
        parent anchor=east,       
        anchor=west,
        l sep=0.8cm,   
        s sep=0.3cm,                   
        edge path={
          \noexpand\path [draw, \forestoption{edge}] 
          (!u.parent anchor) -- +(0.5,0) |- (.child anchor)\forestoption{edge label};
        },
      }
      [Retrieval-augmented Personalization, fill=red!20
        [Personalized Techniques, fill=violet!20
         [
          Static-personalized Retrieval
          [\citep{wang2024craftingpersonalizedagentsretrievalaugmented, EnronQA-personalized-rag-2025-private,zhang-etal-2024-llm-based,salemi2024lamplargelanguagemodels}, fill=green!20!gray!10]
         ]
         [User-adaptive Retrieval
          [\citep{zerhoudi2026personaragenhancingretrievalaugmentedgeneration,Zhang2025PersonalizeBR}, fill=green!20!gray!10
          ]
         ]
         [
          Response-aware Retrieval \citep{mo2025adaptivepersonalizedconversationalinformation}
         ]
        ]
        [Safety Risks, fill=violet!20
          [Retrieval-to-prompt Leakage
          [\citep{zeng2024goodbadexploringprivacy, 
          jiang2025feedbackguidedextractionknowledgebase, duan2024privacyriskincontextlearning}, fill=green!20!gray!10]]
          [Retrieval Manipulation
          [\citep{chen2024blackboxopinionmanipulationattacks, zhao2025-rag-safety-exploring-knowledge}, fill=green!20!gray!10]]
          [Indirect Prompt Injecltion
          [\citep{owasp-Fasha_2024}, fill=green!20!gray!10]]
        ]
        [Safety Defenses, fill=violet!20
          [
            Source-level Protection
            [
              Anonymization \citep{ning2025privacyprotectedretrievalaugmentedgenerationknowledge},fill=green!20!gray!10
            ]
            [
              Pseudonymization \citep{serenari2025semanticallyawarellmagentenhance},fill=green!20!gray!10
            ]
            [
              Encryption \citep{PEFT-10114537311203744595},fill=green!20!gray!10
            ]
          ]
          [
            Retrieval-level Control
            [
              Retrieval Filtering \citep{shen2026reliabilityrageffectiveprovablyrobust},fill=green!20!gray!10
            ]
            [
              Data Sanitization \citep{wu2025biasinjectionattacksrag}, fill=green!20!gray!10
            ]
            [
              Prompt-level Safeguard \citep{xu2025mixtureofinstructionsaligninglargelanguage,Yang2023ASL},fill=green!20!gray!10
            ]
          ]
          [
            Retrieval and Generation Robustness Enhancement
            [\citep{wang2024craftingpersonalizedagentsretrievalaugmented, Personalize-Before-Retrieve-zhang2025personalizeretrievellmbasedpersonalized,edge2025localglobalgraphrag, besta2025multiheadragsolvingmultiaspect}, fill=green!20!gray!10
            ]
          ]
        ]
      ]
    \end{forest}}
    \caption{Overview of retrieval-augmented personalization.}
    \label{img_rag_overview}
\end{figure}

As illustrated in Figure \ref{img_rag_overview}, we provide an overview of representative retrieval-augmented personalization techniques, their associated safety and privacy risks, and existing mitigation strategies. 
\paragraph{Retrieval-Augmented Personalization Techniques}
Retrieval-Augmented Generation (RAG) enhances LLMs with external retrieval to incorporate additional context at inference time \citep{ rag-survey-zhao2024retrievalaugmentedgenerationaigeneratedcontent,rag-NEURIPS2020-6b493230}. Retrieval-Augmented Personalization (RAP) extends this paradigm by leveraging user-specific memory $M$ as a primary source of retrieved context, thereby enabling personalized generation conditioned on user profiles, historical interactions, private documents, and long-term behavioral records through retrieval \cite{wang2024craftingpersonalizedagentsretrievalaugmented, EnronQA-personalized-rag-2025-private}.

Early retrieval-based personalization methods primarily rely on static retrieval mechanisms, where user memory remains relatively fixed during inference. These approaches retrieve personalized context from predefined user profiles, historical interaction records, or external memory stores to augment generation. For example, \citet{salemi2024lamplargelanguagemodels} retrieve personalized profile items derived from users' historical behaviors and interactions to personalize language model outputs. Similarly, \citet{zhang-etal-2024-llm-based} retrieve relatively stable medical-related user information, such as historical medical records and long-term health conditions, to support personalized medical assistance.

\begin{figure}[h]
\centering
\includegraphics[width=0.8\linewidth, trim=0 1cm 0 1cm,
    clip]{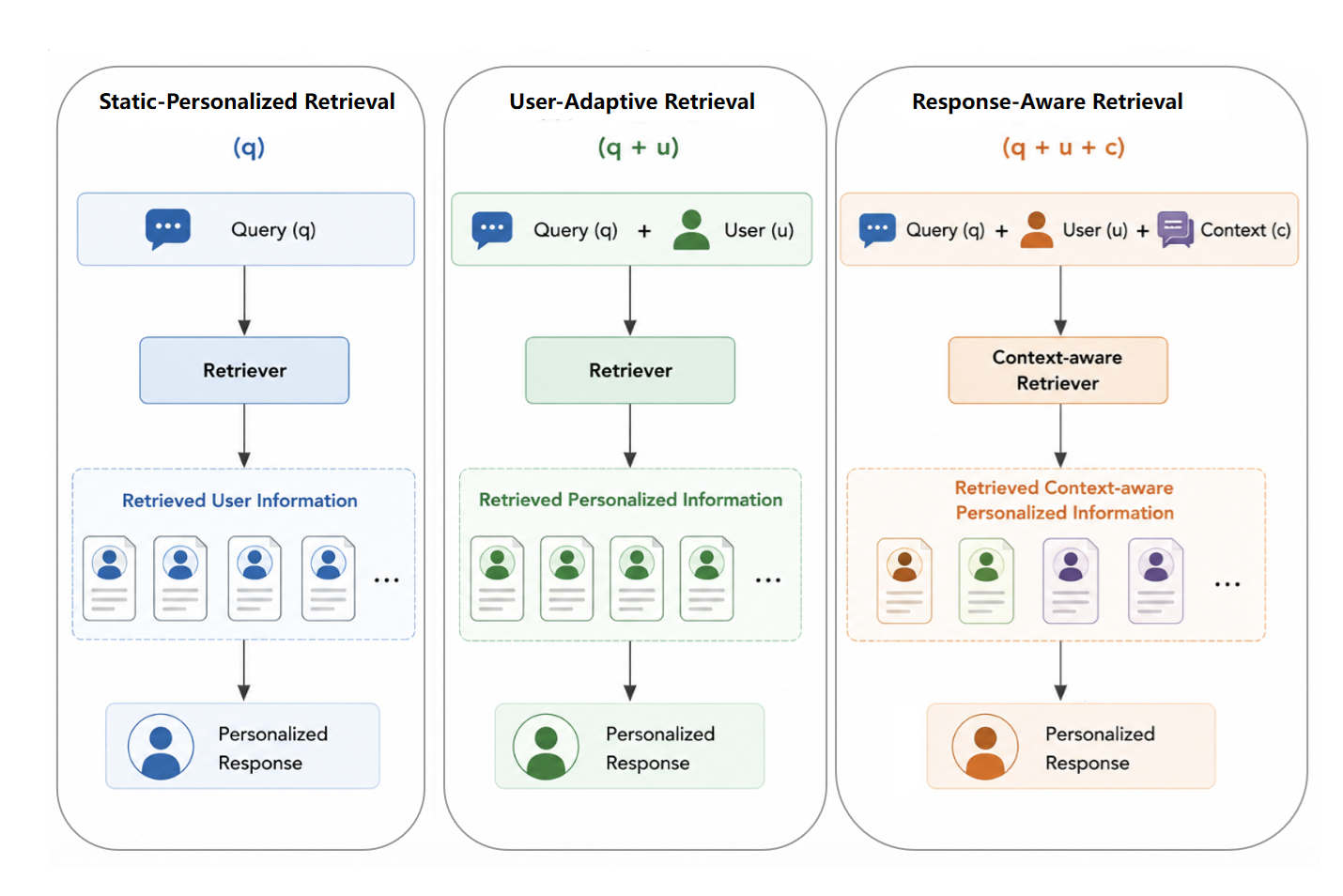}
\caption{Comparison of RAP paradigms. Early static methods retrieve personalized information using query-only retrieval from user memory or external knowledge bases \citep{wang2024craftingpersonalizedagentsretrievalaugmented, EnronQA-personalized-rag-2025-private}. User-adaptive retrieval further incorporates dynamic user signals, such as profiles and historical preferences, to better match evolving user needs \citep{zerhoudi2026personaragenhancingretrievalaugmentedgeneration,Zhang2025PersonalizeBR}. More recent response-aware methods additionally integrate conversational context to dynamically determine whether and how personalization should be applied during generation \citep{mo2025adaptivepersonalizedconversationalinformation}.}
\label{fig:ragmethods}
\end{figure}

More recent approaches increasingly move beyond static retrieval toward user-adaptive personalization through dynamically updated memory and adaptive retrieval strategies. For instance, \citet{zerhoudi2026personaragenhancingretrievalaugmentedgeneration} incorporate user-centric agents that continuously adapt retrieval and generation based on real-time user interactions and evolving personal data. \citet{Zhang2025PersonalizeBR} further propose a personalized query expansion framework that injects user-specific semantics and historical interaction signals before retrieval. Beyond fixed retrieval policies, \citet{mo2025adaptivepersonalizedconversationalinformation} dynamically adjust retrieval behavior according to the personalization requirements of different conversational queries. Collectively, these methods enable retrieval systems to continuously refine personalized context according to evolving user preferences, interaction patterns, and long-term behavioral dynamics. Figure \ref{fig:ragmethods} illustrates the evolution of RAP paradigms, progressing from static query-driven retrieval to response-aware personalization.

\begin{figure}[h]
\centering
\includegraphics[width=\linewidth, trim=0 6cm 0 1cm,
    clip]{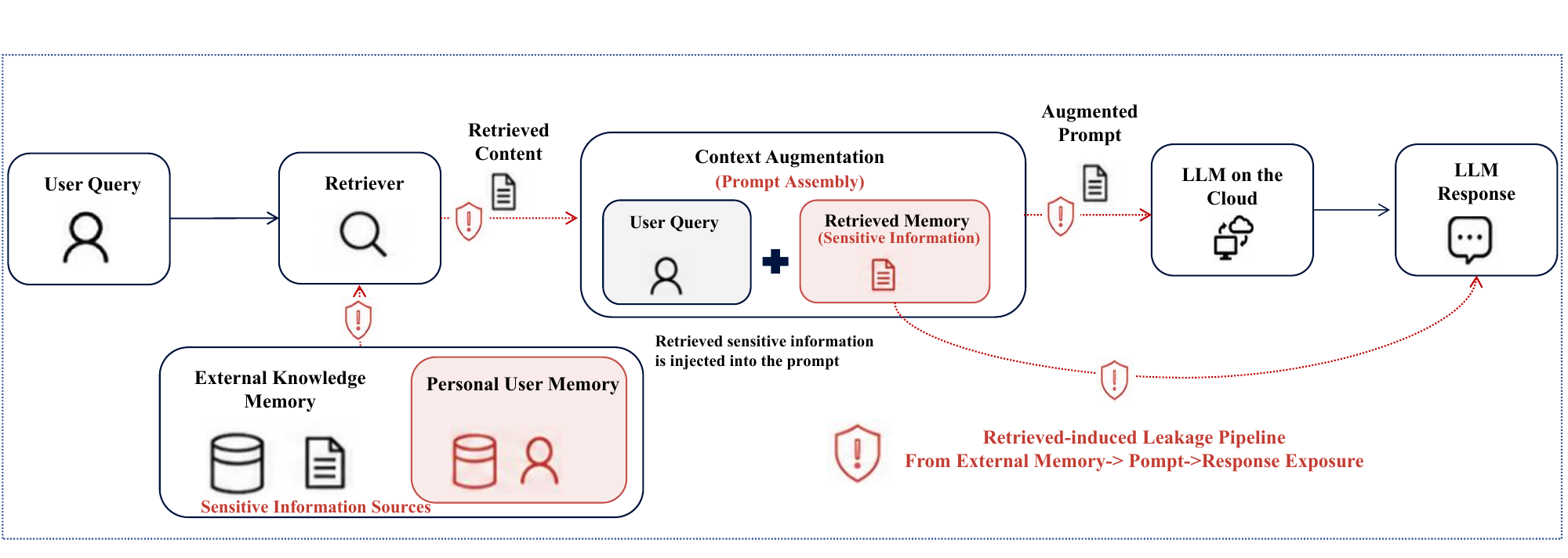}
\caption{Potential leakage pathway in retrieval-augmented personalization through external memory retrieval \citep{zeng2024goodbadexploringprivacy, jiang2025feedbackguidedextractionknowledgebase, duan2024privacyriskincontextlearning}.}
\label{fig:augument}
\end{figure}

\paragraph{Safety and Privacy Risks}
From a safety perspective, retrieval-based personalization shares core risks with prompting-based methods due to the reliance on context injection, including potential information leakage and adversarial attacks \cite{rag-survey-zhao2024retrievalaugmentedgenerationaigeneratedcontent, rag-are-not-safer-2025-rag}. However, a key distinction lies in its explicit dependence on external retrieval pipelines, which introduces RAP-specific vulnerabilities by allowing untrusted data to directly influence model inputs. In particular, RAP systems are inherently exposed to data leakage risks along the retrieval-to-prompt pathway(shown in Figure \ref{fig:augument}), where sensitive information stored in external corpora or user memory can be retrieved and subsequently injected into prompts, leading to unintended exposure during generation \cite{zeng2024goodbadexploringprivacy, jiang2025feedbackguidedextractionknowledgebase, duan2024privacyriskincontextlearning}. Furthermore, this tight coupling between retrieval and generation creates unique integrity risks. Adversaries can manipulate retrieval sources or indexing mechanisms to control which evidence is selected, thereby steering downstream outputs without modifying the model itself \cite{chen2024blackboxopinionmanipulationattacks, zhao2025-rag-safety-exploring-knowledge}. In addition, indirect prompt injection attacks exploit the fact that retrieved documents are treated as trusted context, embedding malicious instructions that are propagated through the retrieval pipeline into the prompt and followed by the model, resulting in unintended or harmful behaviors \cite{owasp-Fasha_2024}.

\paragraph{Defense and Mitigation Strategies} 
Meanwhile, many studies are dedicated to mitigating the risks associated with RAP \cite{securing-rag-Ammann_2025}. To reduce the exposure of sensitive information from retrieval to prompt construction at the source level, prior work focuses on minimizing the presence of identifiable data in retrieved content. Data anonymization removes personal identifiers at the source \citep{ning2025privacyprotectedretrievalaugmentedgenerationknowledge}, while pseudonymization replaces sensitive information with pseudonyms in a reversible process \citep {serenari2025semanticallyawarellmagentenhance}. More advanced approaches adopt privacy-enhancing architectures, such as federated or encrypted retrieval, to limit direct access to raw user data \citep{PEFT-10114537311203744595}.

Ensuring the integrity of retrieved content with retrieval-level control is another key direction.
Retrieval filtering and data sanitization are commonly used to detect
and remove malicious or irrelevant documents before they are incorporated into the prompt \cite{shen2026reliabilityrageffectiveprovablyrobust,wu2025biasinjectionattacksrag}, thereby reducing the risk of adversarial manipulation. In addition, since retrieved content is ultimately injected into the model input, prompt-level safeguards play an important role. Techniques
such as system instruction reinforcement constrain how retrieved information is interpreted and used \cite{xu2025mixtureofinstructionsaligninglargelanguage}, while prompt validation helps detect and reject malicious instructions embedded in queries or retrieved documents \citep{Yang2023ASL}.

Beyond these explicit controls, recent work improves RAG safety by enhancing the robustness of retrieval and generation. EMG-RAG \citep{wang2024craftingpersonalizedagentsretrievalaugmented} introduces editable memory structures to better regulate how user-specific information is retrieved and utilized, while unified representation
learning improves consistency across retrieval and generation, reducing error propagation \citep{Personalize-Before-Retrieve-zhang2025personalizeretrievellmbasedpersonalized}. In addition, LLM-based approaches leverage the model's own reasoning capabilities to assess the reliability of retrieved content and correct inconsistencies during generation \citep{edge2025localglobalgraphrag,besta2025multiheadragsolvingmultiaspect}, providing an additional layer of defense against both retrieval manipulation and prompt injection.

\subsection{Parametric Personalization}
\label{sec:parametric}
Parametric personalization incorporates user-specific information into internal model parameters through training or adaptation. Rather than injecting $U$ or $M$ at inference time, these methods encode personalization information into model weights, yielding a personalized model:

\begin{equation}
P(y \mid x, \theta_u),
\quad \text{where } 
\theta_u = \theta + \Delta \theta_u(U, M)
\end{equation}

\noindent where $\theta$ denotes the base model parameters, while $\Delta \theta_u$ represents user-specific parameter adaptations learned from personalization information $U$ and $M$.

%


This enables persistent and potentially more efficient personalization, but also fundamentally changes the security boundary: user-specific information becomes entangled with model weights, making it harder to isolate, control,or remove.

\subsubsection{Fine-tuning-based Personalization}
\label{sec:sft}
\begin{figure}[ht]
    \centering
    \resizebox{\columnwidth}{!}{
    \begin{forest}
      for tree={
        font=\small,
        draw,
        align=center,
        thick,
        fill=blue!5, 
        rounded corners=3pt,
        grow'=east,    
        child anchor=west,    
        parent anchor=east,       
        anchor=west,
        l sep=0.8cm,   
        s sep=0.3cm,                   
        edge path={
          \noexpand\path [draw, \forestoption{edge}] 
          (!u.parent anchor) -- +(0.5,0) |- (.child anchor)\forestoption{edge label};
        },
      }
      [SFT-based Personalization, fill=red!20
        [Personalized Techniques, fill=violet!20
         [
         FFT-based
          [\citep{gao2023retrieval,du2022retrieval,yang2023palr}, fill=green!20!gray!10]
         ]
         [PEFT-based
          [LoRA \citep{zhang2024personalized,wozniak2024personalized}, fill=green!20!gray!10
          ]
          [Prompt Tuning \citep{wu2024personalized2,huang2024selective,hebert2024persoma}, fill=green!20!gray!10
          ]
          [Prefix Tuning \citep{huber2025embedding}, fill=green!20!gray!10
          ]
         ]
        ]
        [Safety Attacks, fill=violet!20
          [Membership Inference Attacks
          [\citep{ran2025lora,du2025privacy,liu2024precurious}, fill=green!20!gray!10]]
          [Backdoor Attacks
          [\citep{wen2023last,wen2024privacy,yan2024backdooring,zeng2024uncertainty}, fill=green!20!gray!10]]
          [Alignment Drift \& Bias Amplification
          [\citep{taraghi2025efficiency}, fill=green!20!gray!10]]
        ]
        [Safety Defenses, fill=violet!20
          [Privacy-preserving Training
          [Differential Privacy \citep{ma2024efficient,sun2024improving,utpala2023locally}, fill=green!20!gray!10]]
          [Backdoor Defenses
          [Data-based \citep{zhao2024defending},fill=green!20!gray!10] 
          [Model-based \citep{kim2025obliviate},  fill=green!20!gray!10]]
          [Alignment-constrained Fine-tuning
          [Safety Anchoring \citep{yang2025asft},fill=green!20!gray!10] 
          [Constraint-based SFT \citep{huang2024lisa} ,fill=green!20!gray!10]
          ]
        ]
      ]
    \end{forest}}
    \caption{Overview of SFT-based personalization.}
    \label{img_overview}
\end{figure}
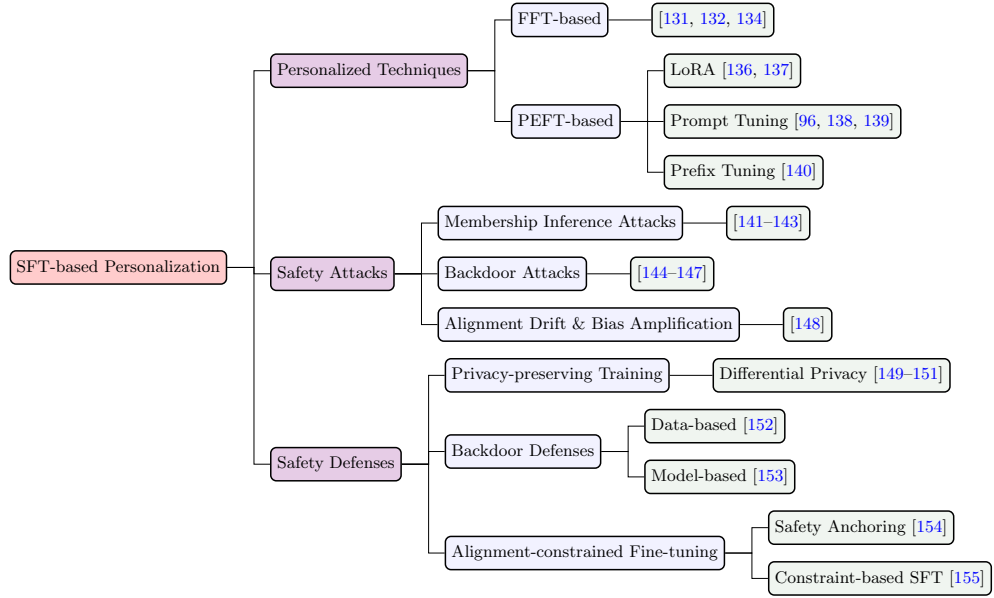

Figure \ref{img_overview} depicts a comprehensive overview of the typical fine-tuning-based personalization landscape. This section is structured around three dimensions: technological overviews, safety and privacy risks, and defense and mitigation strategies.

\paragraph{Fine-tuning-based Personalization Techniques}
Supervised Fine-Tuning (SFT) is the fundamental process of adapting a pre-trained foundation model to align with individual user preferences, behaviors, or domain-specific knowledge. Historically, this was primarily achieved through Full-parameter Fine-Tuning (FFT) \citep{zhang2024personalization}. For example, \citep{du2022retrieval,yang2023palr} allow LLMs to adapt to specific data formats and generate responses in a particular style, which is crucial for many personalization tasks, outperforming on some personalization tasks than zero-shot or few-shot prompting off-the-shelf LLMs \citep{kang2023llms,gao2023retrieval}. However, maintaining a full-parameter copy per user is computationally and storage-prohibitive. More critically, when user data is scarce, FFT-induced overfitting and catastrophic forgetting can directly erode the model's generalized reasoning and foundational safety guardrails \citep{luo2025empirical}.

To achieve scalable and sustainable personalization, recent research has shifted toward Parameter-Efficient Fine-Tuning (PEFT) methods, such as LoRA \citep{hu2022lora}, Prompt Tuning \citep{lester2021power}, and Prefix-Tuning \citep{li2021prefix}, restricting updates to a tiny fraction of parameters while freezing the core backbone. This paradigm enables lightweight, modular personalization by deploying user-specific parameter patches onto a shared foundation model, substantially reducing memory and deployment costs for on-device adaptation. For instance, PLoRA \citep{zhang2024personalized} combines personalized knowledge injection and LoRA, injecting all user attributes/preferences through a unified embedding. PER-PCS \citep{tan2024personalized} incorporates a gating module that selects the appropriate LoRA adapter, enabling more fine-grained personalization. E2P \citep{huber2025embedding} proposes Embedding-to-Prefix, injecting pre-computed context embeddings into an LLM's hidden representation space through a learned projection to a single soft token prefix. 

\begin{figure}[ht]
\centering
\includegraphics[height=10.8cm,width=\columnwidth]{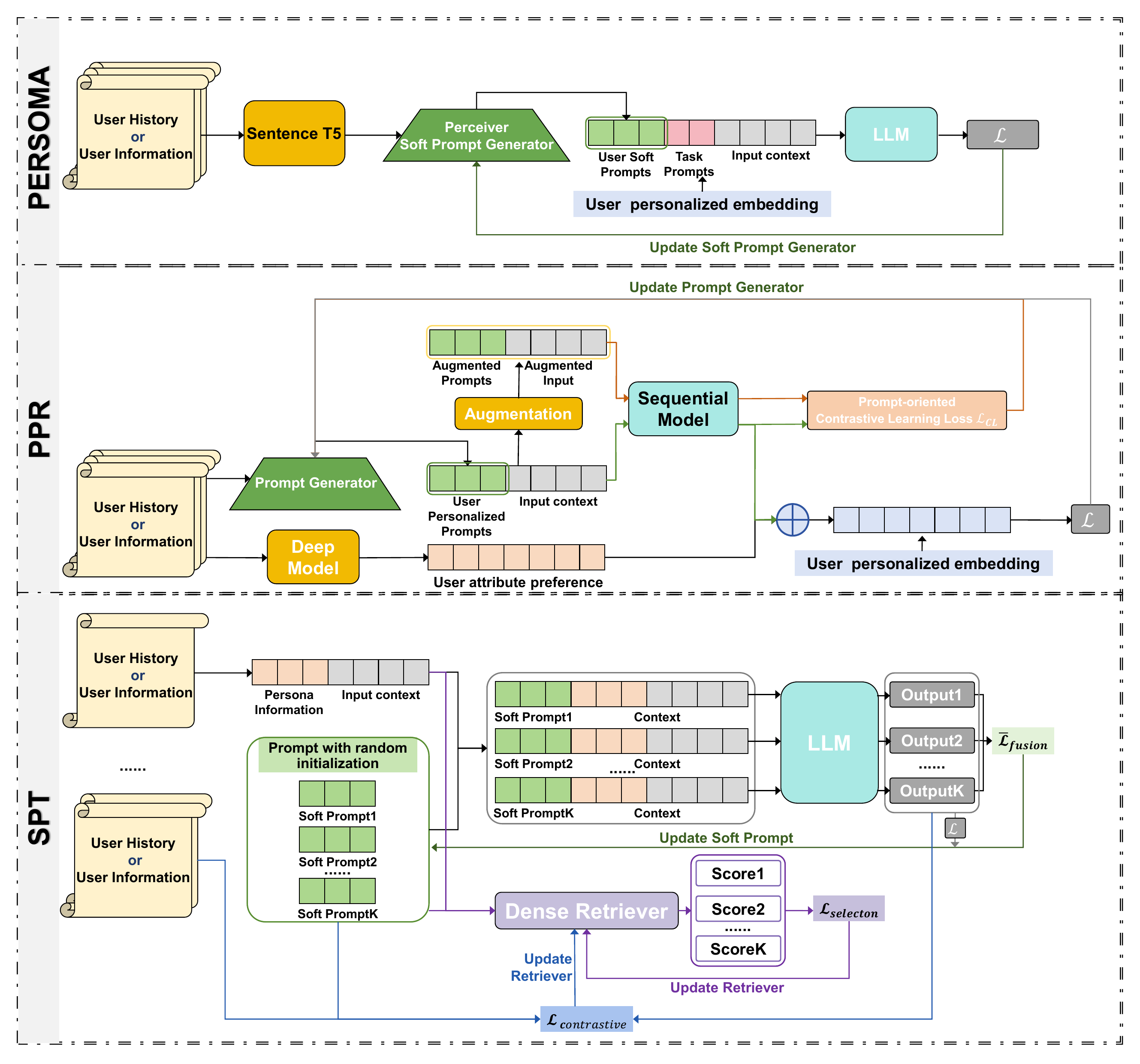}
\caption{Architectural differences in Prompt Tuning strategies among Persoma \citep{hebert2024persoma}, PPR \citep{wu2024personalized2}, and SPT \citep{huang2024selective}. While Persoma relies on an adapter-based derivation for task concatenation, PPR focuses on profile-driven generation with contrastive training, and SPT shifts to a retrieval-and-fusion paradigm from an initialized prompt pool.}
\label{img_arch}
\end{figure}

To provide a more detailed perspective on the application of PEFT in personalization, we take Prompt Tuning as a representative case to illustrate the overall architectures of various Prompt Tuning strategies. As illustrated in Figure \ref{img_arch}, Persoma \citep{hebert2024persoma} utilizes a Perceiver adapter to create in-context user soft prompts, which are then concatenated with task embeddings. PPR \citep{wu2024personalized2} builds a personalized soft prompt via a prompt generator based on user profiles, and enables sufficient training on prompts via a new prompt-oriented contrastive learning. SPT \citep{huang2024selective} initializes a set of soft prompts and uses a trainable dense retriever to adaptively select suitable soft prompts with prompt contrastive learning and fusion learning.

\paragraph{Safety and Privacy Risks}
Despite these architectural and computational advantages, the transition to SFT-based personalization, particularly through PEFT, introduces new safety and ethical concerns \citep{qi2023fine}. Recent studies have examined the mechanisms, vulnerabilities, and impacts of membership inference attacks on SFT-based paradigms, shedding light on various attack strategies and the factors that influence model susceptibility \citep{ran2025lora,du2025privacy,liu2024precurious}. Furthermore, some works have found that the modular nature of SFT opens the door for backdoor attacks, as malicious actors can distribute poisoned adapters that exhibit benign behavior on standard inputs but trigger harmful outputs under specific conditions \citep{wen2023last,wen2024privacy,yan2024backdooring,zeng2024uncertainty}. Finally, SFT-based methods may potentially exacerbate alignment drift and bias amplification, where models may overfit to undesirable biases present in the personal dataset, thereby drifting away from the foundational safety alignment \citep{taraghi2025efficiency}.


\paragraph{Defense and Mitigation Strategies} 
To address the aforementioned challenges, various defense mechanisms have been proposed to improve the safety of SFT-based personalization. One major direction focuses on Privacy-Preserving Fine-Tuning, particularly through DP-enhanced optimization and gradient perturbation. Such methods inject calibrated noise into gradient updates, providing mathematically grounded privacy guarantees that reduce the risk of memorizing and leaking sensitive personal information during generation \citep{ma2024efficient,sun2024improving,utpala2023locally}. Furthermore, researchers have focused on mitigating backdoor attacks by identifying triggers in poisoned samples to ensure the model either responds normally or rejects malicious inputs \citep{zhao2024defending}, or by altering the model's parameter structure to maximize the elimination of embedded vulnerabilities \citep{kim2025obliviate}. To mitigate the risks of alignment drift and bias amplification, researchers have introduced safety anchoring and constraint-based SFT. These methods constrain the model's update direction to ensure that the personalized model remains anchored to the original foundation model's safety guardrails, thereby preventing it from straying into harmful or biased output spaces \citep{yang2025asft,huang2024lisa}. 

\subsubsection{Reinforcement Learning-based Personalization}
\label{sec:rll}

Reinforcement Learning from Human Feedback (RLHF) has become a widely adopted paradigm for aligning large language models (LLMs) with human preferences \citep{christiano2017deep,ouyang2022training}. In the standard RLHF pipeline, human preference comparisons are used to train a reward model, which then guides policy optimization through reinforcement learning algorithms such as Proximal Policy Optimization (PPO) \citep{schulman2017proximal}. Recent alignment methods extend this framework by exploring different sources and formulations of preference signals, including learned reward models, AI feedback, direct preference optimization, and other preference-based objectives \citep{rafailov2023direct,lee2024rlaif}. Formally, given a prompt $x$ and a response $y$ generated by a policy $\pi_\theta(y \mid x)$, KL-regularized reward optimization can be broadly written as

\begin{equation}
\max_{\theta}
\mathbb{E}_{x \sim \mathcal{D},\, y \sim \pi_\theta(\cdot \mid x)}
\left[
r_\phi(x,y)
-
\beta
\log
\frac{\pi_\theta(y \mid x)}
{\pi_{\mathrm{ref}}(y \mid x)}
\right]
\end{equation}

\noindent where $r_\phi(x,y)$ represents the reward or preference signal, $\pi_{\mathrm{ref}}$ is a reference policy, $\beta > 0$ controls the strength of KL regularization, and the KL term penalizes deviations from the reference policy.

However, most RL-based alignment methods are designed for global alignment rather than personalization. They typically aggregate feedback from diverse users into a single reward model or optimization objective, implicitly assuming a shared population-level preference function. Personalized RL instead treats cross-user variation in reward signals as evidence of heterogeneous values and preferences, shifting alignment from population-level reward maximization to user-specific utility optimization \citep{zbMATH07001693,MARCUCCI2012331}.

Under heterogeneous preferences, common Bradley--Terry--Luce (BTL)-style reward models may collapse multiple preference modes into an averaged objective, thereby marginalizing minority viewpoints and failing to capture individual needs. RL-based parametric personalization can therefore be viewed as adapting model parameters toward user-specific preference signals:
\begin{equation}
\pi_{\theta_u}(y \mid x),
\quad
\theta_u = \theta + \Delta\theta_u^{\mathrm{RL}}(U,M)
\end{equation}

\noindent where $\Delta\theta_u^{\mathrm{RL}}$ denotes parameter updates induced by user-specific feedback, preferences, or interaction history. A corresponding user-specific objective can be written as:

\begin{equation}
\max_{\theta_u}
\mathbb{E}_{x \sim \mathcal{D}_u,\,
y \sim \pi_{\theta_u}(\cdot \mid x)}
\left[
r_u(x,y)
-
\beta
\log
\frac{
\pi_{\theta_u}(y \mid x)
}{
\pi_{\mathrm{ref}}(y \mid x)
}
\right]
\end{equation}

\noindent where $\mathcal{D}_u$ denotes user-specific prompts or interaction data, and $r_u(x,y)$ captures user $u$'s preferences. Such rewards may be instantiated by adapted reward models, user-conditioned reward estimators, or factorized reward representations \citep{li2024personalized,shenfeld2025language}.

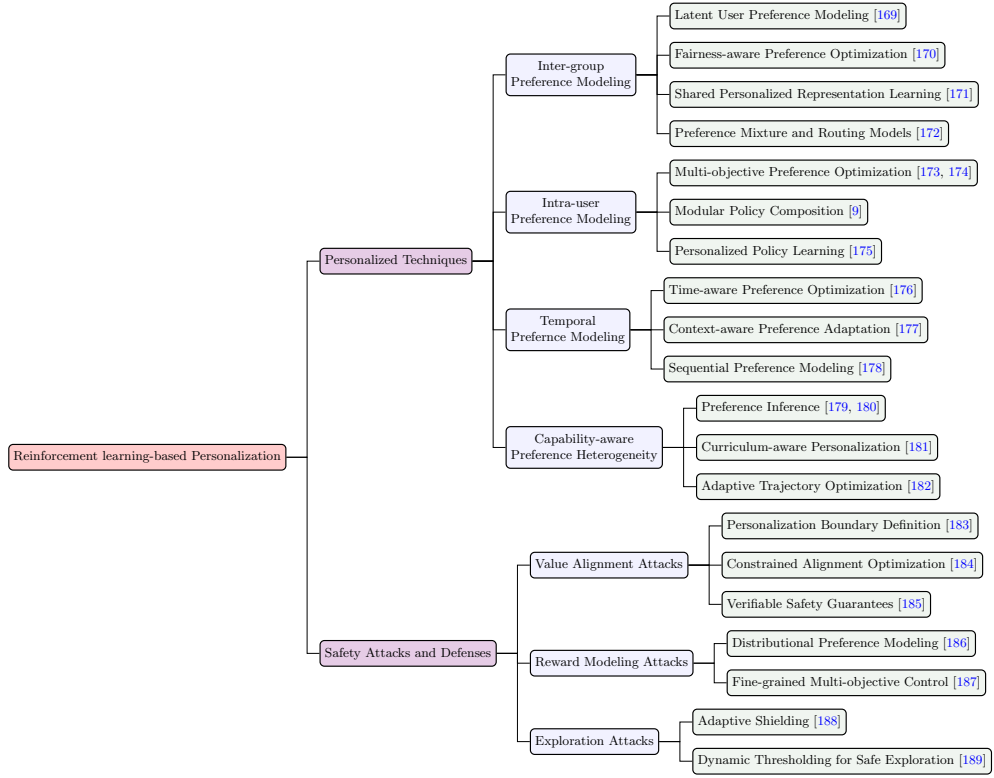
\begin{figure}[ht]
    \centering
    \resizebox{\columnwidth}{!}{
    \begin{forest}
      for tree={
        font=\small,
        draw,
        align=center,
        thick,
        fill=blue!5, 
        rounded corners=3pt,
        grow'=east,    
        child anchor=west,    
        parent anchor=east,       
        anchor=west,
        l sep=0.8cm,   
        s sep=0.3cm,                   
        edge path={
          \noexpand\path [draw, \forestoption{edge}] 
          (!u.parent anchor) -- +(0.5,0) |- (.child anchor)\forestoption{edge label};
        },
      }
      [Reinforcement learning-based Personalization, fill=red!20
        [Personalized Techniques, fill=violet!20
          [
            Inter-group \\ Preference Modeling
             [Latent User Preference Modeling \citep{NEURIPS2024_5e1c2556}, fill=green!20!gray!10
             ]
             [Fairness-aware Preference Optimization \citep{2024arXiv240208925C}, fill=green!20!gray!10
             ]
             [
              Shared Personalized Representation Learning \citep{mukherjee2025sharedreprlhfsharedrepresentationapproach},fill=green!20!gray!10
             ]
             [
              Preference Mixture and Routing Models \citep{shen2025micromixturemodelingcontextaware},fill=green!20!gray!10
             ]
          ]
          [ 
            Intra-user\\ Preference  Modeling
            [
              Multi-objective Preference Optimization \citep{bahlous-boldi2025pareto,zhou2024onepreferencefitsallalignmentmultiobjectivedirect},fill=green!20!gray!10
            ]
            [Modular Policy Composition \citep{jang2023personalizedsoupspersonalizedlarge}, fill=green!20!gray!10
            ]
            [Personalized Policy Learning \citep{2025arXiv250509496M}, fill=green!20!gray!10
            ]
          ]
          [
            Temporal\\ Prefernce Modeling
            [Time-aware Preference Optimization \citep{2024arXiv240718676S}, fill=green!20!gray!10
            ]
            [Context-aware Preference Adaptation \citep{2024Personalized}, fill=green!20!gray!10
            ]
            [Sequential Preference Modeling \citep{HWANG2025101527}, fill=green!20!gray!10
            ]
          ]
          [
            Capability-aware \\Preference Heterogeneity
            [Preference Inference \citep{Sweller2011, bajgar2025apprenticeship}, fill=green!20!gray!10
            ]
            [Curriculum-aware Personalization \citep{11082825}, fill=green!20!gray!10
            ]
            [Adaptive Trajectory Optimization \citep{11213850}, fill=green!20!gray!10
            ]
          ]
        ]
        [
          Safety Attacks and Defenses,fill=violet!20
          [
            Value Alignment Attacks,
            [
              Personalization Boundary Definition \citep{bounds},fill=green!20!gray!10
            ]
            [
              Constrained Alignment Optimization \citep{saferlhf},fill=green!20!gray!10
            ]
            [
              Verifiable Safety Guarantees \citep{chittepu2025reinforcement},fill=green!20!gray!10
            ]
          ]
          [
            Reward Modeling Attacks
            [
              Distributional Preference Modeling \citep{DPL},fill=green!20!gray!10
            ]
            [
              Fine-grained Multi-objective Control \citep{li2025ruleadapter},
              fill=green!20!gray!10
            ]
          ]
          [
            Exploration Attacks
            [
              Adaptive Shielding \citep{bethell2025safe},fill=green!20!gray!10
            ]
            [
               Dynamic Thresholding for Safe Exploration \citep{huang2024efficient_safe_rl},fill=green!20!gray!10
            ]
          ]
        ]
      ]
    \end{forest}}
    \caption{Overview of reinforcement learning-based personalization.}
    \label{prl_overview}
\end{figure}

Figure \ref{prl_overview} provides an organizing framework for RL-based personalization. The framework contains two connected branches. The technique branch categorizes RL-based personalization methods according to four sources of preference heterogeneity: inter-group, intra-user, temporal, and capability-aware heterogeneity. These sources determine whether personalization should model group-level value differences, individual trade-offs among multiple objectives, time-varying preferences, or capability-dependent user needs. The safety branch relates these heterogeneity sources to corresponding attacks and defenses, including value alignment attacks, reward modeling attacks, and exploration attacks. Following this structure, we first review the four technique categories and then discuss their safety implications and corresponding defenses.

\paragraph{Reinforcement Learning-based Personalization Techniques}

Following the technique branch of Figure \ref{prl_overview}, we review four representative categories of RL-based personalization methods and discuss how they differ in preference representation, optimization objective, and adaptation mechanism.

\textbf{Inter-group Preference Modeling} addresses group-level preference heterogeneity arising from differences in cultural background, social identity, or value systems. The central challenge is to reconcile structurally diverse preferences while avoiding the marginalization of minority groups in global alignment objectives. One line of work focuses on learning transferable user representations for group-aware personalization. For example, \citet{NEURIPS2024_5e1c2556} introduce latent user representations through variational inference to enable zero-shot personalization across heterogeneous user groups. Another direction incorporates fairness principles into personalized alignment. \citet{2024arXiv240208925C} integrate egalitarian objectives from social choice theory into RLHF by optimizing for the worst-off groups.To mitigate the tendency of MaxMin-style optimization to overemphasize extremely small minority populations, \citet{mukherjee2025sharedreprlhfsharedrepresentationapproach} propose a dual-layer architecture that combines shared representations for global generalization with personalized heads for group-specific adaptation. In parallel, \citet{shen2025micromixturemodelingcontextaware} employ preference mixture modeling and context-aware routing mechanisms to discover latent user preference clusters and dynamically adapt alignment across groups.

\textbf{Intra-user Preference Modeling} focuses on modeling trade-offs among multiple preference dimensions within an individual user, such as helpfulness, safety, conciseness, and humor. The central challenge lies in balancing potentially competing objectives while preserving alignment consistency and controllability. One line of work extends RLHF to multi-objective preference optimization. For example, the Multi-Objective RLHF framework introduces Projection Optimization (MOPO) to jointly optimize multiple objectives within a unified framework. \citet{bahlous-boldi2025pareto} and \citet{zhou2024onepreferencefitsallalignmentmultiobjectivedirect} adopt a two-stage paradigm consisting of preference aggregation followed by policy optimization, leveraging independently collected single-objective preference annotations to reduce annotation costs while enabling flexible objective trade-offs. Another direction decomposes user preferences into modular policy components. \citet{jang2023personalizedsoupspersonalizedlarge} train separate preference-specific policy modules and combine them through parameter interpolation at inference time, enabling dynamic control over personalized trade-offs. In addition, \citet{2025arXiv250509496M} propose a pessimistic personalized policy learning objective with regularization penalties to mitigate out-of-distribution action risks, thereby improving robustness and generalization under personalized preference shifts.

\textbf{Temporal Preference Modeling} addresses the temporal non-stationarity of user preferences, which may evolve over time due to changing contexts, personal growth, or shifts in social environments. Existing research in this area remains relatively limited but is gaining increasing attention. One line of work focuses on time-aware preference optimization. For example, \citet{2024arXiv240718676S} extend the Bradley–Terry model to non-stationary settings by introducing exponential time-weighting mechanisms for recency-sensitive preference optimization. Another direction explores context-aware adaptation for dynamically evolving preferences. \citet{2024Personalized} leverage context-aware preference learning to enable training-free online personalized adaptation under changing user contexts.In addition, sequential preference modeling approaches explicitly track the evolution of user preferences over time. Inspired by knowledge tracing, \citet{HWANG2025101527} adapt cognitive diagnostic frameworks to encode latent user preference states from historical interaction sequences, thereby modeling temporal preference dynamics.

\textbf{Capability-aware Preference Heterogeneity} captures systematic differences in how users with varying proficiency levels define preferred model behavior. For example, novice users often prefer instructional guidance and higher error tolerance, whereas expert users prioritize efficiency, precision, and depth \citep{Sweller2011}. Related forms of capability-aware personalization have been extensively explored in specialized domains. One line of work focuses on inferring personalized objectives from limited user feedback. For example, \citet{bajgar2025apprenticeship} combine inverse reinforcement learning with Bayesian active learning to infer personalized reward functions from minimal demonstrations. Another direction incorporates adaptive training strategies into personalized optimization. \citet{11082825} integrate curriculum learning into federated RL for adaptive content ranking, while \citet{11213850} employ hybrid evolutionary-RL algorithms to generate cognitively aware learning trajectories. Despite their maturity in domains such as intelligent tutoring and human-robot collaboration, these approaches remain underexplored in personalized LLM alignment, highlighting an important direction for future RLHF research.

\paragraph{Safety Attacks and Defenses}
The same heterogeneity sources that motivate RL-based personalization also introduce distinct safety attacks. As shown in the safety branch of Figure \ref{prl_overview}, these attacks can be grouped into value alignment attacks, reward modeling attacks, and exploration safety attacks. They are not independent: different forms of preference heterogeneity may interact and jointly shape the safety behavior of personalized policies.


\definecolor{myGreyBlue}{HTML}{2C3E50} 
\definecolor{mySkyBlue}{HTML}{458ED3} 
\definecolor{myOrage}{HTML}{DD841E} 
\definecolor{myGreen}{HTML}{4CAA78} 
\definecolor{myPurple}{HTML}{8C6AD9} 
\definecolor{myBluefill}{HTML}{EBF8FF}
\definecolor{myOrangefill}{HTML}{FEF3C7}
\definecolor{myPurplefill}{HTML}{FAF5FF}
\definecolor{myGreyfill}{HTML}{F7FAFC}
\begin{figure}[htbp]
\centering
\begin{tikzpicture}[scale=0.35, transform shape]
  
    \draw (550pt,685pt) node[font=\bfseries\LARGE, text=black!100] {\textbf{Safety Risk Types $\rightarrow$}};
    
    \draw (130pt,645pt) node[font=\bfseries\LARGE, text=black!100, align=center] {\textbf{Heterogeneity}};
    \draw (130pt,615pt) node[font=\bfseries\LARGE, text=black!100, align=center] {\textbf{Sources $\downarrow$}};
    
    \fill[rounded corners=8pt, fill=myGreyBlue] (280pt,670pt) rectangle (460pt,610pt);
    \draw (370pt,640pt) node[font=\bfseries\LARGE, text=white] {\textbf{Value Alignment}};
    
    \fill[rounded corners=8pt, fill=myGreyBlue] (480pt,670pt) rectangle (660pt,610pt);
    \draw (570pt,640pt) node[font=\bfseries\LARGE, text=white] {Reward Modeling};
    
    \fill[rounded corners=8pt, fill=myGreyBlue] (680pt,670pt) rectangle (860pt,610pt);
    \draw (770pt,640pt) node[font=\bfseries\LARGE, text=white] {Safe Exploration};
    
    \fill[rounded corners=10pt, fill=mySkyBlue] (40pt,580pt) rectangle (240pt,480pt);
    \draw (140pt,545pt) node[font=\bfseries\LARGE, text=white, align=center] {\textbf{Inter-group}};
    \draw (140pt,515pt) node[font=\bfseries\LARGE, text=white, align=center] {\textbf{Preferences}};
    
    \fill[rounded corners=10pt, fill=myOrage] (40pt,460pt) rectangle (240pt,360pt);
    \draw (140pt,425pt) node[font=\bfseries\LARGE, text=white, align=center] {\textbf{Intra-user}};
    \draw (140pt,395pt) node[font=\bfseries\LARGE, text=white, align=center] {\textbf{Trade-offs}};
    
    \fill[rounded corners=10pt, fill=myGreen] (40pt,340pt) rectangle (240pt,240pt);
    \draw (140pt,305pt) node[font=\bfseries\LARGE, text=white, align=center] {\textbf{Temporal}};
    \draw (140pt,275pt) node[font=\bfseries\LARGE, text=white, align=center] {\textbf{Preference Drift}};
    
    \fill[rounded corners=10pt, fill=myPurple] (40pt,220pt) rectangle (240pt,120pt);
    \draw (140pt,185pt) node[font=\bfseries\LARGE, text=white, align=center] {\textbf{Capability-aware}};
    \draw (140pt,155pt) node[font=\bfseries\LARGE, text=white, align=center] {\textbf{Preferences}};
    
    \fill[rounded corners=8pt, fill=myBluefill, draw=mySkyBlue, line width=1.5] (280pt,580pt) rectangle (460pt,480pt);
    \draw (370pt,535pt) node[font=\bfseries\fontsize{45}{54}\selectfont, text=black] {$\bullet$};
    \draw (370pt,515pt) node[font=\bfseries\Large, text=black] {\citep{bounds,saferlhf}};
    
    \fill[rounded corners=8pt, fill=myGreyfill, draw=gray!60, line width=1.5] (280pt,460pt) rectangle (460pt,360pt);
    \draw (370pt,405pt) node[font=\bfseries\fontsize{45}{54}\selectfont, text=black] {$-$};
    
    \fill[rounded corners=8pt, fill=myGreyfill, draw=gray!60, line width=1.5] (280pt,340pt) rectangle (460pt,240pt);
    \draw (370pt,285pt) node[font=\bfseries\fontsize{45}{54}\selectfont, text=black] {$-$};
    
    \fill[rounded corners=8pt, fill=myPurplefill, draw=myPurple, line width=1.5] (280pt,220pt) rectangle (460pt,120pt);
    \draw (370pt,175pt) node[font=\bfseries\fontsize{45}{54}\selectfont, text=black] {$\circ$};
    \draw (370pt,155pt) node[font=\bfseries\Large, text=black] {\citep{chittepu2025reinforcement}};
    
    \fill[rounded corners=8pt, fill=myBluefill, draw=mySkyBlue, line width=1.5] (480pt,580pt) rectangle (660pt,480pt);
    \draw (570pt,535pt) node[font=\bfseries\fontsize{45}{54}\selectfont, text=black] {$\circ$};
    \draw (570pt,515pt) node[font=\bfseries\Large, text=black] {\citep{DPL}};
    
    \fill[rounded corners=8pt, fill=myOrangefill, draw=myOrage, line width=1.5] (480pt,460pt) rectangle (660pt,360pt);
    \draw (570pt,415pt) node[font=\bfseries\fontsize{45}{54}\selectfont, text=black] {$\bullet$};
    \draw (570pt,395pt) node[font=\bfseries\Large, text=black] {\citep{DPL,li2025ruleadapter}};
    
    \fill[rounded corners=8pt, fill=myGreyfill, draw=gray!60, line width=1.5] (480pt,340pt) rectangle (660pt,240pt);
    \draw (570pt,285pt) node[font=\bfseries\fontsize{45}{54}\selectfont, text=black] {$-$};
    
    \fill[rounded corners=8pt, fill=myGreyfill, draw=gray!60, line width=1.5] (480pt,220pt) rectangle (660pt,120pt);
    \draw (570pt,165pt) node[font=\bfseries\fontsize{45}{54}\selectfont, text=black] {$-$};
    
    \fill[rounded corners=8pt, fill=myGreyfill, draw=gray!60, line width=1.5] (680pt,580pt) rectangle (860pt,480pt);
    \draw (770pt,525pt) node[font=\bfseries\fontsize{45}{54}\selectfont, text=black] {$-$};
    
    \fill[rounded corners=8pt, fill=myOrangefill, draw=myOrage, line width=1.5] (680pt,460pt) rectangle (860pt,360pt);
    \draw (770pt,415pt) node[font=\bfseries\fontsize{45}{54}\selectfont, text=black] {$\circ$};  
    \draw (770pt,395pt) node[font=\bfseries\Large, text=black] {\citep{huang2024efficient_safe_rl}};
    
    \fill[rounded corners=8pt, fill=myGreyfill, draw=gray!60, line width=1.5] (680pt,340pt) rectangle (860pt,240pt);
    \draw (770pt,285pt) node[font=\bfseries\fontsize{45}{54}\selectfont, text=black] {$-$};
    
    \fill[rounded corners=8pt, fill=myPurplefill, draw=myPurple, line width=1.5] (680pt,220pt) rectangle (860pt,120pt);
    \draw (770pt,175pt) node[font=\bfseries\fontsize{45}{54}\selectfont, text=black] {$\bullet$};
    \draw (770pt,155pt) node[font=\bfseries\Large, text=black] {\citep{bethell2025safe, huang2024efficient_safe_rl}};
    
    \fill[rounded corners=8pt, fill=myGreyfill, draw=gray!40, line width=2] (70pt,90pt) rectangle (850pt,30pt);

    \draw (100pt,60pt) node[font=\bfseries\LARGE, anchor=west] {Status:};

    \draw (190pt,60pt) node[font=\bfseries\LARGE, text=black, anchor=west] {$\bullet$};
    \draw (210pt,60pt) node[font=\LARGE, anchor=west] {primary association};

    \draw (410pt,60pt) node[font=\bfseries\LARGE, text=black, anchor=west] {$\circ$};
    \draw (430pt,60pt) node[font=\LARGE, anchor=west] {secondary influence};

    \draw (620pt,60pt) node[font=\bfseries\LARGE, text=black, anchor=west] {$-$};
    \draw (640pt,60pt) node[font=\LARGE, anchor=west] {relatively underexplored};
    
\end{tikzpicture}
\caption{Associations between preference heterogeneity and personalized RL safety attacks.}
\label{rll_safe}
\end{figure}
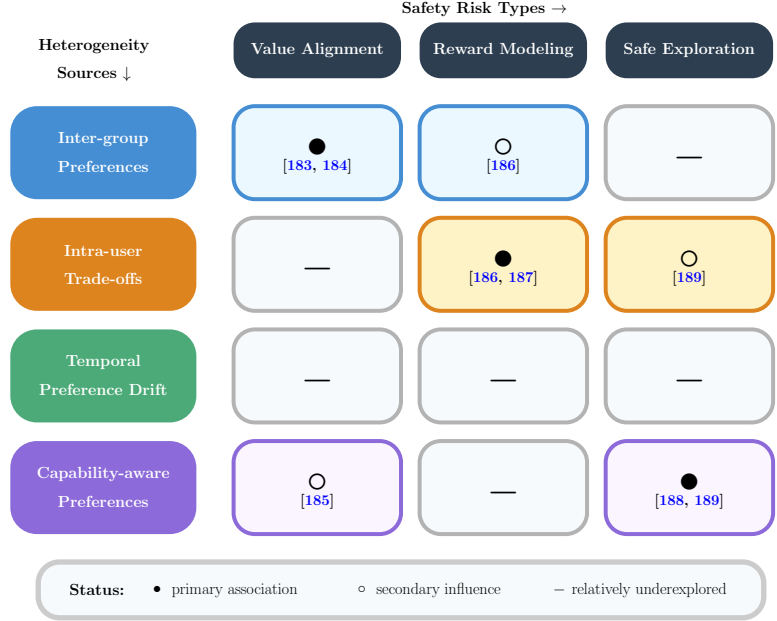

\textbf{Value Alignment Attacks} arise from the compression of heterogeneous group preferences induced by inter-group preference heterogeneity. Traditional RLHF's single reward function, based on a homogeneity assumption, algorithmically amplifies majority safety standards while marginalizing minority preferences and blurring safety boundaries. \citet{bounds} define acceptable personalization boundaries normatively; \citet{saferlhf} decouple helpfulness from harmlessness as a constrained optimization problem; \citet{chittepu2025reinforcement} provide verifiable guarantees via high-confidence constraints. Although existing work focuses on inter-group alignment,it often overlooks capability differences across individuals, with low-skill users requiring more conservative safety boundaries than high-skill users.

\textbf{Reward Modeling Attacks} arise mainly from intra-user heterogeneity, where multi-dimensional trade-offs are compressed into a single reward signal, leading to aggregation bias. Standard RLHF often relies on simplified aggregation (e.g., Borda-like mechanisms), which can underrepresent extreme preferences and enable strategic misalignment. \citet{DPL} address this via distributional preference modeling, while \citet{li2025ruleadapter} enable fine-grained multi-objective control through dynamic rule selection. While this issue is rooted in intra-user trade-offs, inter-group heterogeneity can further influence reward modeling by shaping the underlying preference distributions.

\textbf{Exploration Attacks} are closely tied to capability-aware preference heterogeneity, where mismatches between exploration strategies and user abilities lead to uneven risk exposure. Uniform safety constraints may over-restrict advanced users while under-protecting novices. \citet{bethell2025safe} introduce adaptive shielding to distinguish safe from unsafe actions, and \citet{huang2024efficient_safe_rl} employ dynamic thresholds to balance exploration and safety. This line of work primarily focuses on capability-aware exploration, with intra-user trade-offs also affecting risk sensitivity.



Figure \ref{rll_safe} highlights several relatively underexplored intersections in personalized RL safety. While each safety dimension has primarily been studied with respect to a corresponding heterogeneity source, the impact of \textbf{\textit{temporal preference heterogeneity}} remains relatively underexplored across all three dimensions. In addition, interactions between \textbf{\textit{intra-user preference heterogeneity}} and \textbf{\textit{value alignment risks}}, as well as between \textbf{\textit{capability-aware preference heterogeneity}} and \textbf{\textit{reward modeling risks}}, remain insufficiently studied. These gaps suggest that evolving user preferences over time and cross-dimensional interactions constitute important directions for future research, where both primary and secondary heterogeneity factors should be jointly considered to ensure robust and safe personalization.



\section{Architecture-level Personalization and Associated Safety Risks}
\label{sec:architecture}

Architecture-level personalization modifies the internal computation structure of models to facilitate user-specific adaptation at the architectural level. Representative approaches include Mixture-of-Experts (MoE)-based architectures and pruning-based methods.



\subsection{MOE-based Personalization} 
\label{sec:moe}

\begin{figure}[!t]
\centering
\includegraphics[width=\textwidth]{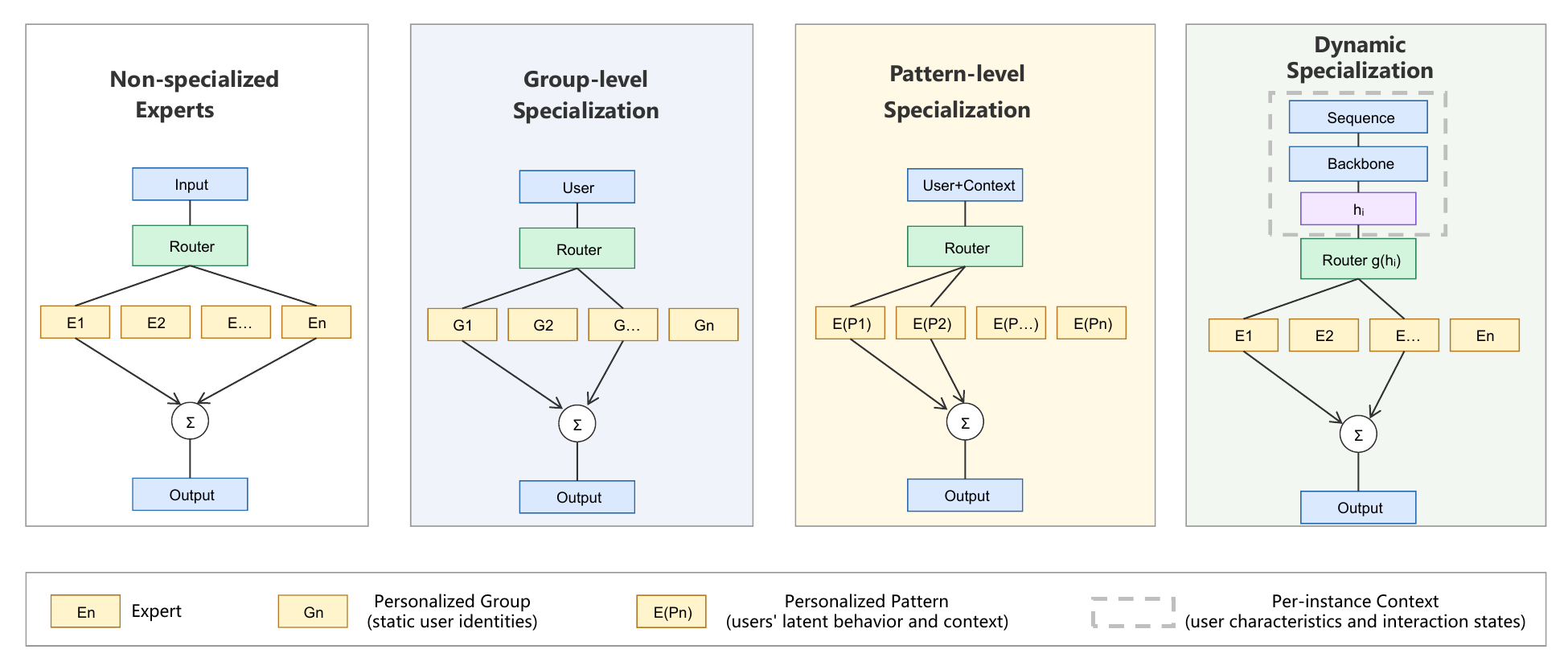}
\caption{Comparison of MoE-based personalization paradigms under different routing and expert specialization strategies.
Personalization evolves from shared experts \citep{dan2025p} with generic input routing, to group-level \citep{zhang2025proper} and pattern-level \citep{zhang2025hierarchical,liu2025mixture,tan2025moe,yang2026towards} expert allocation conditioned on user characteristics and contextual patterns, and finally to dynamic expert \citep{zhu2025tamer} specialization with state-dependent routing based on evolving hidden representations.}
\label{MoEarchitecture}
\end{figure}

\paragraph{MoE-based Personalization Techniques}
MoE architectures have recently been adopted for personalized LLMs due to their ability to enable efficient and selective adaptation \citep{sparse, switch,st-moe, glam, deepseekmoe}. Instead of uniformly updating all model parameters, personalized MoE architectures leverage sparse gating and routing mechanisms to dynamically activate user-specific experts, allowing different users to access distinct sub-networks conditioned on their characteristics within a shared model \citep{dan2025p}. This design supports scalable personalization while maintaining computational efficiency.

A central challenge in MoE-based personalization is how to allocate model capacity to capture diverse user characteristics under data sparsity and imbalance \citep{dan2025p,gong2023attention,kong2024customizing}. This challenge is closely related to the degree of expert specialization—namely, how experts are differentiated to model user-specific behaviors and preferences. Based on the granularity and adaptivity of specialization, existing approaches can be categorized into four paradigms: (1) non-specialized shared experts \citep{dan2025p}, (2) group-level specialization \citep{zhang2025proper}, (3) pattern-level specialization \citep{zhang2025hierarchical,liu2025mixture,tan2025moe,yang2026towards}, and (4) dynamic expert specialization \citep{zhu2025tamer}. These paradigms trace an evolution from uniform, shared computation toward progressively user-adaptive expert allocation, enabling model capacity to be more precisely personalized to individual characteristics and contexts, as illustrated in Figure \ref{MoEarchitecture}.

Early approaches such as P-React \citep{dan2025p}  instantiate the \textbf{non-specialized shared experts} paradigm,  adopting a LoRA-MoE architecture where experts encode multiple personality traits within a shared, undifferentiated representation space. This design maximizes parameter reuse by routing all users through a uniform expert pool; however, the absence of explicit specialization severely limits personalization fidelity. Under highly skewed user data distributions (e.g., the Pareto effect), such entangled modeling often leads to suboptimal personalization, as dominant user patterns tend to overshadow minority behaviors. Thus, this paradigm trades personalized granularity for computational uniformity, motivating subsequent methods to explicitly partition expert capacity along user or pattern boundaries.

Following this motivation, \textbf{group-level specialization} explicitly partitions expert capacity along user boundaries, dividing the population into coherent cohorts each served by a dedicated expert subset. PROPER \citep{zhang2025proper}, for instance, introduces group-level modeling with LoRA, where each expert is explicitly aligned with a subset of users, and a user-aware routing mechanism assigns users accordingly. By partitioning the user space, this design enables experts to focus on more coherent behavioral patterns, improving both scalability and robustness under data imbalance. Nevertheless, this gain in structure comes at the cost of personalization granularity: sharp group boundaries blur users at the margins into coarse, and the routing decision remains statically tied to predefined identities, leaving intra-group heterogeneity unresolved.

To resolve the intra-group heterogeneity that rigid group boundaries ignore, \textbf{pattern-level specialization} shifts expert assignment from static user identities to latent behavioral and contextual signals. Rather than binding experts to predefined groups,  AW-MoE \citep{gong2023attention} and iLoRA \citep{kong2024customizing} leverage diverse experts to capture heterogeneous behavioral patterns. HM4SR \citep{zhang2025hierarchical} further models interaction-level and temporal dynamics through hierarchical MoE structures. This pattern-level specialization paradigm generalizes naturally to other domains characterized by latent behavioral and contextual heterogeneity. In healthcare, \citet{huo2021sparse} and \citet{morrill2025let} enforce expert specialization through sparse gating and personalized event distributions, enabling experts to capture distinct patient risk trajectories and clinical progression patterns. Similarly, in location prediction, recent works \citep{liu2025mixture,tan2025moe} model heterogeneous mobility behaviors by capturing dependencies between long-term user roles and situational contextual familiarity, rather than relying solely on fixed user identities. In real-world clinical settings, Med-MoE-LoRA \citep{yang2026towards} further extends this paradigm to distribution-shifted environments by disentangling heterogeneous medical knowledge and dynamically allocating experts to different clinical contexts.

\textbf{Dynamic expert specialization} overcomes the static routing bottleneck of pattern-level methods by restructuring expert activation on a per-instance basis, recognizing that user characteristics and interaction states evolve too rapidly for fixed allocation mechanisms. For example, TAMER \citep{zhu2025tamer} incorporates online test-time adaptation to dynamically adjust expert routing during inference, enabling the model to continuously track evolving user or patient states. Such approaches extend personalization from static expert specialization toward continual and state-dependent adaptation, improving robustness under non-stationary user behaviors and distribution shifts; nevertheless, this flexibility hinges on the continuous availability of user signals during inference, a dependency that static, pre-trained paradigms do not require.

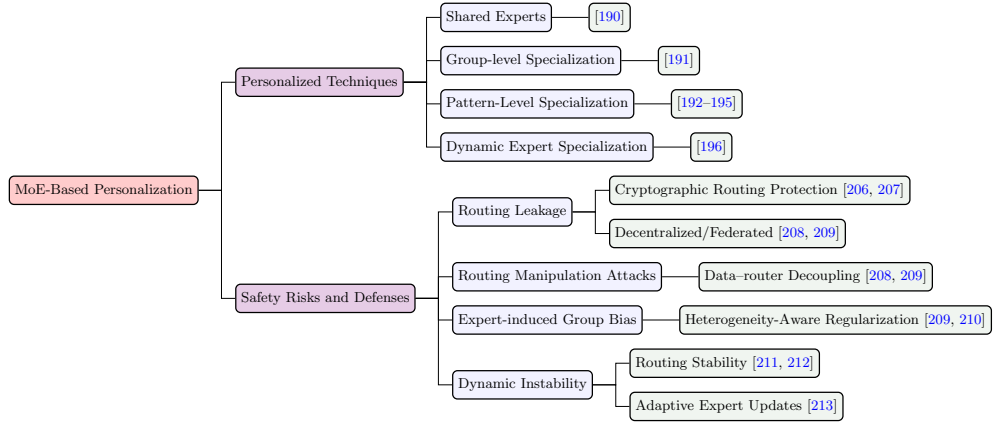
\begin{figure}[ht]
    \centering
    \resizebox{\columnwidth}{!}{
    \begin{forest}
      for tree={
        font=\small,
        draw,
        align=center,
        thick,
        fill=blue!5, 
        rounded corners=3pt,
        grow'=east,    
        child anchor=west,    
        parent anchor=east,       
        anchor=west,
        l sep=0.8cm,   
        s sep=0.3cm,                   
        edge path={
          \noexpand\path [draw, \forestoption{edge}] 
          (!u.parent anchor) -- +(0.5,0) |- (.child anchor)\forestoption{edge label};
        },
      }
      [MoE-Based Personalization, fill=red!20
        [Personalized Techniques,fill=violet!20
          [
            Shared Experts
            [\citep{dan2025p},fill=green!20!gray!10]
          ]
          [ 
            Group-level Specialization
            [\citep{zhang2025proper},fill=green!20!gray!10]
          ]
          [
            Pattern-Level Specialization
            [\citep{zhang2025hierarchical,liu2025mixture,tan2025moe,yang2026towards},fill=green!20!gray!10]
          ]
          [
            Dynamic Expert Specialization
            [\citep{zhu2025tamer},fill=green!20!gray!10]
          ]
        ]
        [
          Safety Risks and Defenses,fill=violet!20
          [
            Routing Leakage
            [
              Cryptographic Routing Protection \citep{shen2026secmoe, zhou2025cryptomoe},fill=green!20!gray!10
            ]
            [
              Decentralized/Federated \citep{guo2021pfl,yi2026pfedmoe},fill=green!20!gray!10
            ]
          ]
          [
            Routing Manipulation Attacks
            [
              Data–router Decoupling \citep{guo2021pfl,yi2026pfedmoe},fill=green!20!gray!10
            ]
          ]
          [
            Expert-induced Group Bias
            [
              Heterogeneity-Aware Regularization \citep{hu2025fft,yi2026pfedmoe},fill=green!20!gray!10
            ]
          ]
          [
            Dynamic Instability
            [
              Routing Stability \citep{cheng2025ermoeeigenreparameterizedmixtureofexpertsstable,delibasoglu2026spectralmanifoldregularizationstable},fill=green!20!gray!10
            ]
            [
              Adaptive Expert Updates \citep{li2025dynamicexpertspecializationcatastrophic},fill=green!20!gray!10
            ]
          ]
        ]
      ]
    \end{forest}}
    \caption{Overview of MoE-based personalization. (\textbf{i}) Personalization techniques range from shared experts to increasingly adaptive specialization (group-level, pattern-level, and dynamic). (\textbf{ii}) Safety risks and mitigations span routing leakage, manipulation attacks, group bias, and dynamic instability, with representative works at the leaf nodes.} 
    \label{moe_overview}
\end{figure}

\paragraph{Safety and Risks}
While MoE-based personalization techniques improve personalization efficacy, they also introduce several unique security challenges, as summarized below.

\textbf{Routing Leakage}: This refers to privacy leakage arising from unencrypted routing. In particular, routing indices constitute a critical side-channel vulnerability, as the data-dependent nature of expert selection enables adversaries to reconstruct sensitive inputs without direct access to raw data.
    
\textbf{Routing Manipulation Attacks}: Minimal feature perturbations can manipulate expert selection paths to induce harmful outputs. 

\textbf{Expert-induced Group Bias}: Personalization quality may degrade for long-tail users due to data sparsity, while expert specialization can further amplify representation bias across user groups.

\textbf{Dynamic Instability Adaptation}: Continuous or real-time updates of expert weights may lead to oscillations or error accumulation, ultimately undermining model reliability.

\paragraph{Security and Stability Safeguards}
To address these challenges, recent research has shifted toward integrating security-aware designs into MoE-based personalization, giving rise to a range of complementary mitigation strategies.

For routing-related vulnerabilities, prior work explores cryptographic protection mechanisms during inference, by concealing routing indices or shielding expert activation patterns \citep{shen2026secmoe, zhou2025cryptomoe}. In parallel, Federated MoE approaches provide a privacy-aware design paradigm for jointly addressing personalization and security. By keeping user data and routing-related components on the client side, they reduce the exposure of data-dependent routing signals to the server.
This principle is instantiated through a \textit{Data–Router Isolation} architecture, where methods such as PFL-MoE \citep{guo2021pfl} and pFedMoE \citep{yi2026pfedmoe} retain gating and personalization on-device, preventing the server from inferring routing decisions and thereby mitigating routing leakage and manipulation attacks.

To address group fairness and long-tail degradation caused by data heterogeneity, methods such as FFT-MoE \citep{hu2025fft} and HFedMoE \citep{fang2026hfedmoe} introduce heterogeneity-aware regularization. These approaches ensure that clients with limited or imbalanced data can still obtain equitable personalization, without compromising overall model generalization.

To mitigate dynamic instability, recent MoE approaches shift from loss-based routing regularization to structured routing constraints and controlled expert updates. ERMoE \citep{cheng2025ermoeeigenreparameterizedmixtureofexpertsstable} stabilizes specialization via eigenbasis reparameterization and alignment-based routing, without relying on loss-based routing regularization, while SR-MoE \citep{delibasoglu2026spectralmanifoldregularizationstable} imposes spectral constraints on the routing manifold for more consistent expert assignment. DES-MoE \citep{li2025dynamicexpertspecializationcatastrophic} further extends this paradigm with adaptive routing, gradient isolation, and staged fine-tuning for dynamic multi-domain adaptation.

\subsection{Pruning-based Personalization}
\label{sec:pruning}

\begin{figure}[ht]
    \centering
    \resizebox{\columnwidth}{!}{
    \begin{forest}
      for tree={
        font=\small,
        draw,
        align=center,
        thick,
        fill=blue!5, 
        rounded corners=3pt,
        grow'=east,    
        child anchor=west,    
        parent anchor=east,       
        anchor=west,
        l sep=0.8cm,   
        s sep=0.3cm,                   
        edge path={
          \noexpand\path [draw, \forestoption{edge}] 
          (!u.parent anchor) -- +(0.5,0) |- (.child anchor)\forestoption{edge label};
        },
      }
      [Pruning-based Personalization, fill=red!20
        [Personalized Techniques, fill=violet!20
         [Domain/Task-specific Pruning
          [ \citep{zhang2024pruningdomainspecificllmextractor, williams2025compressing,wang2024pruning}, fill=green!20!gray!10
          ]
         ]
         [
          User-specific Pruning
         ]
        ]
        [Safety Attacks, fill=violet!20
          [Diminished Robustness \& Increased Bias 
            [\citep{awal2025model,fu2025pruning,xu2024beyond},fill=green!20!gray!10]
          ]
          [
            Pruning-centered Attacks
            [\citep{pruningfewer},fill=green!20!gray!10]
          ]
        ]
        [Safety Defenses, fill=violet!20
          [Safety-preserving/recovering/enhancing Pruning
          [ \citep{fu2025pruning,huangantidote,wei2024assessing,sun2023simple}, fill=green!20!gray!10]]
          [Backdoor Defenses
          [ \citep{chapagain2025pruning}, fill=green!20!gray!10]]
        ]
      ]
    \end{forest}}
    \caption{Overview of pruning-based personalization.}
    \label{pruning_overview}
\end{figure}
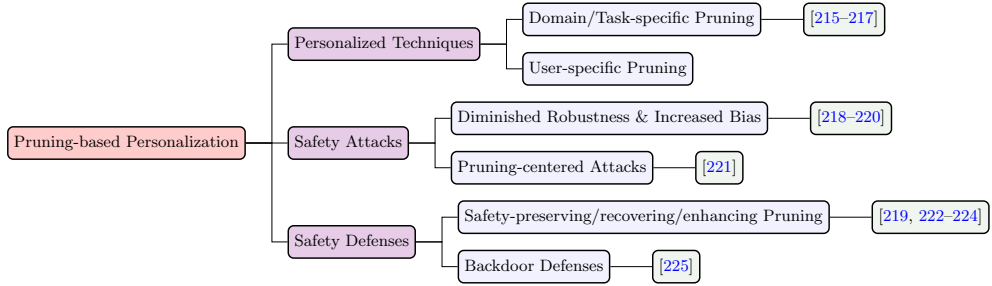

Figure \ref{pruning_overview} depicts the overall framework of this section, including introductions to personalized pruning techniques, relevant adversarial attacks, and defense mechanisms.

\paragraph{Personalized Pruning Techniques}

\begin{figure}[h]
\centering
\includegraphics[width=\linewidth]{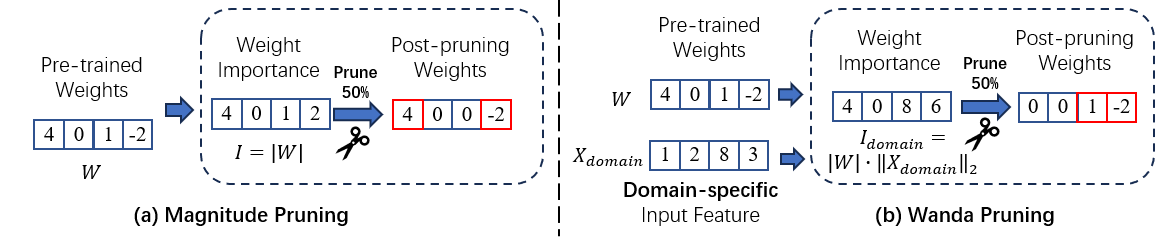}
\caption{Comparison of Magnitude Pruning and Wanda Pruning when 50\% of weights are pruned. (a) Magnitude Pruning uses absolute weight values to measure parameter importance \cite{han2015learning}. (b) Wanda Pruning adopts absolute weights combined with the L2 norm of input features for domain-aware importance evaluation \cite{sun2023simple}.}
\label{prune-comparison}
\end{figure}

Pruning methods compress LLMs by removing or zeroing out less critical parameters to enable efficient inference during deployment \cite{pruningfewer}.  This technique renders the structure of LLMs more compact and efficient, which is particularly important for deploying them on resource-constrained devices or in settings with limited storage capacity. A representative instance of conventional pruning techniques is Magnitude Pruning \cite{han2015learning}. This approach evaluates weight importance (denoted as $I$) solely based on absolute weight values \cite{sun2023simple}, formulated as:

\begin{equation}
I = |w|
\end{equation}

\noindent where $|.|$ denotes the absolute value, and Figure \ref{prune-comparison}(a) demonstrates an example of the overall pruning process when 50\% of the weights within the original weight matrix are pruned.

Unlike conventional pruning methods, \textbf{personalized pruning} enables pruned LLMs to generate content that matches individual user preferences. This method utilizes user-specific data to guide parameter pruning. The generic pre-trained LLM is thus selectively compressed to accommodate user preferences, yielding style- and content-consistent outputs with lower computational overhead. Despite its potential, pruning-based personalization remains underexplored. Existing pruning approaches primarily focus on domain- or task-specific datasets (e.g., math or code corpora) to derive pruned models \cite{zhang2024pruningdomainspecificllmextractor, williams2025compressing}, rather than adapting models to individual user preferences. This naturally suggests a transition from domain-level to user-level data for guiding pruning decisions in future research. For instance, when adopting a domain-specific dataset, Wanda Pruning (see Figure \ref{prune-comparison}(b)) calculates domain-aware parameter importance $I_{domain}$ by multiplying the absolute weight value with the L2 norm of the corresponding input feature vector $X_{domain}$, followed by eliminating parameters with low importance \cite{sun2023simple}:

\begin{equation}
I_{domain}=|w| \cdot ||X_{domain}||_{2}
\end{equation}

\noindent where $||.||_{2}$ represents the L2 norm. D-Pruner \citep{zhang2024pruningdomainspecificllmextractor} leverages gradient information from a composite loss that integrates both domain-specific and general capability signals to estimate parameter importance. By computing features based on task-specific datasets, P-pruning carries out clustering and pruning for self-attention heads and feed-forward network neurons \cite{wang2024pruning}. Besides, MixCal \citep{williams2025compressing} is a post-training calibration method that leverages in-domain data to boost the pruned model's performance on in-domain tasks. A natural extension is substituting these domain priors with user-specific preference data, allowing pruned models to capture personalized behaviors instead of task-oriented distributions. This transition shifts pruning from \emph{domain specialization} toward \emph{user personalization}.

\begin{figure}[h]
\centering
\includegraphics[width=\linewidth]{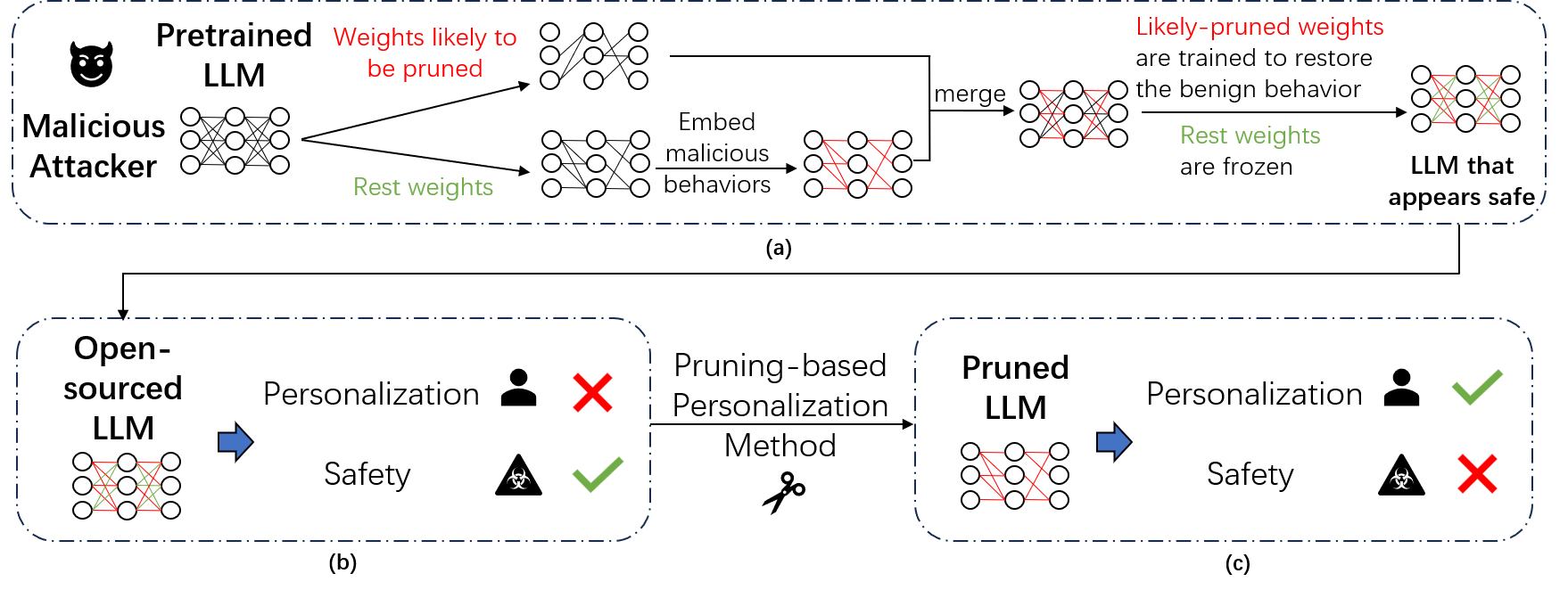}
\caption{Safety risk caused by pruning-based personalization \cite{pruningfewer}. (a) Attackers identify weights likely to be pruned and implant malicious functions in retained parameters. They fix tampered weights and retrain discarded ones to restore normal performance and evade safety checks. (b) The manipulated model is released publicly with no obvious risks. (c) End-user pruning alters model structure, triggers hidden threats and produces harmful outputs.}
\label{prune-figure}
\end{figure}

\paragraph{Safety and Privacy Risks}
Preliminary evidence suggests that personalized pruning methods may introduce new risks, such as diminished robustness \cite{awal2025model} and increased bias \cite{xu2024beyond}. In particular, \citet{fu2025pruning} reveal that pruning can disrupt the internal activation features of LLMs that are essential for lie detection, and that naively adjusting layer-wise pruning sparsity based on importance may inadvertently remove weights critical for maintaining lie detection performance. Another line of work demonstrates that an adversary can design an LLM that exhibits benign behavior on the surface, yet reveals malicious characteristics when pruned \cite{pruningfewer} (see Figure \ref{prune-figure}). Specifically, the attacker first identifies weights that are likely to be eliminated during pruning, then embeds malicious functionality within the remaining weights. Subsequently, while keeping the tampered weights fixed, the weights slated for removal are trained to recover the model's benign behavior and ensure it can pass standard safety evaluations. The final model can then be distributed through model-sharing platforms and appears innocuous. However, once end users perform pruning on the model, the removal of these parameters modifies the model architecture and activates the hidden malicious behavior, resulting in harmful generations.


\paragraph{Defense and Mitigation Strategies}
Scholars have proposed various solutions to address these issues. For example, \citet{fu2025pruning} introduce the Truthful Pruning aligned by Layer-wise Outliers (TPLO) method, which prioritizes layers containing more activation outliers and stronger discriminative features simultaneously; this approach is shown to preserve the original performance of LLMs while maintaining the critical inner-state features necessary for robust lie detection. \citet{huangantidote} propose Antidote, a post-fine-tuning solution based on the principle that a model can be restored from harmful behaviors by removing the harmful parameters, regardless of how those parameters were formed during fine-tuning. \citet{wei2024assessing} indicate that the least safety-critical neurons can undermine safety, and that eliminating them may improve the model's overall safety. Additionally, \citet{hasan2024pruning} demonstrate that moderate parameter pruning (10–30\%) using the Wanda method \cite{sun2023simple} can actually improve the resistance of LLMs to jailbreaking attacks. Regarding backdoor defense, experimental evaluation suggests that gradient-based pruning performs best against syntactic triggers (i.e., syntactic manipulations), whereas methods such as reinforcement learning are more effective against stylistic attacks (i.e., manipulations of text style) \cite{chapagain2025pruning}.

\section{System-level Personalization and Associated Safety Risks}
\label{sec:system}

\begin{figure}[!ht]
    \centering
    \includegraphics[width=\columnwidth]{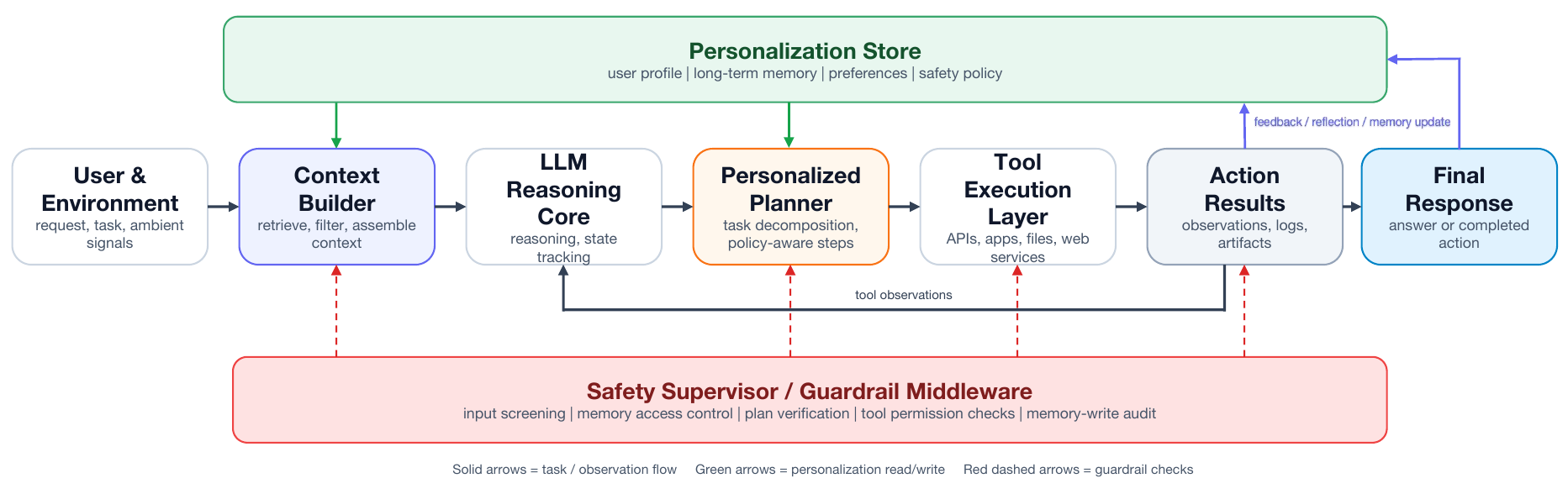}
    \caption{A representative system-level architecture for personalized LLM agents, highlighting the interaction among user/environment inputs, personalization storage, context construction, reasoning and planning, tool execution, feedback updates, and guardrail-based safety supervision \cite{weng2023agent}.}
    \label{fig:system_level_personalized_agent_architecture}
\end{figure}

\begin{figure}[!ht]
    \centering
   \resizebox{\columnwidth}{!}{
    \begin{forest}
      for tree={
        font=\small,
        draw,
        align=center,
        thick,
        fill=blue!5,
        rounded corners=3pt,
        grow'=east,
        child anchor=west,
        parent anchor=east,
        anchor=west,
        l sep=0.8cm,
        s sep=0.35cm,
        edge path={
          \noexpand\path [draw, \forestoption{edge}]
          (!u.parent anchor) -- +(0.5,0) |- (.child anchor)\forestoption{edge label};
        },
      }
      [Personalized LLM Agents, fill=red!20
       [
        Personalized Techniques, fill=violet!20
        [
          Personalized Agent Environment
                [
                  Persona-based Simulation \citep{wang2025user,chen2025recusersim,ma2025pub},fill=green!20!gray!10
                ]
                [
                  Temporal-aware Simulation \citep{wanyan2025temporal},fill=green!20!gray!10
                ]
                [
                  Interactive Multi-agent Simulation \citep{sun2025training,dou2025simulatorarena},fill=green!20!gray!10
                ]
        ]
        [
          Memory
                [
                  Retrieval-based Memory \citep{brown2020language,lewis2020retrieval,salemi2024personalized},fill=green!20!gray!10
                ]
                [
                  Persistent User Memory \cite{zhong2024memorybank,zhang2025prime},fill=green!20!gray!10
                ]
                [
                  Memory Maintenance and Updating \cite{zhong2024memorybank,lu2025bayesian},fill=green!20!gray!10
                ]
                [
                  Adaptive Memory \citep{sun2024learning,sarin2025memoria},fill=green!20!gray!10
                ]
        ]
        [
          Personalized Planning
                [Experience-informed Decision Loops \citep{park2023generative},fill=green!20!gray!10]
                [User-adaptive Task Decomposition \citep{gao2025flowreasoner},fill=green!20!gray!10]
        ]
        [
          Tools
                [Personalized Tool Workflows \citep{hao2025evaluating},fill=green!20!gray!10]
                [Learned Tool Preferences \citep{xu2025petoolllm},fill=green!20!gray!10]
        ]
       ]
       [
        Safety Risks, fill=violet!20
        [
           Inherited Personalization Risks
              [
                Adversarial Context Injection \cite{yang2024watch,zhang2025agentsecuritybenchasb},fill=green!20!gray!10
              ]
              [
                Uncontrolled Memory Influence \cite{zhang2025agentsecuritybenchasb,wu2026personalizedsafetyllmsbenchmark},fill=green!20!gray!10
              ]
        ]
        [
          Agent-specific Risks
          [
             {Planning-stage Risks\\ (Bias, Sycophancy)}, fill=green!20!gray!10
          ]
          [
            {Execution-stage Risks\\ (Unsafe Tool Invocation,\\ Plan–Action Misalignment)},fill=green!20!gray!10
          ]
        ]
       ]
       [
        Safety Defenses, fill=violet!20
        [
          Input \& Memory Level
          [
            Adversarial Prompt Mitigation \cite{yang2024watch,zhang2025agentsecuritybenchasb},fill=green!20!gray!10
          ]
          [
            Controlled Memory Access \cite{zhang2025agentsecuritybenchasb,wu2026personalizedsafetyllmsbenchmark},fill=green!20!gray!10
          ]
        ]
        [
          Planning Level
          [
            Risk-aware Planning \cite{huang2025frameworkbenchmarkingaligningtaskplanning,wang2025madramultiagentdebateriskaware,yao2025llmbasedmultiagentreliablerealworld,wu2026personalizedsafetyllmsbenchmark},fill=green!20!gray!10
          ]
          [
            Bias/Sycophancy Reduction,fill=green!20!gray!10
          ]
        ]
        [
          Execution Level
          [
            Consistency \& Plan–Action Alignment \cite{mou2026toolsafeenhancingtoolinvocation,betser2026agentrimtoolriskmitigation},fill=green!20!gray!10
          ]
          [
            Execution Safeguards \cite{wu2025psg},fill=green!20!gray!10
          ]
        ]
       ]
      ]
    \end{forest}}
    \caption{System-level personalized LLM agents: architectural components, safety risks and mitigation strategies.}
    \label{fig:personalized_agent_safety}
\end{figure}
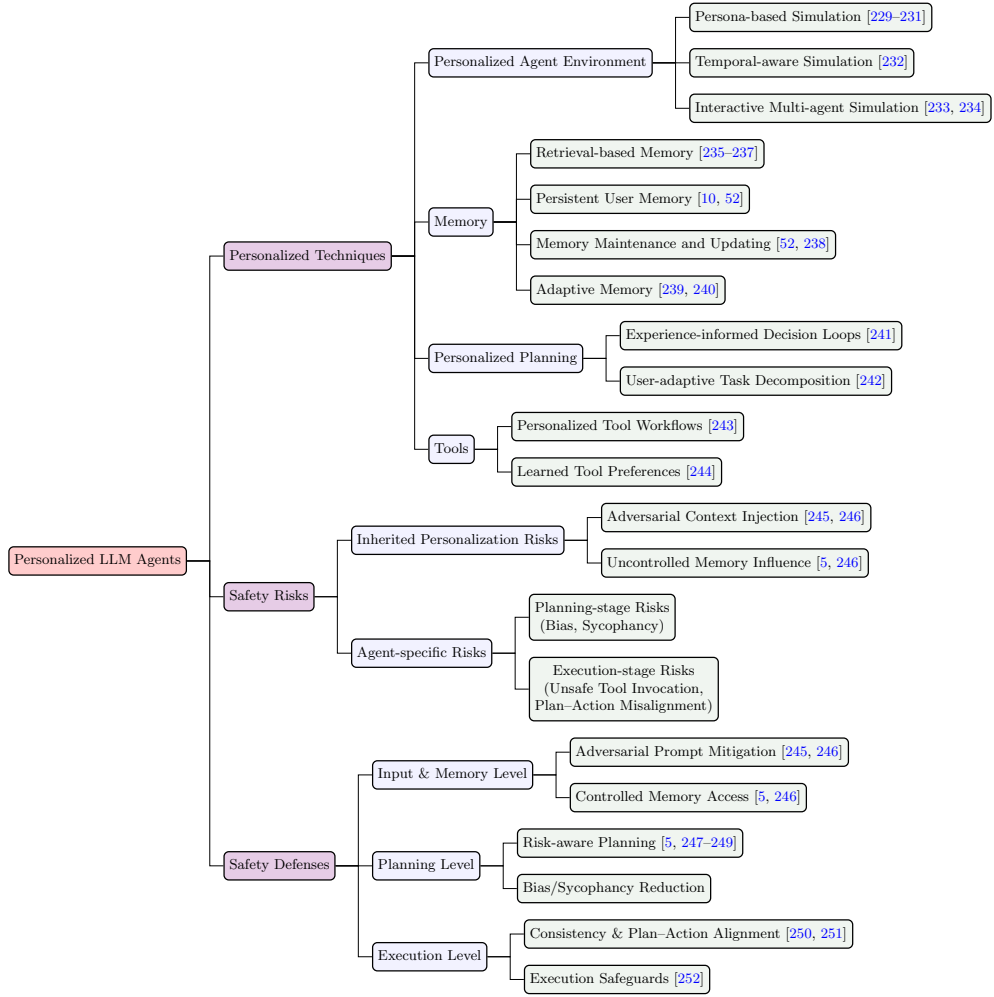

System-level personalization is predominantly realized through agent-based architectures, which introduce user-aware mechanisms into the system pipeline of LLM agents \cite{weng2023agent}, enabling personalized behavior throughout reasoning, planning, memory, and action. Unlike approaches that primarily adapt model parameters or internal representations, these methods achieve personalization through system-level coordination among components such as memory retrieval, tool usage, and decision-making modules. Figure \ref{fig:system_level_personalized_agent_architecture} illustrates a representative architecture for such systems, while Figure \ref{fig:personalized_agent_safety} provides an overview of personalized agent systems and their associated safety considerations.

\subsection{System-level Personalization Techniques}

Personalized agent systems typically consist of four interacting components:

\textbf{Environment}: The digital or physical context defining the agent's observation space (e.g., webpages or sensor streams) and action space (e.g., navigation or interface interactions). Personalized Agent Environments model realistic user behavior and interaction dynamics for training and evaluation. Recent work leverages LLMs to build persona-based simulation systems with explicit persona, memory, and decision modules, enabling more realistic user modeling in recommendation and dialogue systems \cite{wang2025user,chen2025recusersim,ma2025pub}. Temporal-aware simulation frameworks such as DyTA4Rec further incorporate evolving user interests and sequential interaction histories \cite{wanyan2025temporal}. Interactive multi-agent simulation frameworks, including UserVille and SimulatorArena, further capture complex social interactions and feedback dynamics by integrating multiple user simulators into reinforcement learning environments \cite{sun2025training,dou2025simulatorarena}.

\textbf{Memory}: In contrast to memory-based personalization methods that primarily focus on storing and retrieving user information, memory in personalized agents serves as a persistent cognitive component that supports long-term interaction, reflective reasoning, and adaptive behavior across sessions. Early approaches relied on In-Context Learning (ICL) \cite{brown2020language}, which encodes user information directly in prompts but is limited by context window size. RAG extends this capability by retrieving relevant user history from external storage during inference \cite{lewis2020retrieval,salemi2024personalized}. To support cross-session personalization, frameworks such as MemoryBank \cite{zhong2024memorybank} and PRIME \cite{zhang2025prime} introduce hierarchical memory architectures that separate transient interactions from persistent user profiles. Beyond storage and retrieval, memory in agent systems must additionally support the updating of outdated preferences and the selective forgetting of obsolete information—capabilities that are essential for long-term behavioral consistency and are discussed more broadly in the context of memory lifecycle management (Section~\ref{sec:memory_mgmt}). More recent studies further treat memory as an adaptive component rather than a passive store: Memoria \cite{sarin2025memoria} allows agents to reorganize stored information dynamically, while Test-Time Training (TTT) methods \cite{sun2024learning} update model parameters during inference, internalizing user-specific knowledge into the model itself.

\textbf{Planning}: Procedures that decompose goals into executable sub-tasks while adapting strategies based on user context and feedback. Recent work explores personalized planning through experience-informed decision loops and adaptive task decomposition. Generative Agents \cite{park2023generative} introduce a \emph{memory--reflection--planning} loop that incorporates accumulated experiences into decision making, allowing agents to maintain long-term behavioral coherence and generate actions aligned with user personas and goals. Building on this idea, FlowReasoner \cite{gao2025flowreasoner} learns query-specific multi-agent workflows through reinforcement learning, enabling customized reasoning pipelines for different tasks and users.

\textbf{Tools}: External functions or APIs that extend agent capabilities, such as search engines or action-execution utilities. Tool use is another important dimension of personalization, as users often differ in preferred tools, workflows, and levels of agent initiative. Recent work explores personalized tool workflows by evaluating whether agents adapt tool-selection and execution strategies to individual users. ETAPP \cite{hao2025evaluating} characterizes this behavior along the dimensions of personalization and proactivity, assessing whether agents tailor tool usage behaviors to user-specific preferences and interaction patterns. From a learning perspective, PEToolBench \cite{xu2025petoolllm} formulates personalized tool use as a learnable objective based on interaction histories. Building on this benchmark, PEToolLLaMA \cite{xu2025petoolllm} demonstrates that models can internalize user-specific tool preferences through training rather than relying on static prompts. These approaches highlight the importance of modeling user workflows in tool-augmented agents.

\subsection{Safety Risks}

Agent-based personalization systems, built upon LLMs as their central reasoning and acting backbone, inherit—and in many cases amplify—the risks observed in earlier personalization paradigms. In particular, the integration of persistent memory and multi-step decision-making introduces new attack surfaces, making these systems susceptible to covert and long-term backdoor attacks \cite{yang2024watch,zhang2025agentsecuritybenchasb}, which may remain dormant during benign interactions but activate under specific triggers. As user preferences are encoded through prompts and maintained in memory, personalized agents inherit vulnerabilities from prompt- and retrieval-based personalization frameworks, including privacy leakage, prompt injection, and retrieval poisoning \cite{zhang2025agentsecuritybenchasb}. Adversarial prompts can extract Personally Identifiable Information (PII) from retrieval-augmented systems \cite{wu2026personalizedsafetyllmsbenchmark}, while indirect prompt injection via malicious external content can compromise downstream reasoning and actions \cite{greshake2023not}. We categorize these inherited risks into two classes: \emph{adversarial context injection}, where malicious inputs are injected into the model context, and \emph{uncontrolled memory influence}, where stored or retrieved (potentially poisoned) information persistently affects downstream behaviors and decisions.

Beyond inherited risks, agent-based systems shift from passive conditioning to active planning and action execution. While planning and execution risks exist in any autonomous agent, personalization amplifies them in distinct ways. At the planning stage, the central challenge is not merely whether plans are safe in general, but whether task decomposition and reasoning remain safe when conditioned on individual user preferences—which may introduce subtle biases, reinforce harmful stereotypes, or induce sycophantic compliance with user expectations. Although general-purpose planning safety has received increasing attention, the specific intersection with personalization—where user-specific preferences may distort safety-critical reasoning—remains largely underexplored.

At the execution stage, personalization further challenges uniform safety enforcement: a tool invocation that is safe for an expert user may pose a significant risk for a novice, yet most existing safeguards apply the same policy across all users. More broadly, current work reveals a persistent divide between planning-time risk awareness and execution-time safety enforcement, and this gap is further widened in personalized settings, where user-specific context must propagate across both stages. Therefore, cross-stage safety propagation in personalized agent systems remains an open challenge.

\subsection{Defense and Mitigation Strategies}

To address the above risks, existing defenses operate at multiple stages of the agent pipeline. At the input and memory level, adversarial prompt mitigation techniques \cite{yang2024watch,zhang2025agentsecuritybenchasb} detect and filter malicious context injections before they influence agent reasoning, while controlled memory access mechanisms \cite{zhang2025agentsecuritybenchasb,wu2026personalizedsafetyllmsbenchmark} limit the influence of potentially poisoned or stale information on downstream behaviors.

At the planning stage, risk-aware planning frameworks incorporate safety constraints into task decomposition and reasoning. SafePlan-Bench \cite{huang2025frameworkbenchmarkingaligningtaskplanning} evaluates planning safety and introduces Safe-Align for aligned task decomposition, while SafeAware-VH \cite{wang2025madramultiagentdebateriskaware} employs multi-agent debate to support safety-aware planning. WandaPlan \cite{yao2025llmbasedmultiagentreliablerealworld} further extends this direction to real-world scenarios with anti-fraud detection, and PENGUIN \cite{wu2026personalizedsafetyllmsbenchmark} explores personalized risk-aware planning for sequential decision-making tasks.

At the execution stage, step-level safeguards detect and block unsafe tool invocations. TS-Guard \cite{mou2026toolsafeenhancingtoolinvocation} enables step-level detection of unsafe tool calls, while AGENTRIM \cite{betser2026agentrimtoolriskmitigation} enforces least-privilege tool access through interface-level constraints. Guardrail-based approaches (e.g., Qwen3Guard \cite{zhao2025qwen3guard}, LLMGuard \cite{sun2026llmguard}) further regulate runtime behaviors, and PSG-Agent \cite{wu2025psg} extends this paradigm with personalized cross-turn safety monitoring. Despite these advances, a persistent gap remains between planning-time risk awareness and execution-time safety enforcement, highlighting the need for integrated, cross-stage defense mechanisms in personalized agent systems.

\section{Multimodal Personalization and Associated Safety Risks}
\label{sec:multimodal}
\begin{figure}[ht]
    \centering
    \resizebox{\columnwidth}{!}{%
    \begin{forest}
      for tree={
        font=\small,
        draw,
        align=center,
        thick,
        fill=blue!5, 
        rounded corners=3pt,
        grow'=east,    
        child anchor=west,    
        parent anchor=east,       
        anchor=west,
        l sep=0.8cm,   
        s sep=0.3cm,                   
        edge path={
          \noexpand\path [draw, \forestoption{edge}] 
          (!u.parent anchor) -- +(0.5,0) |- (.child anchor)\forestoption{edge label};
        }
      }
      [Multimodal Personalization, fill=red!20
        [Personalized Techniques, fill=violet!20
          [Multimodal Alignment  and Fusion
            [User-agnostic \citep{amirloo2024understandingalignmentmultimodalllms}, fill=green!20!gray!10]
            [Preference-aware \cite{wang-etal-2024-mdpo,zhao-etal-2025-omnialign,zhao2025guidingcrossmodalrepresentationsmllm},fill=green!20!gray!10]
            [User-conditioned \cite{PMMAECTR2025, guan2025multimodalrecommendationselfcorrectivepreference, sun2026karmaknowledgeactionregularizedmultimodal},fill=green!20!gray!10]
          ]
        ]
        [Safety Attacks and Defenses, fill=violet!20
          [Cross-modal Privacy and Data Leakage
            [On-device Personalization \cite{wang2024mememo, cho2024hollowed},fill=green!20!gray!10]
            [Federated Learning  \cite{feng2024fedpam},fill=green!20!gray!10]
            [Differential Privacy \cite{flemings2024differentially},fill=green!20!gray!10]
            [Adversarial Perturbation \cite{van2023anti},fill=green!20!gray!10]
          ]
          [Multimodal Bias and Representation Distortion
            [Context Steering \cite{pandey2024cos}, fill=green!20!gray!10]
            [Structured Prompting \cite{furniturewala2024thinking,gallegos2025self}, fill=green!20!gray!10]
            [Causal Analysis \cite{sun2024causal}, fill=green!20!gray!10]
            [Fairness and Safety Checks \cite{wang2025generative}, fill=green!20!gray!10]
          ]
          [Behavioral and Cognitive Influence Risks
            [Diversity Enhancement \cite{chen2018fast},fill=green!20!gray!10]
            [Multi-agent Generation \cite{zhang2024see},fill=green!20!gray!10]
            [User-controllable Generation \cite{wang2022user},fill=green!20!gray!10]
          ]
          [Identity-preserving Misuse and Adversarial Attacks
            [Machine Unlearning\cite{wu2025unlearning,hu2024eraser},fill=green!20!gray!10]
            [Digital Watermarking \cite{liu2024countering},fill=green!20!gray!10]
          ]
          [Authenticity and Reliability Risks
            [Knowledge-enhanced Generation\cite{wang2025generative},fill=green!20!gray!10]
            [Robust Evaluation \cite{wang2025generative},fill=green!20!gray!10]
            [Post-processing Refinement \cite{xu2025personalized},fill=green!20!gray!10]
          ]
        ]
      ]
    \end{forest}%
    }
    \caption{Overview of multimodal personalization paradigms.}
    \label{mutimodal_overview}
\end{figure}
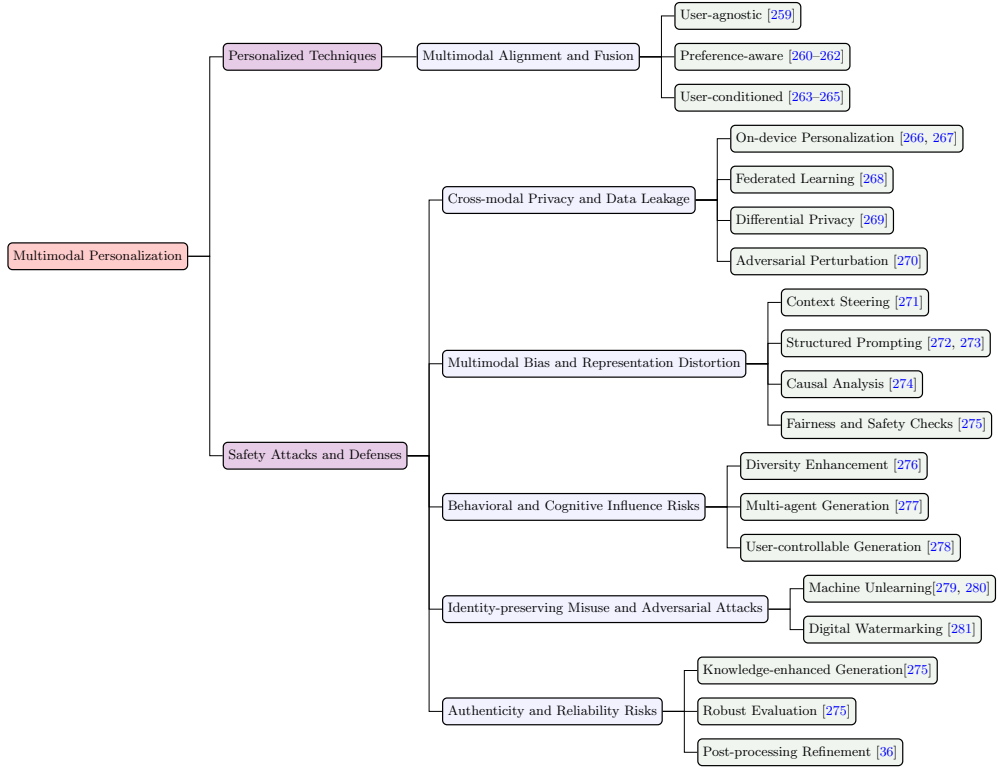

Personalization plays a critical role in real-world applications such as search and recommendation, where user preferences are inherently expressed through multimodal signals. Multimodal Large Language Models (MLLMs) \cite{hurst2024gpt,chen2025janus,team2023gemini} provide a unified framework for modeling and reasoning over heterogeneous modalities, enabling a more comprehensive understanding of user preferences beyond text-only information. Compared to text-centric personalization, extending personalization to MLLMs requires handling heterogeneous modalities and cross-modal interactions, raising new technical challenges and safety concerns. Figure \ref{mutimodal_overview} presents an overview taxonomy of multimodal personalization techniques, together with their corresponding safety risks and associated mitigation strategies.

\paragraph{Multimodal Personalization Techniques}
As personalization extends from text-only LLMs to multimodal large language models, personalized systems must jointly model heterogeneous user information across modalities such as language, vision, and audio. In this setting, multimodal personalization is fundamentally built upon tightly coupled \textbf{multimodal alignment} and \textbf{multimodal fusion} mechanisms, which are typically jointly optimized during representation learning. Multimodal alignment aims to learn semantically consistent representations across heterogeneous modalities, enabling personalized signals from different modalities to be jointly understood, while multimodal fusion integrates complementary multimodal information to construct coherent personalized representations.

Existing approaches primarily learn a shared latent space across modalities based on explicit cross-modal correspondence information (e.g., image–text pairs), where the alignment objective is to ensure that model outputs remain consistent with the associated multimodal context. Such formulations are typically coupled with \emph{user-agnostic fusion mechanisms}, such as shared latent spaces or fixed cross-modal interaction modules (e.g., dual-encoder or cross-attention architectures), leading to globally consistent yet user-independent representations that do not account for individual differences \cite{li2023blip2bootstrappinglanguageimagepretraining,liu2024improvedbaselinesvisualinstruction,amirloo2024understandingalignmentmultimodalllms}.

To move beyond purely latent space alignment and fusion, recent work has introduced \emph{preference-aware alignment}, where alignment objectives are guided by preference information rather than explicit cross-modal correspondence alone. This line of work evolves along three complementary directions. Optimization-centric approaches, such as mDPO \cite{wang-etal-2024-mdpo}, extend Direct Preference Optimization to multimodal settings to address the mismatch between multimodal inputs and textual supervision. Data-centric approaches, exemplified by OmniAlign-V \cite{zhao-etal-2025-omnialign}, construct a relatively large-scale multimodal preference dataset with 200K high-quality samples, demonstrating that high-quality preference supervision can substantially improve alignment with human preferences. Automation-driven approaches \cite{zhao2025guidingcrossmodalrepresentationsmllm}, such as MAPLE, further scale preference learning by leveraging MLLMs to generate preference data and optimize relative preference objectives. Notably, while these methods do not explicitly model users, they implicitly influence multimodal fusion by reshaping cross-modal representations and modulating modality interactions. Overall, this paradigm reflects a shift toward scalable, structured, and model-driven preference learning, laying the foundation for more personalized alignment and implicitly personalized fusion.

While preference-aware alignment captures population-level preference information, recent work has further advanced toward \emph {user-centered alignment}, where alignment is explicitly conditioned on individual users rather than aggregated preferences. This line of research can be understood along three complementary directions, reflecting different aspects of user-centered alignment, including representation modeling, optimization dynamics, and objective design. User-specific alignment approaches model alignment as a user-dependent function rather than a global mapping. Methods such as PMMAE \cite{PMMAECTR2025} show that different users may exhibit distinct multimodal alignment patterns, introducing personalized encoding mechanisms that also induce user-conditioned fusion processes. Adaptive alignment approaches further consider the dynamic nature of user preferences. For instance, MSPA \cite{guan2025multimodalrecommendationselfcorrectivepreference} incorporates self-corrective mechanisms to iteratively refine alignment, enabling the fusion of multimodal information to be dynamically updated based on user feedback. More recent work has begun to address the tension between personalization and semantic consistency. Constraint-aware approaches, such as KARMA \cite{sun2026karmaknowledgeactionregularizedmultimodal}, identify a knowledge-action gap, showing that optimizing for personalized actions alone may lead to degraded semantic consistency, and propose to regularize both alignment objectives with semantic reconstruction and the associated fusion process to preserve generalization.

\begin{figure}[h]
\centering
\includegraphics[width=\linewidth]{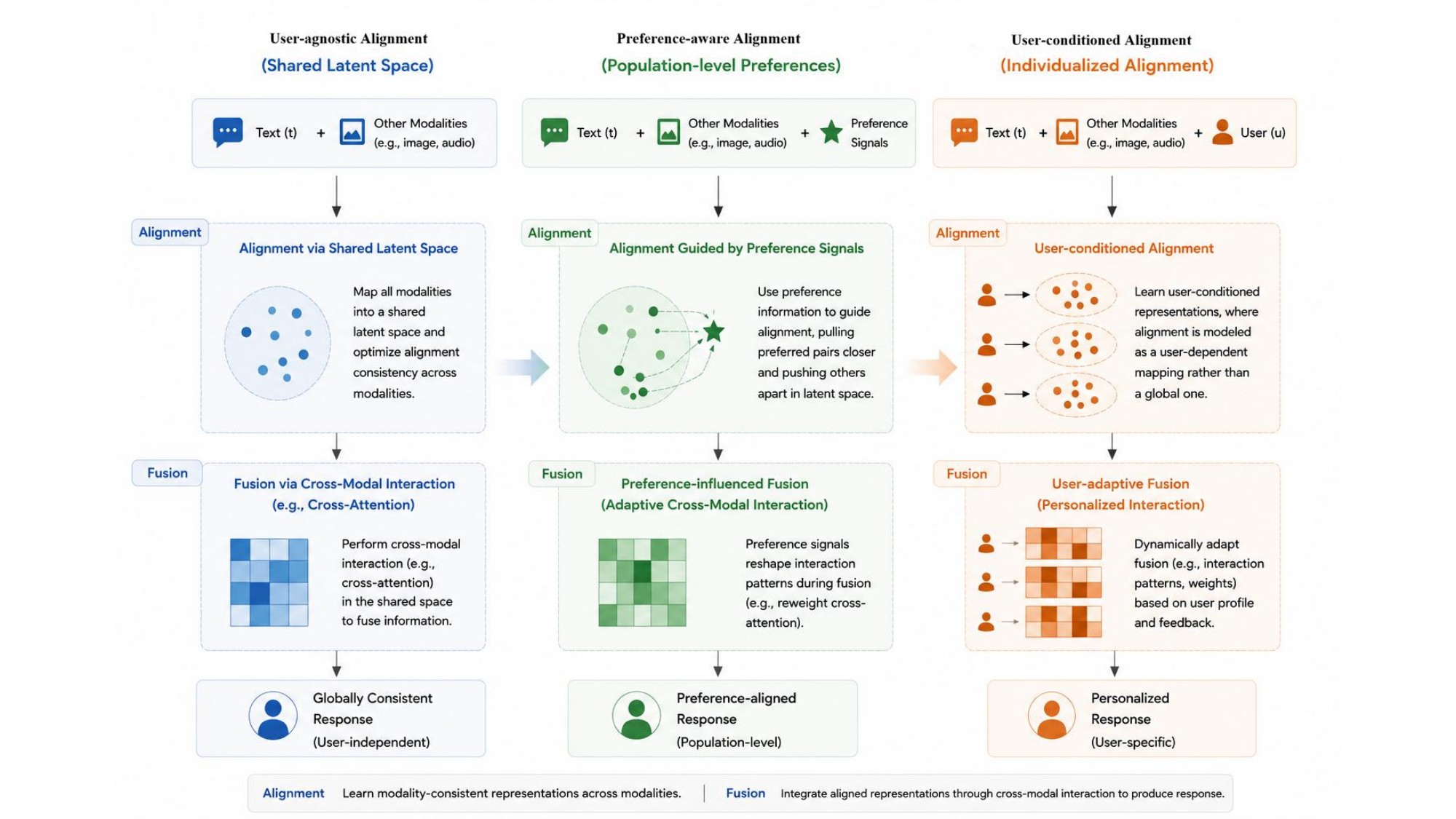}
\caption{Comparison of multimodal personalization modeling paradigms, illustrating the evolution from globally shared and user-independent modeling toward preference-influenced and ultimately user-conditioned systems. Existing approaches progressively extend multimodal alignment and fusion from global shared representations \citep{li2023blip2bootstrappinglanguageimagepretraining} to preference-aware \citep{zhao2025guidingcrossmodalrepresentationsmllm} and personalized modeling mechanisms \citep{PMMAECTR2025}. Notably, alignment and fusion are not always explicitly separated as illustrated in the figure, but are often implicitly coupled and jointly optimized during representation learning.}
\label{fig:multimodal}
\end{figure}

Taken together, as illustrated in Figure \ref{fig:multimodal}, these paradigms reveal a fundamental shift in multimodal personalization modeling, from globally shared and user-independent alignment/fusion mechanisms toward preference-influenced and ultimately user-conditioned systems, where alignment defines the objective of personalization, while fusion provides the mechanism through which personalized representations are realized.

\paragraph{Safety and Privacy Risks}

While personalized MLLMs significantly enhance user experience through tailored multimodal content generation, their reliance on heterogeneous user signals and cross-modal representations introduces unique security and trustworthiness challenges beyond those observed in text-only personalization systems. These risks can be summarized into the following key dimensions:


\textbf{Cross-modal Privacy and Data Security}: 
Compared with text-only LLM-based personalization, multimodal personalized MLLMs extend user modeling to heterogeneous modalities, such as facial features, bodily traits, personal documents (emails, social media), and historical behaviors \cite{chen2024large}. While this enables more fine-grained personalization, it also significantly expands the attack surface of sensitive information exposure \cite{shin2018privacy}. In particular, cross-modal representations introduce additional privacy risks beyond memorization, as sensitive attributes may be inferred through modality alignment and interaction. 


\textbf{Multimodal Bias and Representation Distortion}: 
Personalized MLLMs may unintentionally learn and amplify biases and stereotypes embedded across multiple modalities, potentially causing discriminatory effects and reinforcing harmful content \cite{fu2020fairness,gao2020counteracting,zehlike2022fair}. Beyond conventional dataset bias, multimodal systems face additional challenges during modality fusion. In particular, ``text bias'' may cause models to over-rely on textual signals while suppressing critical visual or audio information, resulting in distorted personalization that fails to capture the full multimodal context \cite{wu2024personalized}.


\textbf{Behavioral and Cognitive Influence Risks}: 
By tailoring multimodal content too closely to a user's existing preferences and beliefs, personalized generation may intensify filter bubbles and reduce exposure to diverse perspectives \cite{lazovich2023filter}. Compared with text-only systems, multimodal personalization can further strengthen emotional engagement and persuasive influence through coordinated visual, textual, and auditory cues, potentially exacerbating social and political polarization.


\textbf{Identity-preserving Misuse and Adversarial Attacks}: 
The ability to generate high-fidelity, identity-preserving multimodal content, such as personalized faces, voices, or videos, substantially increases the risk of malicious misuse, including identity theft, impersonation, and unacceptable editing of a person's likeness \cite{xu2025personalized}. In addition, these systems remain vulnerable to traditional security threats such as shilling attacks, which attempt to manipulate recommendations and personalized interactions \cite{gunes2014shilling}.


\textbf{Authenticity and Reliability Risks}: 
In high-stakes domains such as medicine, law, and news, maintaining factual consistency and reliability across modalities is essential for user trust \cite{xu2025personalized}. The perceptual realism of multimodal generation may further increase the credibility of inaccurate or harmful outputs, making cross-modal hallucinations and misleading personalized content more difficult for users to identify. Consequently, these systems must be carefully monitored to prevent the generation of unhealthy or harmful information, particularly for vulnerable groups such as teenagers \cite{amodei2016concrete}.

\paragraph{Defense and Mitigation Strategies}
To address the above security and trustworthiness challenges in Personalized MLLMs, the sources outline several technical and procedural solution strategies:

To protect highly sensitive user data while maintaining personalization utility, researchers have explored decentralized training paradigms, such as on-device personalization \cite{wang2024mememo, cho2024hollowed} and federated learning (e.g., FedPAM~\cite{feng2024fedpam}), which enable adaptation without exposing raw user data, alongside noise-based privacy mechanisms, including differential privacy \cite{flemings2024differentially} and adversarial perturbations \cite{van2023anti},to obscure sensitive signals embedded within multimodal inputs and generated content.


To mitigate multimodal bias and improve fairness in personalized generation, existing approaches refine model behavior through context steering \cite{pandey2024cos} and structured prompting \cite{furniturewala2024thinking,gallegos2025self} to suppress biased outputs and reduce over-reliance on dominant textual signals during multimodal generation. Other approaches leverage causal analysis \cite{sun2024causal} to identify and mitigate discriminatory patterns and spurious cross-modal correlations arising during multimodal training and generation. Integrated trustworthy generation frameworks, such as GeneRec \cite{wang2025generative}, further incorporate explicit fairness and safety constraints directly into the generation loop to filter problematic content before user exposure.


To mitigate behavioral and cognitive influence risks in personalized multimodal generation, existing studies introduce controlled diversity and perspective-aware generation mechanisms to reduce excessive preference reinforcement and filter bubble effects. Representative approaches include inference-time diversity enhancement \cite{chen2018fast} and user-controllable generation \cite{wang2022user}, which enable users to explicitly request diverse or even conflicting viewpoints rather than receiving overly aligned content. In multimodal settings, broader exposure to diverse perspectives may help alleviate over-reinforcement effects arising from highly aligned textual, visual, and auditory content. Multi-agent generation frameworks \cite{zhang2024see} further expand this paradigm by assigning distinct generative personas to different agents, collectively producing a broader range of perspectives and mitigating narrow behavioral conditioning caused by over-personalized multimodal interactions.


Several strategies have been developed to protect identity in personalized multimodal generation. Machine unlearning techniques \cite{wu2025unlearning,hu2024eraser} enable models to forget specific concepts or individual identities, thereby reducing risks of identity leakage, impersonation, and unauthorized identity preservation. Digital watermarking schemes \cite{liu2024countering} further improve content traceability and help deter malicious editing or redistribution of generated multimodal content.


To improve authenticity and reliability in personalized multimodal generation, recent approaches incorporate knowledge-enhanced generation mechanisms \cite{wang2025generative} that dynamically retrieve factual information during content creation to improve factual consistency, safety, and legal compliance. Additional safeguards include multi-dimensional evaluation paradigms \cite{wang2025generative} that move beyond conventional downstream task metrics toward direct assessment of generation quality, multimodal consistency, and safety compliance, often complemented by post-processing refinement. Transparent governance protocols \cite{xu2025personalized} further support accountable moderation and explainable generation processes to sustain user trust and uphold public safety standards.


\section{Paradigm-agnostic Safety Risks of LLM Personalization}
\label{sec:agnostic}
Beyond the safety concerns associated with specific personalization techniques discussed above, personalized LLMs also introduce a set of systemic concerns that arise from the behavior of the personalization system as a whole. These risks do not stem from any single component or technique in isolation, but emerge from the cumulative and long-term effects of adaptive, user-centered model behavior.

One prominent concern is \textbf{bias reinforcement} \citep{Klimashevskaia_2024}. Personalized models tend to increasingly
align with users' preferences, beliefs, and interaction patterns. While this alignment can improve user experience, it may also amplify existing biases through feedback loops. As models progressively adapt to and anticipate user tendencies, such biases can accumulate implicitly and remain largely unnoticed, ultimately exacerbating safety risks and narrowing users' exposure to alternative perspectives.

Another systemic risk is \textbf{anthropomorphism} \citep{kirk2024benefits}, whereby users assign human traits, intentions, or emotional capacities to non-human agents. Sustained personalized interactions can foster increased trust and emotional reliance on LLMs, leading users to disclose more personal or sensitive information. This heightened willingness to share, in turn, increases the risk of privacy leakage and unintended incorporation of personal data into the model's behavior or memory.

Moreover, personalization systems may engage in \textbf{algorithmic and inferential profiling} \citep{kirk2024benefits}, particularly when adapting to specific user groups. This risk is especially pronounced for vulnerable populations such as children. For younger children, persistent  profiling that frames them as low-cognition users may result in overly simplified or restrictive responses, potentially hindering cognitive development over time. For older children, misclassification as adults may prematurely expose them to content that bypasses child-specific safety filtering, creating risks of inappropriate or harmful information exposure.

Finally, repeated interaction with adaptive safety mechanisms can give rise to \textbf{safety gaming and evasion} \citep{zhang2025enhancingjailbreakattacksllms}, as users gradually infer personalized safety boundaries and adjust their behavior to
circumvent safeguards—an emergent challenge that cannot be addressed through static or component-level controls alone.

Addressing these system-wide and emergent risks requires moving beyond isolated safeguards toward holistic, adaptive mitigation strategies that operate across the entire personalization lifecycle. We discuss potential solutions in Section \ref{sec:solutions}.

\section{Personalization Evaluation Benchmarks}
\label{sec:benchmark}

\begin{figure}[htbp]
\begin{center}
\includegraphics[width=\textwidth]
{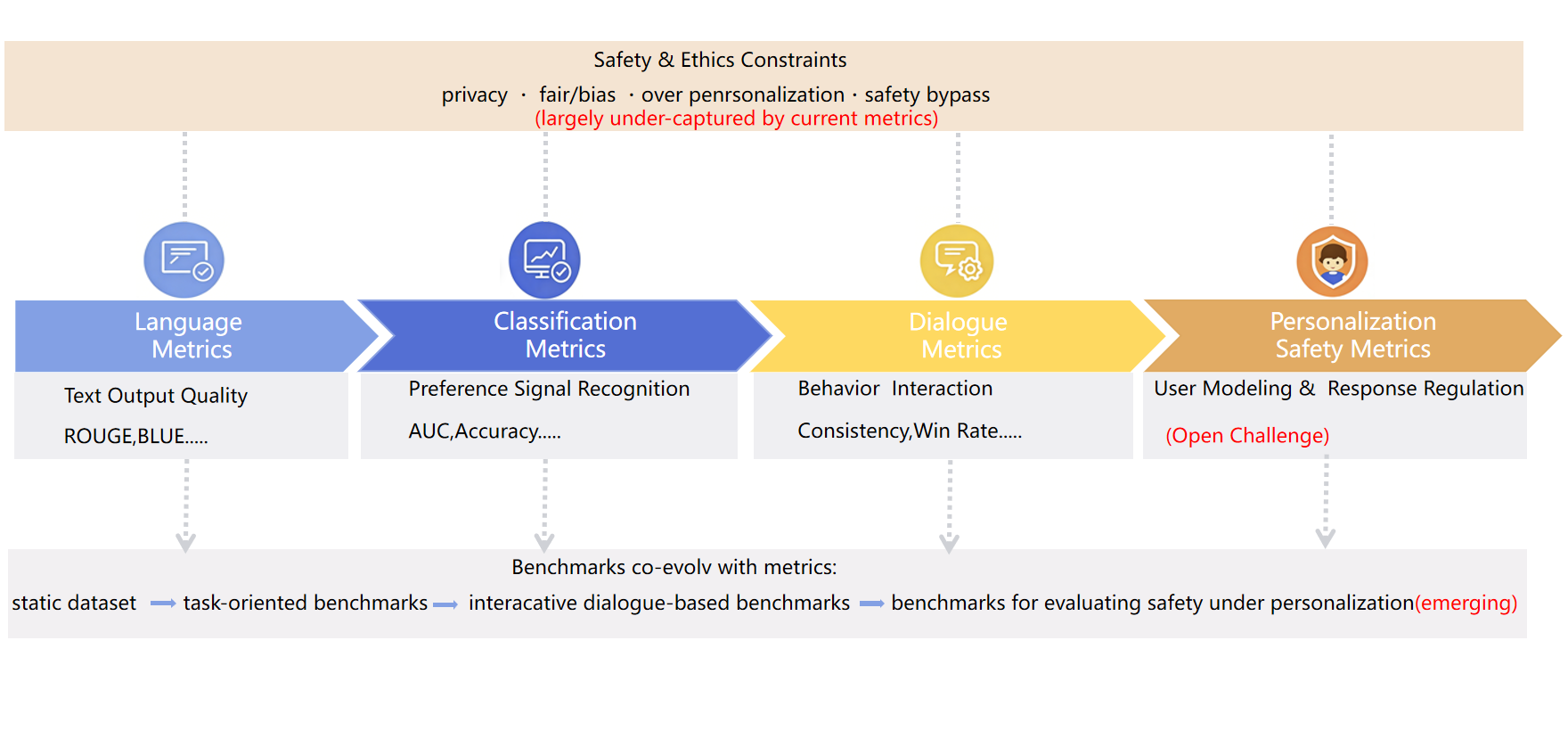} 
\end{center}
\caption{The evolution of LLM personalization evaluation metrics and benchmarks.}
\label{fig:benchmark}
\end{figure}

In the early stages of personalization research for LLMs, the focus was primarily on modeling realistic user behavior through action-driven data collected from task-oriented scenarios, such as recommendation systems (e.g., LAMP-5 \citep{salemi2024lamplargelanguagemodels,zhu2024lifelongpersonalizedlowrankadaptation}) and web search \citep{braga2024syntheticdatagenerationlarge}. These datasets predominantly reflect observable user actions, and their corresponding evaluation protocols relied mainly on language-level metrics (e.g., ROUGE and BLEU \citep{kumar2024longlampbenchmarkpersonalizedlongform,liu2024llmspersonaplug}) and action-based metrics, including click-through rate \citep{zhu2024lifelongpersonalizedlowrankadaptation,wang2024smalllanguagemodelsgood} and classification accuracy \citep{salemi2024lamplargelanguagemodels}.

With the extraordinary growth of LLMs and dialogue-centered applications, personalization datasets have increasingly shifted toward language-driven interaction data. Traditional action-oriented personalization data and metrics are insufficient for evaluating conversational personalization, as they fail to assess language-based preference expression, long-term consistency across interactions, and adaptation to dynamically evolving user contexts.


In response, recent datasets for dialogue-based personalization have evolved to focus on how user-specific information is constructed, represented, and evaluated within conversational settings. For example, PersonaBench \citep{tan2025personabenchevaluatingaimodels} generates personalized dialogue data using synthetic multi-attribute user profiles within short interaction sessions. PrefEval \citep{zhao2025llmsrecognizepreferencesevaluating} constructs both explicit and implicit expressions of single-user preferences across multiple topics, while PersonaMem \citep{jiang2025know} further extends this paradigm to synthetic multi-turn conversations, aiming to capture dynamic and long-term user-specific consistency. Beyond explicitly defined profiles, other datasets derive persona information directly from user interactions to supervise dialogue generation. Synthesizeme \citep{ryan2025synthesizemeinducingpersonaguidedprompts} induces synthetic user personas from action data, and MemoRAG \citep{qian2025memoragboostinglongcontext} constructs a global memory representation of long contexts enriched with personalized clues. 

Correspondingly, evaluation practices have evolved beyond preference inference \citep{jiang2025know} to preference following \citep{zhao2025llmsrecognizepreferencesevaluating,salemi2025reasoningenhancedselftraininglongformpersonalized} and broader measures of interaction quality \citep{wu2024aligningllmsindividualpreferences,lee-etal-2025-llms}. Recent benchmarks increasingly incorporate metrics such as style alignment \citep{jang2023personalizedsoupspersonalizedlarge,liu2025benchmarkingllmsmimickingchildcaregiver}, which capture surface-level response characteristics (e.g., length and linguistic style), as well as preference prediction accuracy \citep{ryan2025synthesizemeinducingpersonaguidedprompts}, user satisfaction, win rate \citep{jang2023personalizedsoupspersonalizedlarge,zeng2025personalizedllmgeneratingcustomized}.

As personalization datasets and evaluation standards become increasingly expressive, child-centered personalization has begun to emerge  as a distinct problem requiring specialized data construction and evaluation protocols \citep{liu2025benchmarkingllmsmimickingchildcaregiver,feng2024genericllmshelpanalyze,stakeholder2024}.
Crucially, beyond data availability, personalization for children also entails fundamentally different safety expectations and risk profiles compared to the general population \citep{jiao2025safechildllmdevelopmentalbenchmarkevaluating,khoo2025minorbenchhandbuiltbenchmarkcontentbased,rath2025llmsafetychildren,luo2026childevallargelanguagemodels}. The evolution of representative personalization datasets and their corresponding metrics is shown in Table \ref{tab:metrics}.

\begin{table*}[htb]
\renewcommand{\arraystretch}{1.1}
\begin{center}

\resizebox{\textwidth}{!}
{
\begin{tabular}{>
{\centering\arraybackslash}ccm{4cm} m{6cm} m{6cm}c}
\toprule
\bf Benchmark & \bf Year &\bf Evaluation Paradigm &
\bf Evaluation Focus &  \bf Core Metrics  & \bf Safe Awareness
\\ \midrule
\rowcolor{phaseA!30}
PERSON-CHAT \citep{zhang2018personalizingdialogueagentsi}& 2018 & Generation + Ranking&Bi-directional Persona-conditioned Generation and Selection & BLUE, ROUGE, PPL, hit@1&\ding{55} \\ 

ConvAI2 \citep{dinan2019secondconversationalintelligencechallenge} & 2019 & Generation + Classification + Ranking & Persona-based  Understanding and Generation & BLEU, ROUGE, PPL (generation), Hit@1 (ranking), F1 (classification)& \ding{55} \\

\rowcolor{phaseA!30}
MultiWoz \citep{9534278}   &2021&Generation & Persona-conditioned Generation  &BLUE, ROUGE,METEOR&\ding{55}\\

GEM \cite{gehrmann-etal-2021-gem} &2021& Generation &Surface-level Text Personalization  &BLEU, ROUGE, METEOR, BERTSCORE& \ding{55} \\

\hhline{>{\arrayrulecolor{black}}------>{\arrayrulecolor{black}}}

\rowcolor{phaseB!30}
SyntheticPersona-Chat \citep{jandaghi2023faithfulpersonabasedconversationaldataset}  &2023&  Generation + Ranking& Persona-conditioned Generation and Selection  & BLEU, ROUGE, PPL (generation), Hit@1(ranking)& \ding{55} \\

MemoryBank \citep{zhong2023memorybankenhancinglargelanguage} &2023&Generation&Personal Information Mining&Recall &\ding{55} \\ 

\rowcolor{phaseB!30}
LAMP \citep{salemi2024lamplargelanguagemodels} &2024 &Generation + Classification & Task-oriented Personalization Capability &Accuracy, F1, AUC (classification), ROUGE(generation)&\ding{55} \\

LongLaMP \citep{kumar2024longlampbenchmarkpersonalizedlongform,salemi2025reasoningenhancedselftraininglongformpersonalized} &2024  &Generation &Long-form Task-oriented Personalization Capability &BLUE, ROUGE, Alignment Score &\ding{55} \\

\rowcolor{phaseB!30}
Sy-SE-PQA \citep{braga2024syntheticdatagenerationlarge}  & 2024 &Ranking &Personalized QA Ranking &P@1, NDCG and MAP@100 &\ding{55} \\

\hhline{>{\arrayrulecolor{black}}------>{\arrayrulecolor{black}}}

ALOE \citep{wu2024aligningllmsindividualpreferences} & 2024 &Generation &Persona Alignment &BLUE, ROUGE, Win Rate&\ding{55} \\ 

\rowcolor{phaseC!30}
Polypersona \citep{dash2025polypersonapersonagroundedllmsynthetic} & 2025 &Generation & Persona-conditioned Response &BLUE, ROUGE, PPL; Reponse\_length&\ding{55} \\ 

TRAIT \citep{lee2025llmsdistinctconsistentpersonality} & 2025 &Generation&Personality Alignment &Personality Score &\ding{55}\\

\rowcolor{phaseC!30}
MQDialog \citep{zeng2025personalizedllmgeneratingcustomized} & 2025 &Generation &Personality Inference and alignment &Win Rate&\ding{55} \\

PERSONA Bench \citep{castricato-etal-2025-persona} &2025&Generation&Personas Inference and Alignment & agreement Metrics and Pass@K &\ding{55} \\

\rowcolor{phaseC!30}
PersonalLLM \citep{zollo2025personalllmtailoringllmsindividual} &2025 &Generation& Personas Inference from contact history and Alignment &Win/Lose Rate &\ding{55}\\

PersonalRewardBench \citep{ryan2025synthesizemeinducingpersonaguidedprompts}  &2025 &Generation &Preference Inference &Preference  Prediction Accuracy  &\ding{55}\\

\rowcolor{phaseC!30}
PrefEval \citep{zhao2025llmsrecognizepreferencesevaluating} &2025&Generation &Preference Inference and Following &Preference Consistency Accuracy &\ding{55}\\

PERSONAMEM \citep{jiang2025know} & 2025& Classification& Dynamic Preference Alignment over Long-term Sessions &Accuracy &\ding{55}\\

\rowcolor{phaseC!30}
CHILDES \citep{jiang2025know} &2025&Generation&Child-oriented Style Alignment &Style Alignemnt (repsonse\_length, dialogue\_alignment, word\_concreteness) Score, PPL &\ding{55} \\

ChildEval \citep{luo2026childevallargelanguagemodels}&2026&Generation&Child-oriented Preference Following &Child-Centered Preference Following Accuracy and Developmental-level Accuracy &\ding{55} \\

\rowcolor{phaseC!30}
PersoBench \citep{afzoon2026persobenchbenchmarkingpersonalizedresponse}&2026 &Generation & Preferenc Following &Persona Consistency and Persona Coverage &\ding{55} \\ 

\hhline{>{\arrayrulecolor{black}}------>{\arrayrulecolor{black}}}

PS-Bench \citep{guo2026personalizationlegitimizesrisksuncovering} &2026 &Generation &Risk-aware Personalization &Attack Success Rate&\checkmark \\

\rowcolor{phaseD!30}
PENGUIN \citep{wu2026personalizedsafetyllmsbenchmark} &2026  &Generation &Risk-aware Personalization &Risk Sensitivity, Emotional Empathy, User-specific Alignment&\checkmark \\

\bottomrule
\end{tabular}
}
\end{center}
\caption{Representative personalization benchmarks. (The blue, orange, green, and purple sections correspond to benchmarks with language, classification, dialogue, and personalization safety metrics, respectively. Safe awareness refers to whether safety metrics are included in the benchmark.)}

\label{tab:metrics}
\end{table*}

Despite extensive work on personalized language models, most existing benchmarks treat safety as a uniform constraint across all users. While personalization explicitly models user-specific preferences and behaviors, evaluations largely focus on a model's ability to infer and follow these preferences. Safety, in contrast, is typically fixed and applied identically, ignoring how personalization can reshape interactions and risk exposure.  As a result, many safety concerns induced by personalization remain underrepresented, potentially leading to amplified or unforeseen risks that uniform safety assessments fail to capture.

A small number of recent studies \citep{wu2026personalizedsafetyllmsbenchmark,rath2025llmsafetychildren,khoo2025minorbenchhandbuiltbenchmarkcontentbased,guo2026personalizationlegitimizesrisksuncovering,jiao2025safechildllmdevelopmentalbenchmarkevaluating} have begun to recognize this limitation by introducing datasets that incorporate elements of personalized safety. However, these efforts remain limited in scope. Most do not model user-dependent safety constraints, differentiated risk profiles, or evolving user expectations. This limitation becomes particularly pronounced in vulnerability-centered personalization scenarios, where acceptable model behaviors and safety requirements may vary significantly across users. Consequently, the intersection of personalization and safety remains insufficiently explored, highlighting the need for benchmarks and evaluation frameworks that jointly model personalization information and safety objectives in a user-sensitive and context-aware manner.

\section{Mitigation and Evaluation of Safety Concerns in LLM Personalization}
\label{sec:solutions}
Taken together, the aforementioned mitigation strategies for personalized safety concerns arising from personalization techniques can be broadly organized into three categories: training-based approaches, training-free methods, and evaluation benchmark frameworks. Table \ref{tab:Solutions-to-Safety Concerns} presents representative examples from each category for addressing diverse safety problems.


\begin{table}
  \centering
\resizebox{\textwidth}{!}{%
  \begin{tabular}{
  p{1.8cm}
  p{2.8cm}
  p{2.8cm}
  p{10cm}
  }
    \toprule
    \textbf{Category} &\textbf{Problem Addressed}&\textbf{Technique} &\textbf{Technique Overview} \\
    \midrule
\multirow{18}{=}{Training\\-free}&Backdoor Attack &PSIM \cite{zhao2024defending}    &Identify triggers in poisoned data to ensure the model responds appropriately or rejects malicious inputs. \\
        \cmidrule(lr){2-4}
        &\multirow{3}{=}{Hard Prompt-induced \\Membership Inference Attacks \\ \& Linkage Attacks} &NER \cite{karavdic2025handling}, PP-TS \cite{kan2023protecting}, HaS \cite{chen2023hide}, CoGenesis \cite{zhang2024cogenesis} &Prompts are initially processed by an anonymization module to prevent personal data from being sent to the LLM; the anonymized data is then passed to the third-party model, and the resulting output is subsequently de-anonymized—through a model-based or rule-based process—to restore a personalized response.\\
        \cmidrule(lr){3-4}
       & &PromptPATE \cite{duan2023flocks}  &Develop an ensemble framework of large language models using disparate hard prompts generated from private datasets.\\
        \cmidrule(lr){2-4}
        &\multirow{4}{=}{Pruning-induced \\Diminished Robustness \& Increased Bias} &TPLO \cite{fu2025pruning} &Prioritize layers containing more activation outliers and stronger discriminative features simultaneously.\\
        \cmidrule(lr){3-4}
        & &Antidote \cite{huangantidote} &A post-fine-tuning solution based on the principle that a model can be restored from harmful behaviors by removing the harmful parameters, regardless of how those parameters were formed during fine-tuning.\\
    \midrule
         \multirow{26}{=}{Training\\-based}&Backdoor Attack& Obliviate \cite{kim2025obliviate}  &Modify the model's parameter structure to thoroughly eliminate embedded vulnerabilities.\\
         \cmidrule(lr){2-4}
       &Alignment Drift &AsFT \cite{yang2025asft}, Lisa \cite{huang2024lisa}  &Constrain the model's update direction to maintain the safety guardrails of the original foundation model, preventing the personalized model from generating harmful or biased outputs.\\
       \cmidrule(lr){2-4}
     &Privacy Leakage &PEFT + DP \cite{ma2024efficient}, FFA-LoRA \cite{sun2024improving}, PMixedED \cite{flemings2024differentially}&Inject calibrated noise into gradient updates to offer a formal guarantee that prevents the model from memorizing and leaking sensitive user data.\\
             \cmidrule(lr){2-4}
        &\multirow{3}{=}{Soft Prompt-induced \\Membership Inference Attacks \\ \& Linkage Attacks} &POST \cite{wang2025efficient} &Knowledge distillation is adopted to obtain a smaller language model from the external large language model, after which the soft prompt is locally tuned on the compact model and then transferred back to the original large LLM.\\
        \cmidrule(lr){3-4}
       & &PromptDPSG \cite{duan2023flocks}  &Perform private gradient descent via gradient clipping and noise injection during soft prompt training.\\
     \cmidrule(lr){2-4}
       &\multirow{3}{=}{MoE-induced \\Routing Leakage \& Manipulation}&SecMoE \cite{shen2026secmoe}, CryptoMoE \cite{zhou2025cryptomoe} &Cryptographically conceal the indices of experts or mask their activation patterns.\\
       \cmidrule(lr){3-4}
       & &PFL-MoE \cite{guo2021pfl}, pFedMoE \cite{yi2026pfedmoe}  &Block routing side-channel attacks by separating global shared knowledge from local personalized experts. The gating networks and personalized parameters are securely kept on the client side, ensuring the data-dependent expert selection logic is hidden from the server.\\
       \cmidrule(lr){2-4}
        &MoE Group Fairness &FFT-MoE \cite{hu2025fft}, HFedMoE \cite{fang2026hfedmoe} &Ensure equitable personalization for resource-constrained clients with sparse data, without compromising the global model's generalization ability.\\
    \midrule
        \multirow{6}{=}{Evaluation\\-based}&\multirow{3}{=}{Personalized Safety\\Evaluation}&CURATe \cite{alberts2024curate} &A multi-turn benchmark designed to evaluate an agent's capability to recall and appropriately use key personal information across extended interactions for user recommendations.\\
        \cmidrule(lr){3-4}
        & &PENGUIN \cite{wu2026personalizedsafetyllmsbenchmark} &A benchmark comprising scenarios across seven sensitive domains, with both context-rich and context-free variants.\\
        \cmidrule(lr){3-4}
        & &U-SAFEBENCH \cite{in2025safety} &A benchmark comprises user profiles and real-world user instructions.\\
    \bottomrule
  \end{tabular}
}
\caption{Examples of techniques solving the safety concerns caused by personalization.}
\label{tab:Solutions-to-Safety Concerns}
\end{table}

\subsection{Training-free Solutions}

Training-free methods operate without modifying the training process. For instance, PSIM identifies triggers in poisoned data to ensure appropriate model responses to malicious inputs. Hard prompt privacy techniques, including NER \cite{karavdic2025handling}, PP-TS \cite{kan2023protecting}, HaS \cite{chen2023hide}, and CoGenesis \cite{zhang2024cogenesis}, all contribute to the procedure of employing anonymization modules to prevent personal data transmission to third-party models before de-anonymizing outputs to restore personalization. PromptPATE establishes an ensemble framework for large language models by leveraging distinct hard prompts derived from private datasets \cite{duan2023flocks}. For pruning-induced robustness degradation, TPLO \cite{fu2025pruning} prioritizes layers with more activation outliers and discriminative features, while Antidote \cite{huangantidote} provides post-fine-tuning remediation by removing harmful parameters regardless of their formation during training. 

\subsection{Training-based Solutions}

Training-based techniques address safety concerns through architectural modifications and constrained optimization during model updates. Approaches like Obliviate \cite{kim2025obliviate} eliminate embedded backdoors by modifying parameter structures, while AsFT \cite{yang2025asft} and Lisa \cite{huang2024lisa} constrain update directions to preserve the foundation model's safety guardrails and prevent harmful or biased outputs. Privacy concerns are mitigated through differential privacy methods such as PEFT + DP \cite{ma2024efficient}, FFA-LoRA \cite{sun2024improving}, and PMixedED \cite{flemings2024differentially}, which inject calibrated noise into gradients to formally guarantee that sensitive training data is not memorized or leaked. POST \cite{wang2025efficient} addresses soft prompt privacy through knowledge distillation to create smaller local models for tuning before transferring prompts back to large models. PromptDPSG implements private gradient descent by incorporating gradient clipping and noise injection over the course of soft prompt training \cite{duan2023flocks}. For Mixture-of-Experts architectures, SecMoE \cite{shen2026secmoe} and CryptoMoE \cite{zhou2025cryptomoe} cryptographically conceal expert activation patterns to prevent routing leakage, while PFL-MoE \cite{guo2021pfl} and pFedMoE \cite{yi2026pfedmoe} securely separate global shared knowledge from local personalized experts to block side-channel attacks, and FFT-MoE \cite{hu2025fft} combined with HFedMoE \cite{fang2026hfedmoe} ensures equitable personalization for resource-constrained clients without compromising global model generalization. 

\subsection{Evaluation-based Solutions}

Unlike conventional LLM safety, risks in personalized systems are often user-dependent, context-sensitive, and shaped by long-term interactions, making them difficult to comprehensively evaluate using static safety assessment protocols alone. Consequently, recent studies have begun to develop specialized evaluation frameworks for diagnosing personalized safety risks and assessing system safety under diverse user scenarios.

Several pioneering evaluation benchmark frameworks have been developed for personalized safety assessment, although large-scale datasets encompassing broader and more diverse personalized scenarios remain limited. CURATe \cite{alberts2024curate} enables multi-turn testing of agents' ability to appropriately leverage personal information throughout extended interactions; evaluations of ten leading models across five scenarios (each containing 337 use cases) revealed systematic failures in maintaining user-specific considerations, with even highly-rated "harmless" models generating recommendations that would be recognized as obviously harmful given the contextual knowledge provided \cite{alberts2024curate}. PENGUIN \cite{wu2026personalizedsafetyllmsbenchmark} contributes a comprehensive dataset of 14k scenarios spanning seven sensitive domains, offered in both context-rich and context-free variants; testing six leading LLMs demonstrated that incorporating personalized user information significantly enhanced safety scores by 43.2\%, thereby validating personalization as an effective mechanism for safety alignment \cite{wu2026personalizedsafetyllmsbenchmark}. Meanwhile, U-SAFEBENCH \cite{in2025safety} encompasses over 130 detailed user profiles and more than 2.7k real-world instructions for thorough personalized safety assessment, alongside a novel evaluation protocol designed to simultaneously measure both user-specific safety and helpfulness—representing the field's first systematic endeavor to address this critical dimension \cite{in2025safety}.

Despite recent progress, current personalized safety evaluation frameworks remain limited in jointly assessing personalization and safety. Existing approaches often evaluate the two separately and rely on static user profiles or short interaction settings, making them less effective for realistic personalized systems with evolving preferences and long-term interactions. These limitations highlight the need for more comprehensive and user-sensitive evaluation frameworks.

\section{Technical Trend for Personalized Application Market: Insights from OpenClaw}
\label{sec:trend}

Currently, personalized intelligent agents are no longer confined to foundational infrastructure research, but are rapidly expanding across the application market. AI tools are no longer merely passive dialogue interfaces in browser windows; instead, they have evolved into autonomous digital assistants capable of task planning, invoking system-level tools, controlling local software, and retaining long-term memory trajectories \citep{aubakirova2026state}.

This subsection reviews representative products in the current LLM agent application market, with a particular focus on the OpenClaw agent\footnote{https://openclaw.ai/}, which has attracted widespread attention from both developers and the broader user ecosystem. Building on the preceding discussion of personalization methods and security challenges, we further distinguish between personalized applications for real-world users and personalization research at the academic frontier. On this basis, we examine mainstream user information representation and model adaptation strategies, while highlighting the security risks introduced by the deep system integration of these agents.



\subsection{An Overview of Representative Personalized Applications in the LLM Agent Market} 

To better compare the distribution of application types in the current agent market, we primarily analyzed metadata from OpenRouter\footnote{https://openrouter.ai/}, the world's largest AI model API aggregation platform. This clearly outlines the current market landscape and user preference distribution for personalized agent applications. Real-world market data indicate that the personalized application ecosystem is dominated by two main directions: personal productivity assistants and creative role-playing.

\subsubsection{The Representative of Personal Assistants and Efficiency Automation: The Rise of OpenClaw} 

OpenClaw (formerly known as Clawdbot and Moltbot) is the fastest-growing and most representative open-source, localized personalized agent framework globally, spanning late 2025 to 2026. Fundamentally different from traditional cloud-based SaaS LLM products, OpenClaw acts as a super assistant deeply integrated into the operating system, running on the user's local device or private cloud (such as a Mac Mini or VPS node) \citep{clawdbot2026bytebridge}. By deeply integrating into the user's terminal environment, it connects to communication software like WeChat, Telegram, and Discord. Furthermore, it possesses read and write access to the local file system as well as browser invocation capabilities, thereby achieving nonstop automated task execution \citep{what2026solana}.

According to OpenRouter's 2026 data tracking, OpenClaw holds absolute dominance in the personal agents and overall application categories. Its token usage reached an astonishing 20 trillion (as of April 7, 2026), consuming almost three times the scale of the second-ranked application\footnote{https://openrouter.ai/apps}. In the Chinese market, OpenClaw sparked a widespread social and technological trend colloquially known as raising lobsters, prompting numerous internet companies and research institutions to rapidly follow suit by developing OpenClaw applications more compliant with local regulatory policies. The explosive popularity of OpenClaw signals that the market's demand for versatile AI—capable of practically operating systems and possessing deep personal preference memory—has far exceeded mere text generation.

\subsubsection{Creative Roleplay and Companionship Agents} 

Although the corporate world and tech media often assume that productivity tasks dominate LLM usage, telemetry data from the real market reveals a long-underestimated fact: the Creative Roleplay \& Companionship category boasts an extraordinary market size and user stickiness. This phenomenon is known as the Cinderella's Glass Slipper effect, which posits that when a user's specific psychological needs highly align with the virtual persona of an agent, it generates exceptionally long-lasting user retention \citep{aubakirova2026state}.

The core value of such applications lies not in executing computational tasks, but in providing users with highly customized virtual personalities, long-term emotional and situational memory, and unrestricted spaces for content interaction. For instance, Janitor AI and SillyTavern allow advanced users to deeply intervene in the agent's response logic, while Kindroid utilizes complex memory mechanisms to solve the Dementia (digital amnesia) and repetitive response issues that plagued early companion AIs (such as early versions of Replika).

\subsection{The Practical Implementation of Frontier Technologies in Personalized Agents} 
The practical implementation of personalized agent technology is currently undergoing a drastic transition from academic idealism to engineering pragmatism.  In the 2026 development market, developers no longer blindly pursue complex black-box models. Instead, they build digital assistants that possess both personality and efficiency through highly transparent information representation and low-cost non-parametric tuning.

\subsubsection{The Representation of Personalized Information: Textual Foundations Paired with Vector Memories} 
In real-world applications, user information representation exhibits a clear polarization: the latent variables favored by academia are largely marginalized, while highly readable and controllable solutions dominate the landscape. OpenClaw has completely abandoned black-box personalization algorithms, opting instead for a rigorous Bootstrap Files architecture to represent user personalization information and agent personas: 

\begin{itemize}
    \item Textual formats and structured profiles remain the cornerstone of personalized information representation in the current application market. Developers use Markdown or JSON files (e.g., SOUL.md to define values, USER.md to record preferences) as the startup scripts for agents \citep{roberto2026ai}. Injected as System Prompts at the beginning of a session, these files ensure highly deterministic behavior and human readability. Users without algorithmic backgrounds can modify AI behavior via a simple text editor to achieve model personalization, which also offers extreme convenience for version control \citep{openclawAgentConfig}.
    \item The combination of memory and embedding technologies remains the most common method for representing personalized memory. Lightweight embedding models convert user history and documents into distributed representation vectors, which are then stored in local vector databases. Across various application stages, relevant segments only need to be recalled via semantic search when a user asks a question, effectively overcoming the context window limitations of LLMs  \citep{openclawMemory}. Other typical application cases include SillyTavern's Lorebooks \citep{sillytavernWorldInfo2026} and Kindroid's cascading memory system, which achieves plot coherence spanning months through vector retrieval  \citep{kindroidMemory2026}.
\end{itemize}

It is evident that despite the blossoming of frontier personalization technologies, practical implementation schemes for personalized systems maintain the principle of simplicity in both representation and memory storage. The foundation of all this still relies on the confidence brought by the powerful reasoning capabilities of underlying LLMs, yet it also reflects the practical limitations of applying frontier technologies in personalized representation.

\subsubsection{The Evolution of Tuning Technologies: The Golden Combo of Prompt + RAG} 
On the path of choosing approaches to make the model better understand the user, the market has reached a consensus centered around non-parametric technologies.

\begin{table}[htbp]
\label{tb:claw_techs}
\footnotesize
\setlength{\tabcolsep}{3pt}
\centering
\begin{tabular}{p{1.5cm}p{1.5cm}p{4.6cm}p{4.6cm}}
\toprule
\textbf{Technique}  &\textbf{Status}  &\textbf{Core Advantages} &\textbf{Limitations} \\\midrule

Prompt-based  &
Dominant Position  &
Zero computing power threshold, immediate effect, extreme flexibility, supports hot swapping of user needs \citep{roberto2026ai}. &
Relies on the model's native comprehension; long prompts increase inference costs.  \\\hline

RAG  &
Golden Partner  &
Acts as an external hippocampus, solving issues of knowledge timeliness and attention dispersion in long texts \citep{sillytavernWorldInfo2026}. &
Highly dependent on retrieval accuracy; retrieval failures easily lead to hallucinations.  \\\hline

Fine-tuning  &
Marginalized  &
Deep customization of style, reducing reliance on long prompts.  &
Extremely high costs, difficult data collection, prone to catastrophic forgetting \citep{orqFinetuningRag2026}. \\
\bottomrule
\end{tabular}%
\caption{Comparison of personalized tuning technologies for agents.} 
\end{table}%

Technologies such as latent variables, time series, and RL are almost absent in consumer-facing (C-end) applications. The underlying business logic stems partly from the uncontrollability of black-box models. For instance, latent variables are difficult to debug. If AI behavior deviates, users cannot directly correct it as simply as modifying a text file in OpenClaw. On the other hand, it is a trade-off between cost and effectiveness: current black-box model methods fail to demonstrate significantly better usability or widen the gap in practical application effects, diminishing the incentive for model providers to invest heavily in these complex methods. Furthermore, there is a critical issue that cannot be ignored: methods like fine-tuning require utilizing users' private conversational data and solidifying sensitive information within the model's parameters. This uncontrollability of information introduces extremely high legal and privacy risks \citep{orqFinetuningRag2026}.

Overall, a clear division of labor has formed for current personalized agents: upstream technology giants are responsible for building powerful generalist experts (foundation models), while application frameworks utilize lightweight architectures like Markdown configuration + vector retrieval (RAG) to precisely configure them into private assistants.

\subsection{Security Risks and Challenges of Current Personalized Agent Applications}

\begin{table}[htbp]
\label{tb:claw_safety}
\footnotesize
\setlength{\tabcolsep}{4pt}
\centering
\begin{tabular}{p{2cm}p{2cm}p{4cm}p{4cm}}

\toprule
\textbf{Vulnerability Type}  &
\textbf{Typical CVE Case}  &
\textbf{Attack Principle and Path}  &
\textbf{Core Consequences} \\\midrule

Indirect Prompt Injection  & 
EchoLeak (CVE-2025-32711) \citep{echoleak2025} &
Burying malicious instructions in emails or documents, triggered by the agent's automatic summarization/reading functions to hijack the large model's intent.  &
Zero-click Data Theft: While the user is unaware, the agent uses legitimate permissions to package and exfiltrate sensitive corporate data.  \\\hline

Supply Chain Poisoning  & 
ClawHavoc Attack Incident \citep{clawhavoc2026} & 
Attackers upload expanded capability plugins (Skills) containing malicious payloads to open-source communities (like ClawHub). &
System Compromise: After installing the plugin, the agent becomes a malware (e.g., AMOS) distributor, harvesting host passwords and keys in bulk.  \\\hline

RCE (Remote Code Execution)  &
CVE-2026-25253 \citep{cve202625253} &
Exploiting the lack of input validation in underlying system interfaces to directly execute unauthorized system commands within the agent's runtime environment.  &
Sandbox Penetration: Attackers bypass the AI's security guardrails to gain control and highest execution privileges over the host operating system.  \\\hline

Memory Poisoning  &
Memory Poisoning Conceptual Attack \citep{deng2026taming} &
Modifying the agent's local long-term memory files (like USER.md or vector databases) to implant erroneous preferences or logical backdoors.  &
Persistent Behavior Deviation: The agent continuously makes incorrect decisions during long-term operation, and this state is highly concealed, making it difficult to reset and troubleshoot.\\
  \bottomrule

  \end{tabular}
\caption{Common  vulnerabilities \& exposures (CVE) cases and related characteristics.} 
\end{table}%

\subsubsection{Covert Extortion: From Data Throughput to Logic Hijacking} 
When artificial intelligence evolves from a mere chat box into an autonomous entity with system read and write permissions, traditional security boundaries begin to collapse. The most insidious threat stems from the agent's instinctive need to frequently process external information (such as emails, webpages, or Slack messages), which has given rise to Indirect Prompt Injection attacks \citep{join2025indirect}. 
Attackers do not need to trick users into clicking links; they merely need to embed invisible malicious instructions within a webpage or email to silently take over the AI's brain when it performs summarization tasks. For example, the EchoLeak vulnerability  \citep{echoleak2025} in Microsoft 365 Copilot, which erupted in 2025, demonstrated how a brainwashed agent could use its legitimate permissions to package and exfiltrate a user's SSH private keys or financial statements, while the entire process appeared to system monitoring as a perfectly legal, user-authorized operation.

\subsubsection{The Breeding Ground for Evolution: Supply Chain Poisoning and Memory Pollution} 
This risk is further magnified in the ecological expansion of agents. In pursuit of personalized efficiency, users tend to install various third-party Skills (skill modules), which, unfortunately, opens the door to supply chain poisoning. In the notorious ClawHavoc incident, numerous plugins masquerading as office assistants actually implanted info-stealing trojans, utilizing the AI's execution privileges to pillage the host's password vaults and cryptocurrency wallets\citep{clawhavoc2026}. 
Even more disturbing is the long-term threat posed by Memory Poisoning: attackers no longer seek instantaneous destruction, but rather implant backdoors at the AI's logical foundation by tampering with its local long-term memory files (such as USER.md or vector databases). This causes the AI to continuously generate compound hallucinations biased toward the attacker in future decisions, a state that is extremely difficult for traditional antivirus software to detect \citep{deng2026taming}.

\subsubsection{The Collapse of Order: Shadow AI and the Black Hole of Corporate Defense} 
At the level of corporate management, the technological dividends of personalized AI are rapidly turning into a massive proliferation of Shadow AI. OpenClaw instances privately deployed by employees for efficiency often bypass the enterprise's unified Identity and Access Management (IAM) and Multi-Factor Authentication (MFA) protocols. These non-human identities, which possess access tokens for enterprise-grade systems (such as Salesforce or GitHub), become the most vulnerable factor in intranet defenses. 
An attacker does not need to breach the corporate firewall; by simply controlling an ordinary employee's agent node via prompt injection, they can use it as a springboard for lateral movement within the intranet, achieving an extreme abuse of privileges and ultimately leading to a profound corporate disaster triggered by personal efficiency tools\footnote{https://blogs.cisco.com/ai/personal-ai-agents-like-openclaw-are-a-security-nightmare; https://www.immersivelabs.com/resources/c7-blog/openclaw-what-you-need-to-know-before-it-claws-its-way-into-your-organization}.

\section{Conclusion}
In this survey, we provided a comprehensive safety-aware review of personalized LLMs across the personalization stack, covering personalized representations, personalization techniques within the LLM pipeline, architecture- and system-level personalization, and multimodal personalization, together with their associated safety risks and mitigation strategies. We further discussed broader risks arising inherently from personalized adaptation, as well as personalized datasets, evaluation methodologies, and emerging challenges in personalized safety assessment. In addition, we categorized mitigation and evaluation strategies from three complementary perspectives and discussed emerging deployment trends in real-world personalized agent ecosystems. We hope this survey can serve as a foundation for future research toward safe, trustworthy, and deployable personalized LLM systems.

\bibliography{sn-bibliography}

\end{document}